\documentclass[sigconf,nonacm]{acmart}
\usepackage{color,soul}
\usepackage{graphicx}
\usepackage{multirow}
\usepackage{subcaption}
\usepackage{comment}
\usepackage{xcolor}
\usepackage{enumitem}
\usepackage{listings}
\usepackage{amsthm}
\usepackage{colortbl}
\usepackage{longtable}
\usepackage{booktabs}
\usepackage{hhline}
\newcommand{\simon}[1]{\textcolor{red}{Simon: #1}}
\newcommand{\sh}[1]{\textcolor{red}{Sha: #1}}

\setcopyright{none} % No copyright notice required for submissions
\begin{document}
% \title{Template for ACM CCS} % TODO: replace with your title
% \title[Am I a Real or Fake Celebrity?]{Am I a Real or Fake Celebrity? Evaluating the Robustness of Commercial Face Recognition Web Services with Deepfakes}
% \title[Am I a Real or Fake Celebrity?]{Am I a Real or Fake Celebrity? \sh{Measuring} the Robustness of Commercial Face Recognition Web Services with Deepfakes}
\title[Am I a Real or Fake Celebrity?]{Am I a Real or Fake Celebrity?}
\subtitle{Measuring Commercial Face Recognition Web APIs under Deepfake Impersonation Attack}

%\subtitle{Measuring the Vulnerabilities of Commercial Recognition APIs under Deepfake Impersonation Attack}

\author{Shahroz Tariq}

\affiliation{%
  \institution{Department of Computer Science\\ and Engineering\\ Sungkyunkwan University}
  \city{Suwon} 
  \country{South Korea} 
}
\email{shahroz@g.skku.edu}

\author{Sowon Jeon}
\orcid{1234-5678-9012}
\affiliation{%
  \institution{Department of Computer Science\\ and Engineering\\ Sungkyunkwan University}
  \city{Suwon} 
  \country{South Korea} 
}
\email{soonej40@g.skku.edu}

\author{Simon S. Woo}
\authornote{Corresponding Author}
\affiliation{%
  \institution{Department of Applied\\ Data Science\\ Sungkyunkwan University}
  \city{Suwon} 
  \country{South Korea}
  }
\email{swoo@g.skku.edu}

%measuring the performance of API
%measuring the attack performance
%measuring the defense performance
%Web API measurement 
\begin{abstract}
Recently, significant advancements have been made in face recognition technologies using Deep Neural Networks. As a result, companies such as Microsoft, Amazon, and Naver offer highly accurate commercial face recognition web services for diverse applications to meet the end-user needs. Naturally, however, such technologies are threatened persistently, as virtually any individual can quickly implement impersonation attacks. In particular, these attacks can be a significant threat for authentication and identification services, which heavily rely on their underlying face recognition technologies' accuracy and robustness. Despite its gravity, the issue regarding deepfake abuse using commercial web APIs and their robustness has not yet been thoroughly investigated. This work provides a measurement study on the robustness of black-box commercial face recognition APIs against Deepfake Impersonation (DI) attacks using celebrity recognition APIs as an example case study. We use five deepfake datasets, two of which are created by us and planned to be released. More specifically, we measure attack performance based on two scenarios (targeted and non-targeted) and further analyze the differing system behaviors using fidelity, confidence, and similarity metrics. Accordingly, we demonstrate how vulnerable face recognition technologies from popular companies are to DI attack, achieving maximum success rates of 78.0\% and 99.9\% for targeted (i.e., precise match) and non-targeted (i.e., match with any celebrity) attacks, respectively. Moreover, we propose practical defense strategies to mitigate DI attacks, reducing the attack success rates to as low as 0\% and 0.02\% for targeted and non-targeted attacks, respectively.

\end{abstract}


\maketitle

\section{Introduction}
\label{sec:intro}
% \simon{motivation...1) Fake Images identy verification application.....impersonation for ID application. 2) DF can be combined with fake news and other services to create riot unrest, other political stuff..3) Fake videos are more severe...}
% \simon{web service}

%\hl{Now, anyone with a few images of a public figure can create a fake video of them, which can create riots and public unrest if combined with fake news. Commercial face recognition API and web services from MS and AWS are majorly used for face verification and identification by clients such as entertainment companies for indexing images and videos, humanitarian groups to identify and rescue human trafficking victims, and law enforcement agencies to protect the public's safety.}

%\hl{(\textbf{measuring the performance of API, measuring the attack performance, measuring the defense performance, Web API measurement})}
%\hl{add About BLACK BOX attack}
Deep learning algorithms have contributed significantly to the advancement of a wide range of applications, including computer vision, speech processing, and natural language processing. In particular, due to the success of face recognition and identification algorithms, companies such as Amazon, Google, Microsoft, Baidu, and Naver, have unveiled commercial face recognition web APIs for various applications~\cite{AmazonAPI,MSAPI,NaverAPI}. These services are widely adopted for face verification and identification in a number of areas, such as the entertainment industry for indexing images and videos~\cite{AmazonMediaAnalysis}, humanitarian groups for identifying and rescuing human trafficking victims~\cite{TrafficJam}, and law enforcement agencies for protecting the public's safety~\cite{AmazonUsecase}. Moreover, Microsoft has released a mobile app named TwinsOrNot in 2015 which helps users find celebrities who resemble them~\cite{twinsornot}. Naturally, these web APIs have a wide range of clients. For example, Kyodo News MediaLab aims to build broader image search services such as auto-tagging of celebrities and classification using custom labels from Amazon Rekognition APIs~\cite{AmazonMediaAnalysis}; the Marinus flagship software~\cite{TrafficJam} offers a tool named Traffic Jam, which can be used by law enforcement agencies to investigate sex trafficking. Using Amazon Rekognition API, investigators can take appropriate actions by searching through millions of records in just a few seconds to identify potential criminals~\cite{AmazonSafety}.
%The results of those APIs provide high reliability due to the advanced technology of the companies.

On the other hand, the rapid development of synthetic image and video generation methods has prompted significant concerns over the authenticity of readily available contents on social media and video hosting sites. Although fabrication and manipulation of digital images and videos are not new, the advent of so-called deepfake methods has made the process of creating convincing fake videos much easier and faster. It is therefore now possible for anyone with a few images of a victim from the general public to create his/her fake videos, which can result in riots and social unrest, especially when combined with fake news~\cite{schwartz_2018,DeepfakeRiot}. In fact, there are positive and creative use cases of deepfakes, such as Disney's plan to replace traditional visual effects (VFX) with deepfakes~\cite{DisneyDeepfake} and the creators of the HBO documentary `Welcome to Chechnya'~\cite{WelcometoChechnya} utilizing deepfakes to anonymize LGBTQ prosecution survivors. However, deepfakes are highly likely to be misused, for example, to shift political opinions using deepfake videos of Richard Nixon's moon disaster speech~\cite{NixonDeepfake} and of Donald Trump's regrets on climate change~\cite{TrumpClimateDeepfake}.
%\hl{Making matters worse,?} 96\% of the deepfake on the Internet \hl{is shown to be} the non-consensual deepfake pornography according to a recent study~\cite{96percentDeepfake}.
To make matters worse, 96\% of the deepfake on the Internet are non-consensual deepfake pornography~\cite{96percentDeepfake}. Most of these deepfake pornography feature women, often celebrities, and are used to degrade, defame, and abuse them~\cite{Newdeepfakesresearch}. 

%In late 2017, when a Reddit account posted non-consensual pornographic videos generated using a deep neural network-based face-swapping method~\cite{FaceSwap} was the first time these deepfakes caught the public's attention. Consequently, the term deepfake generally describes all types of AI-generated impersonating videos.

In addition to privacy violations, we believe that this deepfake technology can be further misused to carry out impersonation attacks. The matter regarding deepfake abuse is not only of great significance from a security and privacy perspective, but also of grave urgency, which must be addressed before further exacerbating misinformation. More specifically, these days, the face recognition APIs have become more essentially integrated into daily life. Therefore, in this work, we showcase a Deepfake Impersonation (DI) attack with a scenario of deceiving commercial face recognition APIs. We perform the attack on the well-known celebrity recognition APIs from Microsoft, Amazon, and Naver; because it is the quintessence of their face recognition capabilities. Nevertheless, this attack can be easily generalized and extended to the masses by attacking the general face recognition APIs.%~\hl{(See Section}~\ref{sec:motivation} for more details). 

Therefore, in this work, we aim to understand how different celebrity recognition APIs respond to celebrity deepfakes. Based on extensive experiments with five deepfake datasets, we demonstrate that it is relatively easy to evade and fool celebrity recognition APIs, achieving up to a 99.9\% attack success rate on the certain commercial celebrity recognition API in a black-box setting. Among the five datasets used to carry out impersonation attacks, two are publicly available benchmark datasets, one is collected online, and the remaining two are created by ourselves. Lastly, we also propose defense strategies that can significantly reduce the impact of DI attacks. Our main contributions are summarized as follows:
%Meanwhile, people rely on the search APIs provided by Google~\cite{GoogleCloudVisionAPI}, Microsoft~\cite{MSAPI}, and Amazon~\cite{AmazonAPI} to find the identity of a celebrity from an image.
%However, there has been no standard method to evaluate the face recognition web APIs' vulnerability against the developed deepfake images. The topic is not only of great significance but also of technical interest. Therefore, in this work, we aim to understand how different celebrity recognition APIs respond to celebrity deepfakes. Further, we show that it is easy to evade and fool Celebrity recognition APIs and provide a solution to defend against them.
\begin{itemize}[leftmargin=10pt]
    \item \textbf{Deepfake Impersonation Attack: } We introduce a novel impersonation attack to deceive commercial face recognition APIs using deepfakes.
    % We introduce a novel approach to deceive  and measure the performance of commercial celebrity recognition web services against deepfakes emulating impersonation attacks.
    We also create two novel datasets, Celebrity First Order Motion (\textit{CelebFOM}) and the Celebrity Blend (\textit{CelebBlend}), for evaluation\footnote{We wish to open our own datasets and share the download link on our server to foster future research in this area. But it would break the double-blind policy. Instead, we upload some samples of our dataset to our anonymized code repository\footnotemark. Once accepted, we will make our GitHub link public along with the link to the full datasets.}.

    % We also create the Celebrity First Order Motion (CelebFOM) and the Celebrity Blend (CelebBlend) dataset for evaluation 

    %\st{We also considered real-world deepfake data by collecting available celebrity deepfakes in an ethically and legally proper method.} 
    
    % \item \textbf{Novel Evaluation Dataset:} \has{
    % We used facial replacement, reenactment, and synthesis as attack methods and generated two novel datasets in this work for evaluation(write something similar), which are Celebrity First Order Motion (CelebFOM) and the Celebrity Blend (CelebBlend)}\footnote{We plan to release our datasets after acceptance of this paper to foster future research in this area.}. %\st{We also considered real-world deepfake data by collecting available celebrity deepfakes in an ethically and legally proper method.} 
    
    \item \textbf{Extensive Measurement and Evaluation of Commercial Web APIs: } We evaluate how popular major commercial celebrity recognition web APIs from Amazon, Microsoft, and Naver respond to celebrity deepfakes in a black-box setting, using four different metrics. We demonstrate that celebrity deepfakes fool all three APIs with an attack success rate of up to 78.0\% and 99.9\% on targeted and non-targeted attacks, respectively. We also analyze the deepfakes and real images using the APIs' face similarity module. 
    
    \item \textbf{Defense Mechanism: } We present a defense mechanism against the DI attack and compare several defense strategies. We show that our proposed mechanism significantly decreases the actual attack success rate from 99.6\% to 1.79\% for all datasets combined.
\end{itemize}

We provide the code used for attacks and other test code to query the ML service in the anonymized code repository\footnotemark[\value{footnote}]. Also, we list the external data sources, models, and papers evaluated in our work on the repository.

% \footnote{}
\footnotetext{We will share the code after the paper acceptance.}
% \footnotetext{\url{https://anonymous.4open.science/r/a3be4ec6-b746-4589-bb4a-03040bd21ecf}}
%We are releasing the experimental details and our code for the evaluation of commercial celebrity recognition web API\footnote{\url{https://anonymous.4open.science/r/a3be4ec6-b746-4589-bb4a-03040bd21ecf/}}. 
%\simon{maybe first and second contribution can be combined. Ideally 3 main contributions are the best.}
% \hl{fix at end}
% The rest of our paper is organized as follows: in Section~\ref{sec:related}, we provide background knowledge about different deepfake attack generation methods and literature review on commercial face recognition web APIs; in Section~\ref{sec:approach}, we present the overview of our attack framework with three different types of deepfake generation methods and two attack scenarios; in Section~\ref{sec:experiment}, we explain our experiments based on the attack analysis metrics and present the results in Section~\ref{sec:results}; in Section~\ref{sec:defense}, we propose a defense mechanism against deepfake impersonation attacks using deepfake detection methods and show the corresponding results; in Section~\ref{sec:discussion}, we discuss the limitation and ethical issues, and offer our conclusion in Section~\ref{sec:conclusion}. %\siho{(Some of the section numbers appear as "??". Don't forget to label the sections and check later.)}
\section{Related Work}
\label{sec:related}
Our research covers several areas, such as deepfake generation, face recognition, and attacks on commercial celebrity recognition APIs. Throughout this section, we briefly cover the history and related works directly relevant to our work.

\noindent
\textbf{Deepfake Generation Methods. }
Mirsky and Lee~\cite{WenkeSurvey} define deepfake as ``believable media generated by a deep neural network.'' They broadly classify deepfake into three categories: reenactment, replacement, and synthesis. Reenactment-based methods transfer the expression, mouth, gaze, pose, or body of the reference image to the target. As an example of the reenactment, Zakharov et al.~\cite{VoxTalkingHead} propose a new system that generates highly realistic and personalized talking head of celebrities.
%They performed lengthy meta-learning with GAN-based framework on a large video dataset.
%Furthermore, it is possible to generate a frame with a few- and one-shot learning of previously unseen neural talking head models as adversarial training problems using efficient generators and discriminators. 
On the other hand, replacement-based methods are to replace the target's content with that of the reference, preserving the reference's identity. Replacement-based methods such as face swaps are widely misused to generate revenge or celebrity porn. FaceSwap~\cite{Deepfakes} is one widely used application of replacement. 
Moreover, face synthesis is where the deepfake is created by altering one or multiple reference images. FaceApp~\cite{FaceApp2018} is one example of synthesis, enabling users to alter their or morph different faces to make a new synthetic identity. 
We use facial replacement, reenactment, and synthesis as attack methods and also generate two new additional datasets in this work. We explain more details of each dataset in Section~\ref{sec:approach}.

\begin{figure*}
    \centering
    \includegraphics[width=1\linewidth]{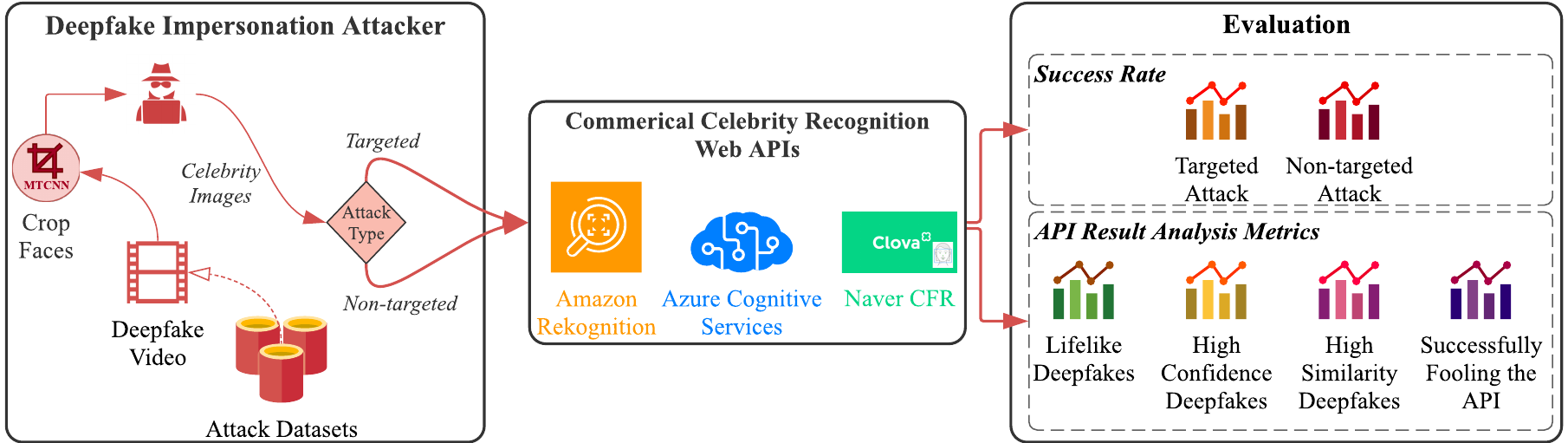}
    \caption{Overview of our Deepfake Impersonation attack and evaluation framework for commercial face recognition APIs.}
    \label{fig:Overview}
    % \vspace{-10pt}
\end{figure*}
\noindent
\textbf{Commercial Face Recognition Web APIs. }
Amazon Rekognition~\cite{AmazonAPI}, Microsoft Azure Cognitive Services~\cite{MSAPI}, Google Cloud Vision~\cite{GoogleCloudVisionAPI}, Baidu AI~\cite{BaiduAI}, and Naver Clova~\cite{NaverAPI} provide various image analysis web APIs for the uses in object, scenes, faces detection and recognition as well as celebrity recognition. Many people use these APIs for visual search, image classification, and recognizing celebrities.
Also, Baidu~\cite{BaiduAI} and Naver~\cite{NaverAPI} are the leading search engine and portal service providers in China and South Korea, respectively, where they both offer web APIs for celebrity recognition. Naver Face Recognition web API can also detect various face image attributes, and Microsoft Azure also provides face image analysis information. 

There are several commercial face recognition APIs. However, in this work, we choose Microsoft (\textit{MS}), Amazon (\textit{AWS}), and Naver (\textit{NAV}) celebrity recognition APIs for the following reasons: First, these three web APIs all commonly offer a service to recognize celebrity faces, returning comparable face similarity scoring metric. This allows us to evaluate APIs in a fair manner. Second, celebrity face images are easily obtainable compared to ordinary people and can be easily spoofed to generate deepfakes and further perform impersonation attack. Moreover, they can be evaluated with the live commercial celebrity recognition APIs. Third, we believe we have representative use cases incorporating major web service companies from different continents. We hypothesize that MS and AWS can effectively identify the western celebrities (Europe or America), while NAV can recognize Asian celebrities more effectively.

On the other hand, we do not use Baidu in this work, as they provide a different output format and detection characteristics, making it difficult to compare with the other three web APIs. Also, Google Cloud Vision API for celebrity recognition~\cite{GoogleCloudVisionAPI} is currently under restricted access and is only available to media and entertainment companies or authorized partners. In fact, we have made a formal request to access the API, but the permission was not granted; therefore, we could not consider Google Cloud Vision API in this work. However, we believe the three APIs we chose can be fairly compared and are representative to evaluate the DI attack.

\noindent
\textbf{Attacks on Face Recognition Web APIs. }
Several studies have recently focused on uncovering vulnerabilities of commercial face recognition web APIs, taking different approaches. However, most of the research is focused on finding vulnerabilities using adversarial attacks~\cite{dong2019efficient,zhong2020towards,kim2018george,SeeingnotBelieving,Lowkey}. In comparison, using the impersonation attack to evaluate the APIs is less explored. We believe that the impersonation attack is equally, if not more important, as it has been one of the critical problems for face recognition. And many multi-factor authentication methods adopt face verification as a primary or secondary authentication method.

Now, we will briefly cover how adversarial attacks are used to attack these APIs. Dong et al.~\cite{dong2019efficient} evaluate the robustness of state-of-the-art face recognition models with the decision-based black-box attack setting, which can only acquire hard-label predictions by sending queries to the target model. %They conducted an impersonation attack to face recognition API using adversarial noise.
Zhong and Deng~\cite{zhong2020towards} investigate the characteristics of transferable adversarial attacks in face recognition by showing that feature-level methods are superior to label-level methods.
%They also proposed DFANet, a dropout-based method, generating a new set of adversarial face pairs that can successfully attack and spoof four commercial APIs by applying DFANet to the LFW database~\cite{LFWTech}.
GenAttack~\cite{kim2018george} is proposed as an attack algorithm to create adversarial examples using an evolutionary genetic algorithm on celebrity images and evaluate those with celebrity recognition web APIs from Amazon~\cite{AmazonAPI} and Naver~\cite{NaverAPI}.
%GenAttack starts from generating adversarial examples by iteratively adding noise to the original images, unlike DNN based adversarial example generations. The GenAttack successfully deceived Amazon and Naver APIs with a success probability of 86.6\%. 
Also, Xiao et al.~\cite{SeeingnotBelieving} present the scaling algorithm that can successfully perform attacks against famous cloud-based image services and cause noticeable misclassifications. In addition, Lowkey~\cite{Lowkey} is proposed as an evasion tool for protecting users' images from adversarial attack against commercial facial recognition APIs. 
Yaun et al.~\cite{yuan2019stealthy} show the effectiveness of adversarial promotional porn images (APPI) as an attack method that can evade the existing explicit content detectors APIs. Recently, more adversarial attack based methods have also been proposed to attack web APIs~\cite{han2019nickel,yang2019neural,yu2020cloudleak}.

However, to the best of our knowledge, no study exploits deepfakes as an impersonation attack method to attack face recognition web APIs. We are the first to demonstrate such attacks and reveal the potential danger from those on commercial APIs that are serviced as of now.
%\st{As deepfake is frequently abused for impersonation, considering deepfakes as a threat to face recognition API is very natural.} 
In this work, we design the practical impersonation attack framework and evaluate the attack scenarios using high-quality deepfake images. Moreover, we successfully propose and demonstrated a defense method against DI attack on face recognition web APIs.

%in comparison with GenAttack ~\cite{kim2018george} and DFANet ~\cite{zhong2020towards}, we regarded deepfake itself as an adversarial example and tried several deepfake datasets on commercial web APIs.

\section{Motivations of DI Attacks}
\label{sec:motivation}
In this section, we describe the concrete motivations behind Deepfake Impersonation (DI) attacks.
Specifically, we consider the following realistic attack use case scenarios:
% \begin{enumerate}[wide=\parindent]%leftmargin=14pt]
    % \item 
    
    \noindent
    (1) \textbf{General Face Tagging on Social and News Media. } Automatic face tagging such as Facebook's face recognition feature~\cite{FacebookFacialRecognition} can be fooled by DI attacks by creating a victim's deepfake that can be misclassified. Such attacks may imply malicious intents, leading to impersonation of certain target people and the creation of false information and fake personalities. For example, Eve is not Obama, but Eve can create a deepfake of Obama~\cite{ObamaDeepfake} and share this on social media, pretending to be Obama. Furthermore, she can claim some misinformation regarding election fraud, climate change~\cite{Belgianpremier} or show Obama visiting certain places during vacations. It can easily be combined with other fake news and extended to create fake digital footprints of a target person. In this scenario, if the commercial APIs fail to filter the deepfakes on social media, it will allow the propagation of false information and harm innocent individuals. Also, it is reported that deepfake-generated LinkedIn profiles are becoming prevalent, even resulting in serious misuse cases~\cite{link1, link2}. 
    %\simon{Add Linkedin Real case to make it more strong!} 
    % \item
    
   %\noindent
    %\sh{(2) \textit{Face Authentication}: In a Smart-home door lock system, which relies on commercial face-recognition such as Amazon Rekognition~\cite{DIYFaceID}, it can be fooled or spoofed using DI attack by presenting deepfake videos or images of the victim (target).} \simon{we may need to expand...maybe sowon has more materials on this.}
    
    % \item
    \noindent
    (2) \textbf{Face Authentication during Remote Meetings. }  Taking advantage of the common occurrences of remote meetings these days, one can use deepfake generation methods to impersonate an original participant. For example, students can hire someone to create deepfake images of themselves during remote exams to cheat. This has already been an issue, when the deepfake of Elon Musk bombed a Zoom call~\cite{ElonMuskZoomDF} combined with Zoombombing~\cite{zoombomb} to hijack the session. 
    %If an underlying face recognition APIs may not recognize the impersonator using the deepfake of the genuine user, then it can cause several security, privacy, and repudiation risks from various frauds.\sh{simon can you review this} 
    Assuming the underlying face recognition API cannot distinguish the deepfake impersonator from the genuine user, it can cause many privacy, security, and repudiation risks, as well as numerous fraud cases. Addressing these issues, especially these days where remote meetings have become common, has become important more than ever.
    
    % \item
    
    \noindent
    (3) \textbf{Deepfake Phishing. } Recently, banks have started to consider voice biometrics as a secure and convenient way of providing access to their services~\cite{BankAudioBiometrics}. However, as shown from the recent deepfake-based voice phishing incident in the UK~\cite{DFVoisePhishingUK}, audio deepfakes have become serious threats. Similarly, the concept of impersonation attacks can be easily extended to fooling commercial voice recognition APIs, such as Amazon Alexa, Apple Siri, and Google Assistant~\cite{DFAudio1,DFAudio2}, analogously as in the case for face recognition APIs. More importantly, voice and video deepfake technologies can be combined to create multi-modal deepfakes and used to carry out more powerful and realistic phishing attacks. In our work, we do not consider voice deepfakes, but the motivations are similar to those of face deepfake impersonation attacks.

% \end{enumerate}

%facing APIs are widely used for recognition and authentication.

%\hl{In addition to privacy,} we believe that these so-called deepfakes can be \hl{misused to carry out} an impersonation attack. \hl{And} this topic is not only of great significance \hl{from the security and privacy perspective} but also \hl{is of grave urgency to deal with further social misuses of deepfakes.} \simon{Please check this sentence w siho}
% \hl{fix this sentence}Previously, impersonation attacks on commercial face recognition APIs by adding adversarial noise in the reference (real) image using deep neural networks~\cite{dong2019efficient,zhong2020towards} or evolutionary algorithms~\cite{kim2018george} have been explored. 

%\noindent
%\textbf{Our Work. } 
Previously, adversarial attacks using deep neural networks~\cite{dong2019efficient,zhong2020towards} and evolutionary algorithms~\cite{kim2018george} have been used to generate attacks on commercial face recognition APIs. However, to the best of our knowledge, our work is the first to use deepfakes to implement the impersonation attack scenarios to deceive face recognition APIs. Furthermore, we are the first to formulate deepfake impersonation attacks for targeted and non-targeted scenarios, as described in Section~\ref{sec:Deepfake Impersonation Attack}. Through our work, we aim to raise awareness of the vulnerabilities and potential dangers regarding commercial APIs against various attack incidents.

%\textit{Note: DI attack is different from Adversarial attack; see Section~\ref{sec:discussion}}.
% \input{2_1_attacks}
% \input{2_2_datasets}

\section{Deepfake Attack Framework}
\label{sec:approach}
In this section, we provide an overview of the deepfake attack framework (Fig.~\ref{fig:Overview}) and threat models against celebrity recognition APIs. First, we provide a brief background on different categories of deepfake generation methods with representative benchmark datasets.
% We use these datasets to perform the deepfake-based attacks. 
Then, we discuss deepfake impersonation (DI) attacks using different deepfake datasets. We also explain how we conduct DI attacks using targeted vs. non-targeted case.

%Figure~\ref{fig:Overview} describes the design of each component. 

% \section{Overview and Threat Model}
% \label{sec:attacks}
% \begin{figure*}
%     \centering
%     \includegraphics[width=0.8\linewidth]{"figures/Adv_DF - Types".pdf}
%     \caption{types}
%     \label{fig:types}
% \end{figure*}

\subsection{Deepfake Generation Methods and Datasets}

\begin{figure}
    \centering
    \includegraphics[width=1\linewidth]{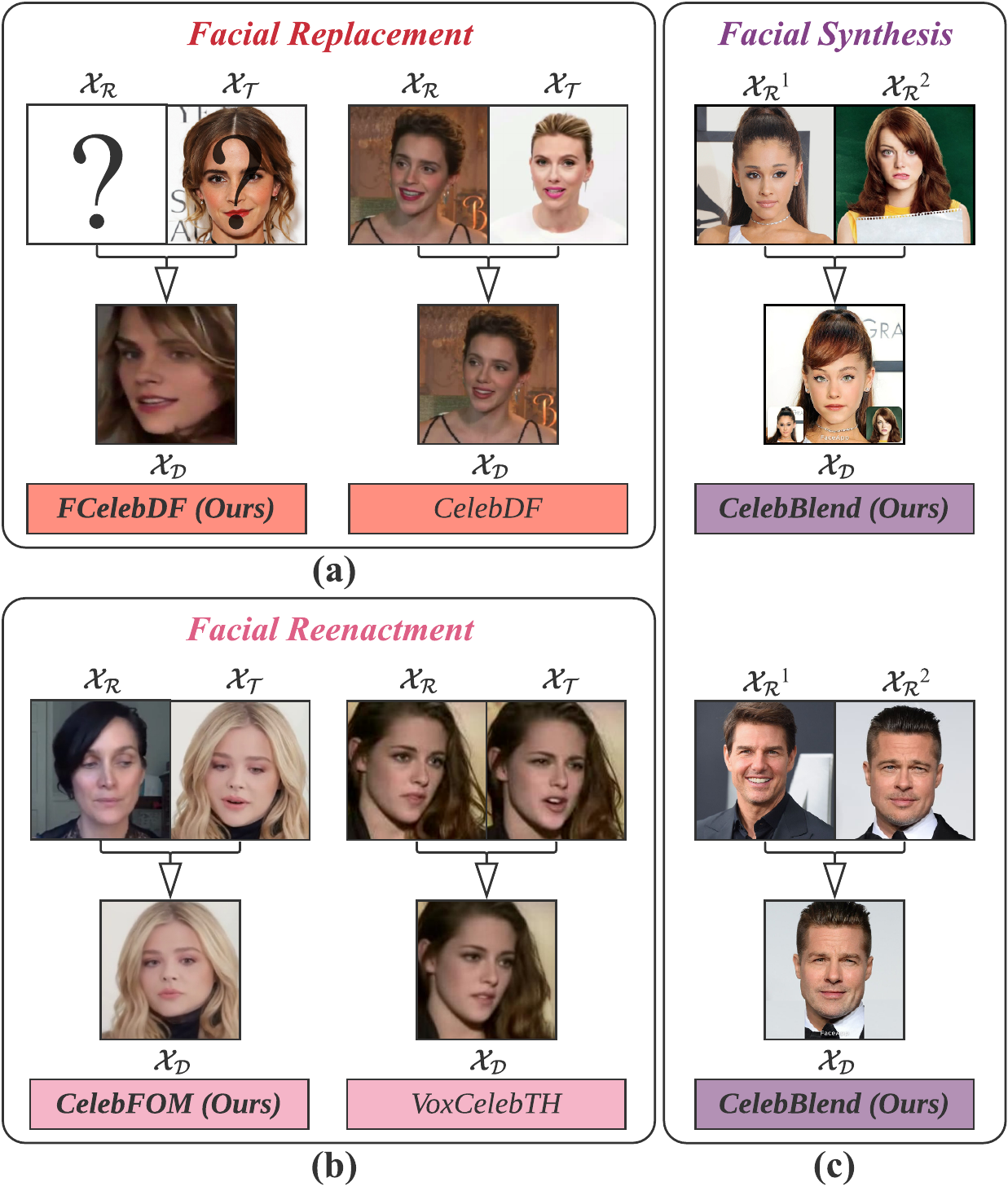}
    \caption{Three general categories of deepfake generation methods with example datasets. a) CelebDF and FCelebDF (\textit{Ours}) are made using Facial Replacement techniques, b) CelebFOM (\textit{Ours}) and VoxCelebTH are made with Facial Reenactment methods and c) CelebBlend (\textit{Ours}) is developed using Facial Synthesis approaches.}
    \label{fig:datasettypes}
    % \vspace{-10pt}
\end{figure}

%We considered three attack models, as shown in Fig~\ref{fig:attackstypes}, to test several commercial celebrity recognition APIs.
There are multiple types of deepfakes; however, most widely used deepfakes can be categorized into three types of deepfake generation methods~\cite{WenkeSurvey}. First, we introduce a brief overview of the most widely used deepfake datasets, where we use five different datasets as shown in Fig.~\ref{fig:datasettypes} to perform deepfake impersonation attack. Two of those datasets, the Celebrity Deepfake (\textit{CelebDF}) and VoxCeleb Talking Head (\textit{VoxCelebTH}) datasets, are publicly available. Furthermore, we collected the Female Celebrity Deepfake (\textit{FCelebDF}) dataset from the Internet, and we generated the Celebrity First Order Motion (\textit{CelebFOM}) and Celebrity Blend (\textit{CelebBlend}) datasets by ourselves. \textit{Note that the focus of this paper is on facial deepfakes; therefore, other types of deepfakes, such as human pose synthesis, are not considered in this work.}

In this paper, the reference (source) and target (victim) identities are denoted by $\mathcal{R}$ and $\mathcal{T}$, respectively; %where generally $\mathcal{R}$ is  a random video from the Internet, and $\mathcal{T}$ is the victim's video or images.
$\mathcal{X}_\mathcal{R}$ refers to a random video from the Internet and $\mathcal{X}_\mathcal{T}$ refers to images or videos of a target individual. The deepfakes generated from $\mathcal{X}_\mathcal{R}$ and $\mathcal{X}_\mathcal{T}$ are denoted by $\mathcal{X}_\mathcal{D}$. A sample deepfake image generated from a reference video featuring Emma Watson and a target video featuring Scarlett Johansson through facial replacement is shown in Fig.~\ref{fig:datasettypes}a--CelebDF.
% \sowon{where generally $\mathcal{X}_\mathcal{R}$ is a random video from the Internet, and $\mathcal{X}_\mathcal{T}$ is the victim/target's video or images. We also denote $\mathcal{X}_\mathcal{D}$ as the deepfake generated from $\mathcal{X}_\mathcal{R}$ and $\mathcal{X}_\mathcal{T}$ (e.g., a deepfake video of target Scarlett Johansson generated from Emma Watson's video by facial replacement technique, as shown in Fig.~\ref{fig:datasettypes}a--CelebDF).} 
% \simon{Reference is what and used for...and target identity is what...and used...}
%\simon{unclear what is target and reference. pls define more precisely using example. ok?} 
%We also denote $\mathcal{X}_\mathcal{R}$ and $\mathcal{X}_\mathcal{T}$ as the images/videos of these identities and $\mathcal{X}_\mathcal{D}$ as the deepfake generated from $\mathcal{R}$ and $\mathcal{T}$ (e.g., a deepfake video of target Scarlett Johansson generated from Emma Watson's video by facial replacement technique, as shown in Fig.~\ref{fig:datasettypes}a--CelebDF).  
A brief description of each deepfake generation method is provided as follows:

% \subsection{Attack Datasets Description}
% \label{sec:dataset}

% \subsubsection{Facial Replacement}
{\setlength{\parindent}{0cm}
\newtheorem{defi}{\textbf{Deepfake Gen. Method}}
% \AfterEndEnvironment{defi}{\noindent\ignorespaces}
\begin{defi}[\textbf{Facial Replacement}]
%$\mathcal{U}$ be the set of $x_s$ and $\mathcal{V}$ be the set of $x_t$, we
Let $\mathcal{X}_\mathcal{T}$ replace the content of $\mathcal{X}_\mathcal{R}$ to generate $\mathcal{X}_\mathcal{D}$ by using popular deepfake generation methods, commonly known as face swaps~\cite{Deepfakes,FaceSwap}, where $\mathcal{X}_\mathcal{D}$ contains the facial features of $\mathcal{T}$, body of  $\mathcal{R}$ and background settings of  $\mathcal{X}_\mathcal{R}$.
\end{defi}}
\noindent
\textbf{Example: }One famous application of this method is where the face of celebrity, such as Scarlett Johansson ($\mathcal{X}_\mathcal{T}$), is pasted on the face of another arbitrary person ($\mathcal{X}_\mathcal{R}$) in a random video from the Internet~\cite{ScarlettDeepfake}. In particular, the following are the two datasets that we use as facial replacement deepfakes to attack celebrity recognition APIs:

\begin{itemize}[leftmargin=10pt]
    \item \textbf{\textit{Celebrity Deepfake} (\textit{CelebDF}): } The CelebDF (Ver. 2) dataset is currently the largest publicly available celebrity deepfake dataset. It contains 5,639 face swap deepfakes video generated from 590 YouTube videos of 61 celebrities of different ages, ethnic groups, and genders speaking in different settings ranging from interviews to TV shows and award ceremonies. We use 58 unique identities in the downloaded dataset.
    \item \textbf{\textit{Female Celebrity Deepfake} (\textit{FCelebDF}): } Giorgio Patrini reported that 96\% of the deepfakes on the Internet are of females, and the majority of them belonged to female celebrities. Therefore, we crawled several online social media sites, explicitly searching for female celebrities' deepfakes. We use the search term ``Celebrity Name + Deepfake'' to find the videos' links and directly evaluate the face-cropped image from the deepfake videos without storing them locally. Since the $\mathcal{X_R}$ and $\mathcal{X_T}$ of this dataset are unknown, i.e., we do not have the real images for this dataset. Therefore, we download these celebrities' representative images from the Internet by querying their name on the search engine. We evaluate 5,000 face-cropped images extracted from numerous videos belonging to 50 female celebrities\footnote{We only stored the links to these videos for references and have not stored the actual video on our local machines due to ethical concerns. The whole evaluation process is done online by directly reading the video frames and sending face cropped images to web API. To prevent misuse and privacy violation, we do not plan to release this dataset.}. We also consulted with our Institutional Review Board (IRB) before evaluating these deepfakes to mitigate potential ethical, legal, and privacy issues. We explain the details of how we dealt with those issues in Section~\ref{sec:discussion}.

\end{itemize}

% \subsubsection{Facial Reenactment}
% \newtheorem{defi}{\textbf{Definition}}
{\setlength{\parindent}{0cm}
\begin{defi}[\textbf{Facial Reenactment}]
Let $\mathcal{X}_\mathcal{R}$ drive the head movement, facial expressions, and gaze of $\mathcal{X}_\mathcal{T}$ to generate $\mathcal{X}_\mathcal{D}$ with the help of different Facial Reenactment techniques such as Talking Head (TH)~\cite{VoxTalkingHead}, first-order motion model (FOM)~\cite{FOM}, Face2Face~\cite{Face2Face} and Neural Textures~\cite{NeuralTextures}, where $\mathcal{X}_\mathcal{D}$ contains the facial features of $\mathcal{T}$, background settings of $\mathcal{X}_\mathcal{T}$ and the expressions, mouth movement or gaze of $\mathcal{X}_\mathcal{R}$.
\end{defi}}
\noindent
\textbf{Example: } Typically, this method is used to animate famous politicians according to attacker's malicious will and intent such as Donald Trump's speech on climate change~\cite{TrumpClimateDeepfake} and Richard Nixon moon disaster speech~\cite{NixonDeepfake}. In Fig.~\ref{fig:datasettypes}, we show an example of facial reenactment where a real video of Carrie-Anne Moss ($\mathcal{X}_\mathcal{R}$) is used as a reference to drive the target image of Chloe Grace Moretz ($\mathcal{X}_\mathcal{T}$) in the CelebFOM dataset. We use the following two datasets for facial reenactment deepfakes:

\begin{itemize}[leftmargin=10pt]
\item \textbf{\textit{VoxCeleb Talking Head} (\textit{VoxCelebTH}): }
VoxCeleb1~\cite{voxceleb1} contains over 100,000 utterances for 1,251 celebrities, while VoxCeleb2~\cite{voxceleb2} contains over 1 million utterances for 6,112 celebrities extracted from videos uploaded to YouTube. Zakharov \textit{et al.}~\cite{VoxTalkingHead} present a system with the few-shot capability to learn highly realistic and personalized talking head models of people and even portrait paintings on a large dataset of videos. We combine two publicly available talking head datasets and formally name them as VoxCelebTH, each containing videos of 50 celebrities generated from VoxCeleb1~\cite{voxceleb1} and VoxCeleb2~\cite{voxceleb2} dataset using Zakharov \textit{et al.'s}~\cite{VoxTalkingHead} method.

\item \textbf{\textit{Celebrity First Order Motion} (\textit{CelebFOM}): }
In the first-order motion (FOM) method, an image animation consists of generating a video sequence so that an object in a reference image is animated according to the motion of a driving video~\cite{FOM}. %FOM decouples appearance and motion information using a self-supervised formulation.% To support complex motions, FOM uses a representation, consisting of a set of learned key points along with their local affine transformations.
We create this dataset by ourselves, generating 2,176 FOM videos from 544 reference video of 58 celebrities\footnote{We plan to release CelebFOM dataset with the paper.}.
\end{itemize}

% \subsubsection{Facial Synthesis}
{\setlength{\parindent}{0cm}
\begin{defi}[\textbf{Facial Synthesis}]
Let $\mathcal{R}^1$ and $\mathcal{R}^2$ be two reference identities, and $\mathcal{X}_\mathcal{D}$ is an image generated by blending $\mathcal{X}_\mathcal{R}^1$ and $\mathcal{X}_\mathcal{R}^2$ using FaceApp~\cite{FaceApp2018}, where $\mathcal{X}_\mathcal{R}^1$ and $\mathcal{X}_\mathcal{R}^2$ are facial images of two celebrities.
%The $x_s$ and $x_t$ are blended to synthesize $x_d$ in this attack.
The blending technique generates a new identity ($\mathcal{D}$) with a mixture of facial features from $\mathcal{R}^1$ and $\mathcal{R}^2$.
\end{defi}}
\noindent
\textbf{Example: } By blending the facial features of Brad Pitt ($\mathcal{X}_\mathcal{R}^1$) and Tom Cruise ($\mathcal{X}_\mathcal{R}^2$), the attacker can synthesize a celebrity lookalike identity ($\mathcal{X}_\mathcal{D}$)~\cite{BradPittFaceApp}.
\begin{itemize}[leftmargin=10pt] 
\item \textbf{\textit{Celebrity Blend} (\textit{CelebBlend}): } We created this dataset by ourselves, where two celebrity faces are blended to generate a synthetic identity. We use faces of 20 different celebrities, ten males and ten females from different age and ethnic groups to generate 180 synthetic but celebrity lookalike identities. We utilize the famous face manipulation application FaceApp~\cite{FaceApp2018} to generate these images\footnote{Final CelebBlend Dataset: We are currently developing more images for the CelebBlend dataset using FaceApp. We aim to release this dataset with images of at least 2,000 synthetic identities generated from over 50 celebrities belonging to different ethnic groups along with this paper.}.
\end{itemize}

%Nothing concrete can be said about the generation methods used to create these deepfake porn videos, and it is nearly impossible to find the original reference and target videos used to generate these deepfake porns (see Deepfake Porn in Fig.~\ref{fig:generation_types}). However, based on our visual analysis and experimental results, we believe that a variety of unknown deepfake generation methods, such as face swaps or methods available on online forums, may have been used. Therefore, it is possible that the celebrity Deepfake Porn dataset has been generated by a mix of different and/or novel methods (see Section~\ref{sec:res} for details).

% \subsection{Attack Design or Threat Model}

\subsection{Deepfake Impersonation (DI) Attack}
\label{sec:Deepfake Impersonation Attack}

% \subsubsection{\textbf{Motivation. }}
% \hl{rename it or put it somewhere else} 
Generally, a few photos of target individuals (victims) can be easily acquired through social media websites or social engineering. Furthermore, attackers can easily generate the corresponding deepfakes using the aforementioned methods, since necessary tools and codes for the generation of these deepfakes are freely available on many online forums ~\cite{Faceswapforum}. 
In addition, commercial face recognition web services are widely utilized for the authentication of a person's identity. %However, the robustness and vulnerability of these web APIs against deepfakes are still unknown. 
Therefore, we incorporate several deepfake datasets to perform impersonation attacks with the goal of deceiving the state-of-the-art commercial celebrity recognition web APIs. We present the overall DI attack framework in Fig.~\ref{fig:Overview}. In the following sections, we refer to deepfake impersonation attacks for targeted and non-targeted attack scenarios as $f(\cdot)$, representing a general deepfake generation method.
% For example, with the Facial Synthesis method, the attacker's objective is to generate a synthetic celebrity look-alike identity by blending $\mathcal{X}_\mathcal{R}^1$ and $\mathcal{X}_\mathcal{R}^2$ to deceive a face recognition system. 

\subsubsection{\textbf{Targeted DI Attack Definition.}} Given a reference image $\mathcal{X}_\mathcal{R}$, a target image $\mathcal{X}_\mathcal{T}$, and a deepfake generation method $f(\cdot)$, it is possible to construct a deepfake image $f(\mathcal{X}_\mathcal{R},\mathcal{X}_\mathcal{T})$ = $\mathcal{X}_\mathcal{D}$, such that $C(\mathcal{X}_\mathcal{D})=C(\mathcal{X}_\mathcal{T})$, with $\mathcal{X}_\mathcal{D}$ close to %$\mathcal{X}_\mathcal{R}$ and 
$\mathcal{X}_\mathcal{T}$ 
over some distance metric $p$, as follows:% \simon{double check the equation again}
\begin{equation}
\label{eq:TAD}
\small
    \begin{split}
%\min_{\mathcal{X}_\mathcal{D}} 
\min%_{\theta} 
\mathcal{L}_p(\mathcal{X}_\mathcal{R},\mathcal{X}_\mathcal{T}, f_{\theta}(\mathcal{X}_\mathcal{R},\mathcal{X}_\mathcal{T}), \theta),
\\s.t.\quad C(\mathcal{X}_\mathcal{D})=C(\mathcal{X}_\mathcal{T}),
\end{split}
\end{equation}
where $\theta \in \mathbb{R}^{m}$ is a set of parameters $m$ of $f(\cdot)$ given a deepfake generation method, $C(\cdot)$ is the class label function, and $\mathcal{L}$ is some loss function to be minimized, such that $\mathcal{X}_\mathcal{D}$ becomes similar to $\mathcal{X}_\mathcal{T}$, hence the impersonation.

\subsubsection{\textbf{Non-targeted DI Attack Definition.}} Given $\mathcal{X}_\mathcal{R}$, $\mathcal{X}_\mathcal{T}$, and $f(\cdot)$, it is possible to construct a deepfake image $f(\mathcal{X}_\mathcal{R},\mathcal{X}_\mathcal{T})$ = $\mathcal{X}_\mathcal{D}$, as follows:
% \simon{double check the equation again}
\begin{equation}
\label{eq:NAD}
\small
    \begin{split}
%\min_{\mathcal{X}_\mathcal{D}} 
\min%_{\theta} 
\mathcal{L}_p(\mathcal{X}_\mathcal{R},\mathcal{X}_\mathcal{T}, f_{\theta}(\mathcal{X}_\mathcal{R},\mathcal{X}_\mathcal{T}), \theta),
\\s.t.\quad C(\mathcal{X}_\mathcal{D})\in \mathbb{C},\quad\quad\quad\hfill
\end{split}
\end{equation}
where $\mathbb{C}$ is the set of all class labels (i.e., all the valid celebrity classes), and $\mathcal{L}$ is a loss function to be minimized, such that $\mathcal{X}_\mathcal{D}$ is similar to a celebrity label, hence the impersonation.

\begin{comment}
\begin{equation}
\label{eq:NAD}
\small
    \begin{split}
\min_{\theta}
\mathcal{L}(\mathcal{X}_\mathcal{R},\mathcal{X}_\mathcal{T}, \theta),
\\s.t.\quad C(\mathcal{X}_\mathcal{D})\in \mathbb{C},
\end{split}
\end{equation}
\end{comment}

\begin{comment}
\begin{equation}
\label{eq:NAD}
\small
    \begin{split}
\min_{\mathcal{X}_\mathcal{D}} \mathcal{L}(\mathcal{X}_\mathcal{R},\mathcal{X}_\mathcal{T}, \theta),
\\s.t.\quad C(\mathcal{X}_\mathcal{D})\in \mathbb{C},
\end{split}
\end{equation}
where $\mathbb{C}$ is the set of all class labels. \simon{what is DI here?}
\end{comment}

According to Eq.~\eqref{eq:NAD}, non-targeted attacks can follow one of three scenarios. Suppose $C(\mathcal{X}_\mathcal{D}) = C(\mathcal{X}_\mathcal{T})$, then it is the same as the targeted attack, whereas $C(\mathcal{X}_\mathcal{D}) = C(\mathcal{X}_\mathcal{R})$ means $\mathcal{X}_\mathcal{D}$ has the same celebrity class as the reference, $\mathcal{X}_\mathcal{R}$ (see VoxCelebTH in Fig.~\ref{fig:datasettypes}b). % \st{However, either in a different video or reanimated in the same video, as the attacker desires??}.
Lastly, if $\{C(\mathcal{X}_\mathcal{D}) \in \mathbb{C} \mid  C(\mathcal{X}_\mathcal{D}) \ne C(\mathcal{X}_\mathcal{R}) \text{ and } C(\mathcal{X}_\mathcal{D}) \ne C(\mathcal{X}_\mathcal{T})\}$, then the predicted class belongs to neither the reference nor the target, but is a valid class corresponding to one of other celebrities.

\noindent
\subsubsection{\textbf{DI Attack vs. Adversarial Example Attack.}} While DI and adversarial attacks~\cite{adv} appear to be similar, they are different. The main goal of the former is to increase \textit{false negatives} by misclassifying, `Real Obama' ($\mathcal{X}_\mathcal{T}$), for instance, as someone else ($\mathcal{X}'$), denoted by $C$($\mathcal{X}_\mathcal{T}$) $\to$ $C$($\mathcal{X}'$). In contrast, the goal of the latter is to increase \textit{false positives} by misclassifying `Fake Obama' ($\mathcal{X}_\mathcal{D}$) as `Real Obama' ($\mathcal{X}_\mathcal{T}$), denoted by $C$($\mathcal{X}_\mathcal{D}$) $\to$ $C$($\mathcal{X}_\mathcal{T}$), %where  is created with an images of A and a target video ($\mathcal{X}_\mathcal{D}$),
where $f(\mathcal{X}_\mathcal{R},\mathcal{X}_\mathcal{T})$ = $\mathcal{X}_\mathcal{D}$.
%It is important to note that $A^*=A \bigotimes X$ and $A^*!=A$,  where $\bigotimes$ is the abstraction of some deepfake generation method. 
Therefore, DI and adversarial attacks are inherently different and can be used to exploit two distinct aspects of attacks (false positive vs. false negative). In this work, we introduce a DI attack to generate false positive scenarios, which have not been examined much in the past.

\subsubsection{\textbf{DI Attack Dataset Generation}}
In order to perform DI attacks, we use five different datasets from three-generation methods, as described earlier: 1) for generating the facial replacement deepfakes, we consider the most general scenario, in which the attacker collects $\mathcal{X}_\mathcal{T}$ from different viewing angles and $\mathcal{X}_\mathcal{R}$ from a reference video. Then, an attacker applies a DNN-based face swap method~\cite{FaceSwap,Deepfakes} to generate a deepfake video $\mathcal{X}_\mathcal{D}$ of the victim. 2) Also, the attacker can use a reference video $\mathcal{X}_\mathcal{R}$, which contains the desired actions, and a set of images $\mathcal{X}_\mathcal{T}$ of the victim. Then, the attacker can apply a Facial Reenactment method, such as Talking Head~\cite{VoxTalkingHead} or First-order Motion~\cite{FOM} to generate $\mathcal{X}_\mathcal{D}$. 3) For Facial Synthesis deepfakes, the attacker can select two celebrity facial images as $\mathcal{X}_\mathcal{R}^1$ and $\mathcal{X}_\mathcal{R}^2$ and uses the image blend functionality of the FaceApp~\cite{FaceApp2018} Android application to generate $\mathcal{X}_\mathcal{D}$. 

To determine how each deepfake generation method performs on various state-of-the-art face recognition systems, we use the CelebDF~\cite{celebdf} and FCelebDF (\textit{Ours}) datasets to emulate Facial Replacement, the VoxCelebTH~\cite{VoxTalkingHead} and CelebFOM (\textit{Ours}) datasets for Facial Reenactment, and the CelebBlend (\textit{Ours}) dataset for Facial Synthesis-based deepfake attacks. %\simon{clean up this paragraph...tangled...better org so that reader can read easily, also repeating from previous...} 
%Section~\ref{sec:dataset} provides the details of these datasets.

\subsubsection{\textbf{Conducting Deepfake Impersonation Attack}}

% We simulate the deepfake generation methods defined in our threat model using the datasets mentioned in Section~\ref{sec:dataset}.
As a first step, we crop the face from an image in a deepfake video using the MTCNN~\cite{MTCNN} method. This cropped face is uploaded to commercial celebrity recognition web APIs of Amazon, Microsoft, and Naver, as shown in Fig.~\ref{fig:Overview}. After uploading, we make two types of requests to the web API:
 \begin{enumerate}[leftmargin=15pt,label=(\roman*)]
     \item \textbf{\textit{Celebrity Recognition} (\textit{CR}): } The web APIs try to identify the uploaded deepfake face image $\mathcal{X_D}$ with their celebrity databases. If a match is found, it returns $\mathcal{P}$ = [$\mathcal{P}^\mathtt{name}$, $\mathcal{P}^\mathtt{conf}$], which is an output containing the celebrity name $\mathcal{P}^\mathtt{name}$ and its match confidence $\mathcal{P}^\mathtt{conf}$. We refer to this web API function as \textit{RecognizeCelebrity}$(\mathcal{X})$ = $\mathcal{P}$.
     \item \textbf{\textit{Face Similarity} (\textit{FS}): } We also use the FS score, since face similarity is used for a wide variety of user verification tasks, people counting, and public safety use cases~\cite{AmazonUsecase}.  
     We upload the real photo of the celebrity $\mathcal{X_T}$ alongside the deepfake photo $\mathcal{X_D}$, and the web API checks the similarity between the two face images. We refer to this web API function as \textit{Similarity}$(\mathcal{X_T},\mathcal{X_D})$. %This can evaluate how closely a generated deepfake is similar to a real image of a celebrity.  
 \end{enumerate}

We use the confidence score to evaluate how certain the API is regarding its prediction. In contrast, we use the similarity score to measure the resemblance between the deepfake and the real image corresponding to a celebrity.

\subsubsection{\textbf{Deepfake Impersonation Attack Scenarios}}

Consider $\mathcal{P}_\mathcal{T}$ and $\mathcal{P}_\mathcal{D}$ are the recognition results from the API for target $\mathcal{X_T}$ and deepfake $\mathcal{X_D}$, where $\mathcal{P}^\mathtt{conf}_\mathcal{T}$ and $\mathcal{P}^\mathtt{conf}_\mathcal{D}$ denote the confidence of the predictions, and the name of the predicted celebrity is denoted by $\mathcal{P}^\mathtt{name}_\mathcal{T}$ and $\mathcal{P}^\mathtt{name}_\mathcal{D}$, respectively. Note that there is no target $\mathcal{X_T}$ in the Facial Synthesis case. Hence, we use two references $\mathcal{X_R}^1$ and $\mathcal{X_R}^2$ instead. 
Next, we perform two types of impersonation attacks, targeted and non-targeted attacks as follows:
\begin{enumerate}[leftmargin=15pt,label=(\roman*)]
    \item \textbf{\textit{Success Definition of Targeted Attack} (\textit{TA}): } Targeted attack determines whether the deepfake $\mathcal{X_D}$ and target $\mathcal{X_T}$ are recognized as the same celebrity by the web API, i.e., $\mathcal{P}^\mathtt{name}_\mathcal{T}$ = $\mathcal{P}^\mathtt{name}_\mathcal{D}$.
    %This attack determines
    %\simon{whether the deepfake of a celebrity is recognized as themselves by the web API.} \simon{make it easier } 
    For example, a deepfake of Emma Watson is recognized as Emma Watson. Therefore, the following two conditions need to be satisfied to perform this attack, as shown in Eq.~\eqref{eq:TA}: %First, the API detects the deepfake as a celebrity, and second, the predictions ($\mathcal{P}$)  for the identity of $\mathcal{T}$ in $\mathcal{X_T}$ and Deepfake in $\mathcal{X_D}$ is the same, as shown in Eq.~\eqref{eq:TA}. 
    \begin{equation}
        \small
        TA=\begin{cases}
            Successful&\begin{matrix}
            \text{ if } \mathcal{P}_\mathcal{D}= Celebrity^\mathtt{Target}\hfill\\ 
            \text{ and } \mathcal{P}^\mathtt{name}_\mathcal{T}=\mathcal{P}^\mathtt{name}_\mathcal{D}\hfill
            \end{matrix}\\
            Fail & \text{ otherwise }
        \end{cases},
        \label{eq:TA}
    \end{equation}
    %where $\mathcal{P}^\mathcal{D}$ is the prediction result from web API function for recognizing the celebrity $RecognizeCelebrity (\mathcal{X_D})$ and $\mathcal{P}^\mathcal{T}_{id}$ and $\mathcal{P}^\mathcal{D}_{id}$ represents the identity of the celebrity for Target ($\mathcal{T}$) and Deepfake ($\mathcal{D}$), respectively \simon{this should be better to be defined eariler, not here.}. 
\item \textbf{\textit{Success Definition of Non-targeted Attack} (\textit{NA}): }
%\simon{Exaplain the goal of this IE attack. Also, give example or scenario using a celebrity}
This attack is very similar to TA; however, it covers a more general attack scenario, where it checks if the deepfake $\mathcal{X_D}$ of a celebrity can be recognized as any celebrity by the web API. For example, a deepfake of Emma Watson is recognized as Emma Watson or any other celebrity. The goal of this attack is to enable any misclassification. Therefore, only one condition from Eq.~\eqref{eq:TA} is required to perform this attack, as follows: %shown in Eq.~\eqref{eq:NA}.
\begin{equation}
\small
NA=\begin{cases}
Successful& \text{ if } \mathcal{P}_\mathcal{D}= Celebrity^\mathtt{Any}\hfill\\ 
Fail & \text{ otherwise }
\end{cases}
\label{eq:NA}
\end{equation}
\end{enumerate}

\subsubsection{\textbf{DI Attack Evaluation Metrics}}
While targeted and non-targeted attack success rates are useful, we aim to examine further the fine-grained details of differing robustness behaviors of each API. Therefore, we introduce four additional evaluation metrics to more carefully examine and analyze the effects of DI attacks on celebrity recognition APIs from different perspectives.
\begin{figure*}[t]
    \centering
    \includegraphics[width=0.99\linewidth]{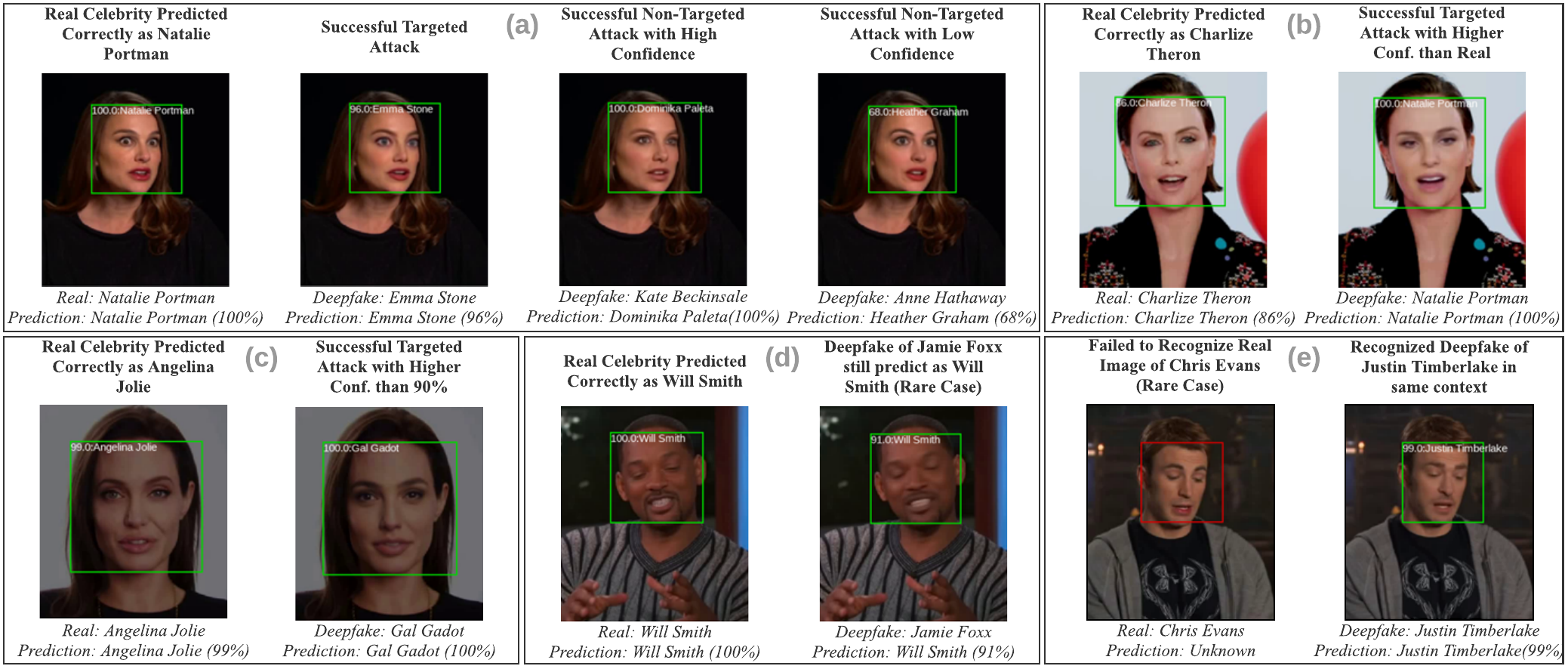}
    \caption{Demonstration of few scenarios of deepfake impersonation attacks and the results from commercial celebrity recognition web APIs such as (a) the successful TA and NA attacks, (b) DHF, (c) DHC, (d) a rare case, where $\mathcal{X_D}$ is predicted as $\mathcal{X_S}$ instead of $\mathcal{X_T}$, and (e) also a rare case, where deepfake is recognized but the real celebrity is not.}
    \label{fig:Attack_Demonstration}
    % \vspace{-10pt}
\end{figure*}
\begin{enumerate}[leftmargin=15pt,label=(\roman*)]
    \item \textbf{\textit{Deepfakes with High Fidelity} (\textit{DHF}):}
    We use DHF evaluation metric to determine how realistic or lifelike the deepfake $\mathcal{X_D}$ of a celebrity is, compared to real celebrity $\mathcal{X_T}$.
    For example, we observe that some deepfakes of Emma Watson are very realistic, i.e., it is tough to distinguish between her real and deepfake images.
    Then, we hypothesize that the API can predict $\mathcal{X_D}$ as real and sometimes provides $\mathcal{X_D}$ an even higher confidence score than the real image $\mathcal{X_T}$, demonstrating the apparent vulnerability of APIs. Therefore, we examine if the API can predict the deepfake $\mathcal{X_D}$ with higher confidence than the real image of the same celebrity $\mathcal{T}$ as follows:
    \begin{equation}
    \small
         DHF=\begin{cases}
            True &\begin{matrix}
            \text{ if } \mathcal{P}_\mathcal{D}= Celebrity^\mathtt{Any}\hfill\\ 
            \text{ and } \mathcal{P}^\mathtt{conf}_\mathcal{D}>\mathcal{P}^\mathtt{conf}_\mathcal{T}\hfill
            \end{matrix}\\
            False & \text{ otherwise }
        \end{cases}
    \label{eq:LD}
    \end{equation}

    \item \textbf{\textit{Deepfakes with High Confidence} (\textit{DHC}): }
    When the API recognizes a deepfake $\mathcal{X_D}$ as a celebrity, this means the attack is successful.
    Also, it is essential to analyze the confidence level the API returns for its prediction $\mathcal{P}^\mathtt{conf}_\mathcal{D}$, when the attack is successful, measuring the certainty about deepfake prediction. The higher the confidence level is, the more the API is vulnerable against deepfakes.
    Therefore, we calculate the number of deepfakes $\mathcal{X_D}$ predicted as a celebrity with high confidence, i.e., greater than $\beta$, as follows:
    \begin{equation}
    \small
    DHC=\begin{cases}
    True &\begin{matrix}
    \text{ if } \mathcal{P}_\mathcal{D}= Celebrity^\mathtt{Any}\hfill\\ 
    \text{ and } \mathcal{P}^\mathtt{conf}_\mathcal{D}>\beta\hfill 
    \end{matrix}\\
    False & \text{ otherwise }
    \end{cases}
    \label{eq:DHC}
    \end{equation}
    
    \item \textbf{\textit{Deepfakes with High Similarity} (\textit{DHS}): }
    There are cases when the API does not recognize the deepfake $\mathcal{X_D}$ of a celebrity, i.e., Celebrity Recognition (CR) fails and returns no prediction for the image. In such a scenario, we perform the second step, Face Similarity (FS) evaluation. The face similarity can provide useful information, such as the percentage of resemblance between deepfake $\mathcal{X_D}$ and real image $\mathcal{X_T}$. If the resemblance is high (>$\gamma$), it means the API's face similarity function is more susceptible to deepfakes. Therefore, we can consider this similarity score as another indicator of API robustness against deepfakes.
    %There are cases when the deepfake $\mathcal{X_D}$ of a celebrity is not recognized by the API, i.e., recognition fails}. In such a scenario, we perform the second step of similarity evaluation which can provide useful information... by using the API's similarity function to check the resemblance between the target celebrity's face $\mathcal{X_T}$ and the celebrity's deepfake $\mathcal{X_D}$ using the similarity function from the web API. 
    %\simon{add the object and explain why this metric is useful}
    %This also helps to evaluate the API's robustness and how APIs recognize the DF and real image similarly.
     If the similarity between $\mathcal{X_T}$ and $\mathcal{X_D}$ is more than the certain threshold $\gamma$, we classify the deepfake being successfully similarly recognized as a celebrity as follows:
    \begin{equation}
    \small
    DHS=\begin{cases}
    True &\begin{matrix}
    \text{ if } \mathcal{P}_\mathcal{D}\neq Celebrity^\mathtt{Any}\hfill\\ 
    \text{ and } Similarity(\mathcal{X_T},\mathcal{X_D})>\gamma\hfill 
    \end{matrix}\\
    False & \text{ otherwise }
    \end{cases}
    \label{eq:DHS}
    \end{equation}
    \item \textbf{\textit{Successful Impersonation of Celebrities} (\textbf{SIC}): } 
    We can generate multiple deepfake images for the same target celebrity. We aim to measure whether any of these deepfakes can successfully deceive the API, demonstrating the existence of the deepfakes on any given target celebrity. Thus, this evaluation metric shows the existence of deepfakes, and it can be used for successful impersonation.
    %To examine the impact of our attack on each dataset, we count the total number of successful celebrity impersonations in attack datasets. For example, if there are ten unique celebrities in a dataset and the API recognizes deepfakes of 7 of those celebrities, then the SIC will be 70\%. 
    %This metric can reveal if the classification of deepfake $\mathcal{X_D}$ as a real image $\mathcal{X_T}$ by the API is limited to only a few celebrities or applies to all celebrities. \simon{unclear}
    On the other hand, the TA and NA attacks are different from SIC such that they are focused on the total number of deepfake images $\mathcal{X_D}$ recognized as real celebrities $\mathcal{X_T}$. In contrast, SIC focuses on the total count of unique celebrities, where any of their deepfakes are successfully detected as real. 
    Let us consider that $\mathcal{N}$ is the total number of deepfakes of a particular celebrity in the attack dataset. If the web API predicts at least one of that celebrity's deepfake as real, we consider it a success, as shown in Eq.~\eqref{eq:SIC}:
    \begin{equation}
    \small
    SIC=\begin{cases}
    True &\begin{matrix}
    \text{ if } \exists~\mathcal{X_D} \mid \mathcal{X_D}\in \mathcal{N}\hfill\\ 
    \text{ and } \mathcal{P}^\mathtt{name}_\mathcal{T}=\mathcal{P}^\mathtt{name}_\mathcal{D}\hfill\\
    \text{ and } \mathcal{P}_\mathcal{D}= Celebrity^\mathtt{Target}\hfill\\
    \end{matrix}\\
    False & \text{ otherwise }
    \end{cases}
    \label{eq:SIC}
    \end{equation}
\end{enumerate}
Note: For evaluating the above DHH, DHC, DHS, and SIC, we measure the percentage (\%) of images in each dataset that successfully satisfy these metrics' criteria, as shown in Fig.~\ref{fig:LD} to~\ref{fig:SIC}.

\begin{table}[t]
\centering
\caption{Dataset Description and Racial Distribution.}
\label{tab:DATASET}
\resizebox{\linewidth}{!}{%
\begin{tabular}{l|c|c|r} 
\toprule
 \textbf{Dataset}  & \textbf{Celebrity}  & \multicolumn{1}{c|}{\textbf{Videos} } & \multicolumn{1}{c}{\textbf{Images Tested (Real, Fake)} } \\ 
\hline
CelebDF & 58 & 5,643 & 6,184 (Real:\ \ \  546,\ \  Fake:\ \ \  5,638) \\
\textbf{FCelebDF (\textit{Ours})}  & 50 & 200 & 5,050 (Real:\ \ \ \ \ 50,\ \  Fake:\ \ \  5,000) \\
VoxCelebTH & 100 & 100 & 3,300 (Real:\ \ \  100,\ \  Fake:\ \ \  3,200) \\
\textbf{CelebFOM (\textit{Ours})}  & 58 & 2,176 & 2,720 (Real:\ \ \  544,\ \  Fake:\ \ \  2,176) \\
\textbf{CelebBlend (\textit{Ours})}  & 20 & \multicolumn{1}{c|}{-} & \ \ \ 200 (Real:\ \ \ \ \ 20,\ \  Fake:\ \ \ \ \ \ 180) \\ 
\hline
Total & 286 & 8,119 & 17,454 (Real: 1,260,\ \  Fake: 16,239) \\ 
\hline
Racial Distribution & \multicolumn{3}{c}{\begin{tabular}[c]{@{}c@{}}White (78.7\%), Asian (7.4\%), Black (4.6\%), Hispanic (3.7\%),\\Asian/Indian (2.8\%), Multiracial (2.8\%)\end{tabular}} \\
\bottomrule
\end{tabular}
}
% \vspace{-10pt}
\end{table}

% \usepackage{graphicx}
% \usepackage{booktabs}

% \begin{table}
% \centering
% \caption{Dataset Description and Racial Distribution.}%. Except CelebBlend dataset all dataset are in video format.}
% \label{tab:DATASET}
% \resizebox{\linewidth}{!}{%
% \begin{tabular}{l|c|c|r} 
% \toprule
%  \textbf{Dataset}  & \textbf{Celebrity}  & \textbf{Videos}  & \multicolumn{1}{c}{\textbf{Images Tested (Real, Fake)} } \\ 
% \hline
% CelebDF & 58 & 5,643 & 6,184 (Real:    546,   Fake:    5,638) \\
% \textbf{FCelebDF (\textit{Ours})}  & 50 & 200 & 5,050 (Real:     0,   Fake:    5,000) \\
% VoxCelebTH & 100 & 100 & 3,300 (Real:    100,   Fake:    3,200) \\
% \textbf{CelebFOM (\textit{Ours})}  & 58 & 2,176 & 2,720 (Real:    544,   Fake:    2,176) \\
% \textbf{CelebBlend (\textit{Ours})}  & 20 & - & 200 (Real:     20,   Fake:      180) \\ 
% \hline
% Total & 286 & 8,119 & 17,454 (Real: 1,260,   Fake: 16,239) \\ 
% \hline
% Racial Distribution & \multicolumn{3}{c}{\begin{tabular}[c]{@{}c@{}}White (78.7\%), Asian (7.4\%), Black (4.6\%), Hispanic (3.7\%),\\Asian/Indian (2.8\%), Multiracial (2.8\%)\end{tabular}} \\
% \bottomrule
% \end{tabular}
% }
% \vspace{-10pt}
% \end{table}

\section{Experimental Results}

The goal of our experiment is to measure the robustness of commercial face recognition web APIs against DI attacks. We apply the attacks to celebrity recognition and face similarity methods of the Amazon Rekognition API, Azure Cognitive Service, Naver Clova Face Recognition using five datasets. %This section explains the experimental setup, the web APIs' details, and how we analyzed the attacks using four attack analysis metrics.
% We applied the attacks to celebrity recognition and face comparison of the Amazon Rekognition  API, Azure Cognitive Service, Naver Clova Face Recognition for six datasets. In this section, we explain the experimental setup including the datasets we used, and the details of web APIs we evaluated. After then, we describe how we analyzed the attack based on four attack analysis metrics.

%\subsection{Experimental Setup}
%\label{sec:experiment}
%Our experiment used five different datasets generated with three types of deepfake generation methods, as explained above. Considering the real-world deepfake videos, we consisted of the datasets with various lines of celebrities, including Hollywood movie stars, singers, sports players, and politicians. We explain the numbers of datasets in detail. Using these datasets, we evaluated the web APIs. Since these APIs are for commercial, we had to pay for the experiments under each web APIs' payment policy as follows.

\subsection{\textbf{Deepfake Impersonation Attack Dataset}}

We use five different datasets generated from three types of deepfake generation methods described in Section~\ref{sec:approach} for our experiments. These datasets consist of real-world deepfake videos of celebrities, including Hollywood movie stars, singers, sports players, and politicians. %Using these datasets, we evaluated the \hl{web} APIs, and since they are commercial APIs, %\st{we had to pay for the experiments} \hl{costing less than 100? dollar in total for testing more than X number of deepfake images.} 

%In Table~\ref{tab:DATASET}, we describe the numbers of celebrities, videos, images, and images tested for each dataset. We extracted the frames from the video and cropped out the faces. For CelebDF, we tested 5,638 fake images extracted from 5,643 videos of 58 celebrities. We have 50 celebrities, a total of 5,000 fake images from 200 videos in FCelebDF. For VoxCelebTH, we tested 3,200 fake images of 100 celebrities. For CelebFOM, we extracted and used 2,176 fake images from 2,176 videos of 58 celebrities, and we tested 180 fake images of 20 celebrities for CelebBlend. We also evaluate the same celebrities' real images to calculate the similarity score and other evaluation metrics, as shown in Table~\ref{tab:DATASET}. %Each celebrity has a different number of images.

In Table~\ref{tab:DATASET}, we describe the numbers of celebrities, videos, images, and images tested for each dataset. We extract the frames from the video and cropped out the faces. For CelebDF, we test 5,638 fake images extracted from 5,643 videos of 58 celebrities. We have 50 celebrities, a total of 5,000 fake images from 200 videos in FCelebDF. For VoxCelebTH, we test 3,200 fake images of 100 celebrities. For CelebFOM, we extract and used 2,176 fake images from 2,176 videos of 58 celebrities. We test 180 fake images of 20 celebrities for CelebBlend. We also evaluate the same celebrities' real images to calculate the similarity score and other evaluation metrics, as shown in Table~\ref{tab:DATASET}. Overall, we use 8,119 deepfake videos for our evaluation, where these datasets' racial distribution is 86.01\% White, 8.04\% Asian, 5.94\% Black, 3.7\% Hispanic, 2.8\% Asian/Indian, and 2.8\% Multiracial by manual inspection. %(See Appendix~\ref{Appendix: Racial Distribution of Datasets and Prediciton from API} for more details).
We will discuss the performance of each API with different race groups (see Section~\ref{sec:discussion} and Appendix~\ref{Appendix: Racial Distribution of Datasets and Prediciton from API}). %\simon{Why we show this? Bias? what does it have to do with results?}

\begin{figure*}[t]
    \begin{subfigure}{0.32\linewidth}
        \centering
        \includegraphics[clip, trim=0pt 0pt 0pt 0pt, width=1\linewidth] {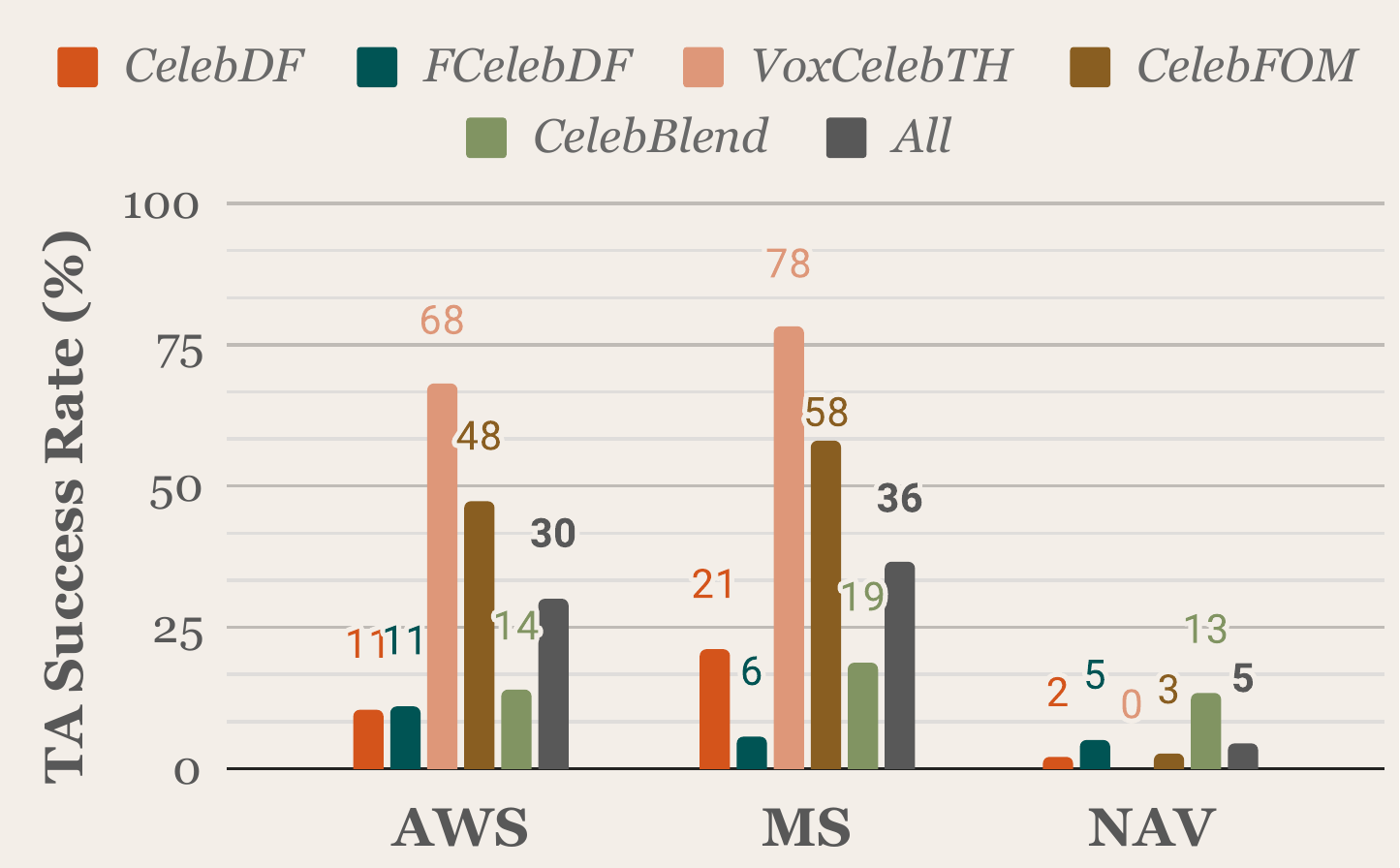}
        \caption{Targeted attack success rate i.e., when $C(\mathcal{X}_\mathcal{D})=C(\mathcal{X}_\mathcal{T})$}
        \label{fig:STA}
    \end{subfigure}\hfill
    % \par\bigskip
    \begin{subfigure}{0.32\linewidth}
        \centering
        \includegraphics[clip, trim=0pt 0pt 0pt 0pt,width=1\linewidth]{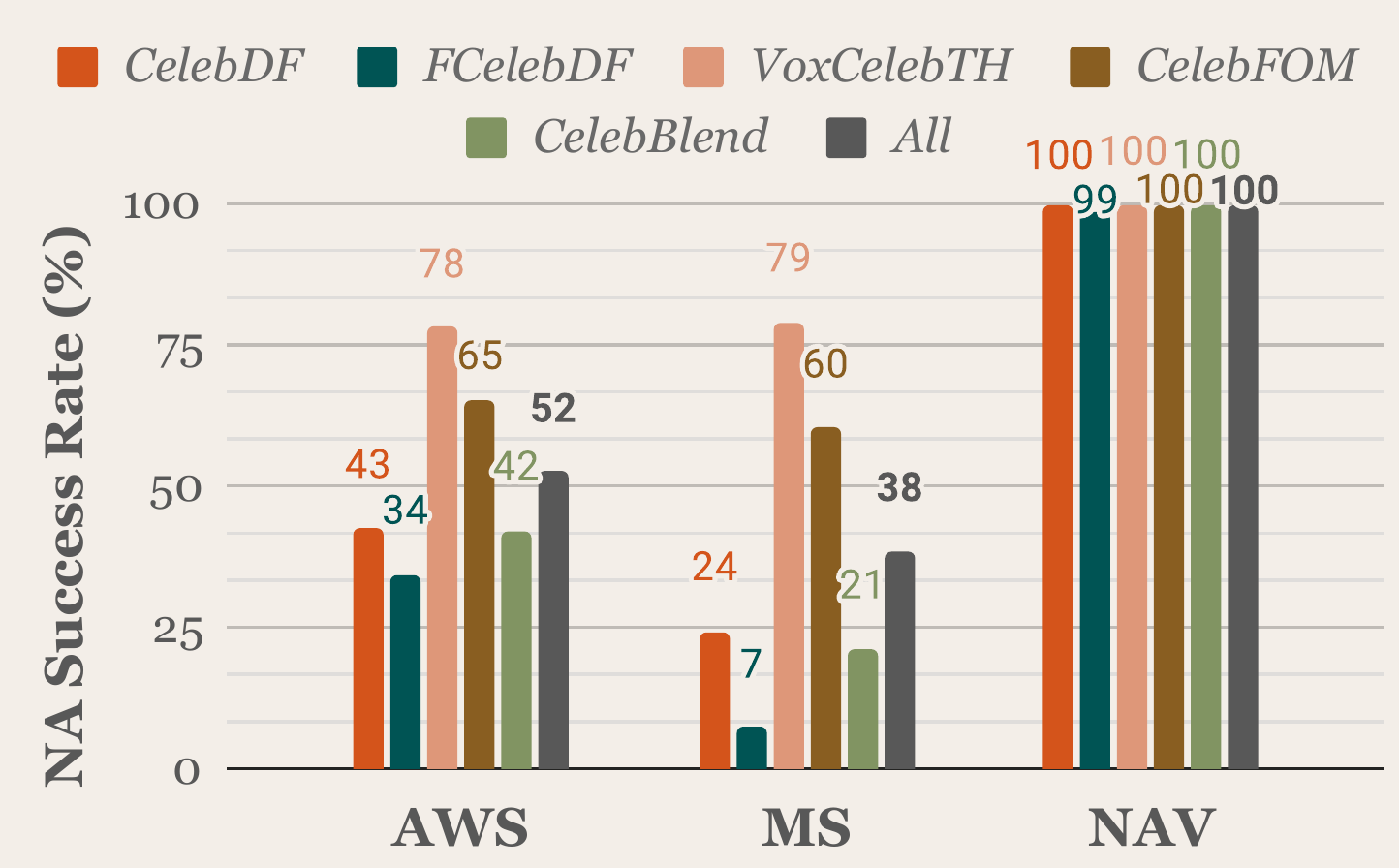}
        \caption{Non-targeted attack success rate i.e., when $C(\mathcal{X}_\mathcal{D})\in \mathbb{C}$}
        \label{fig:SNA}
    \end{subfigure}\hfill
    \begin{subfigure}{0.32\linewidth}
        \centering
        \includegraphics[clip, trim=0pt 0pt 0pt 0pt,width=1\linewidth]{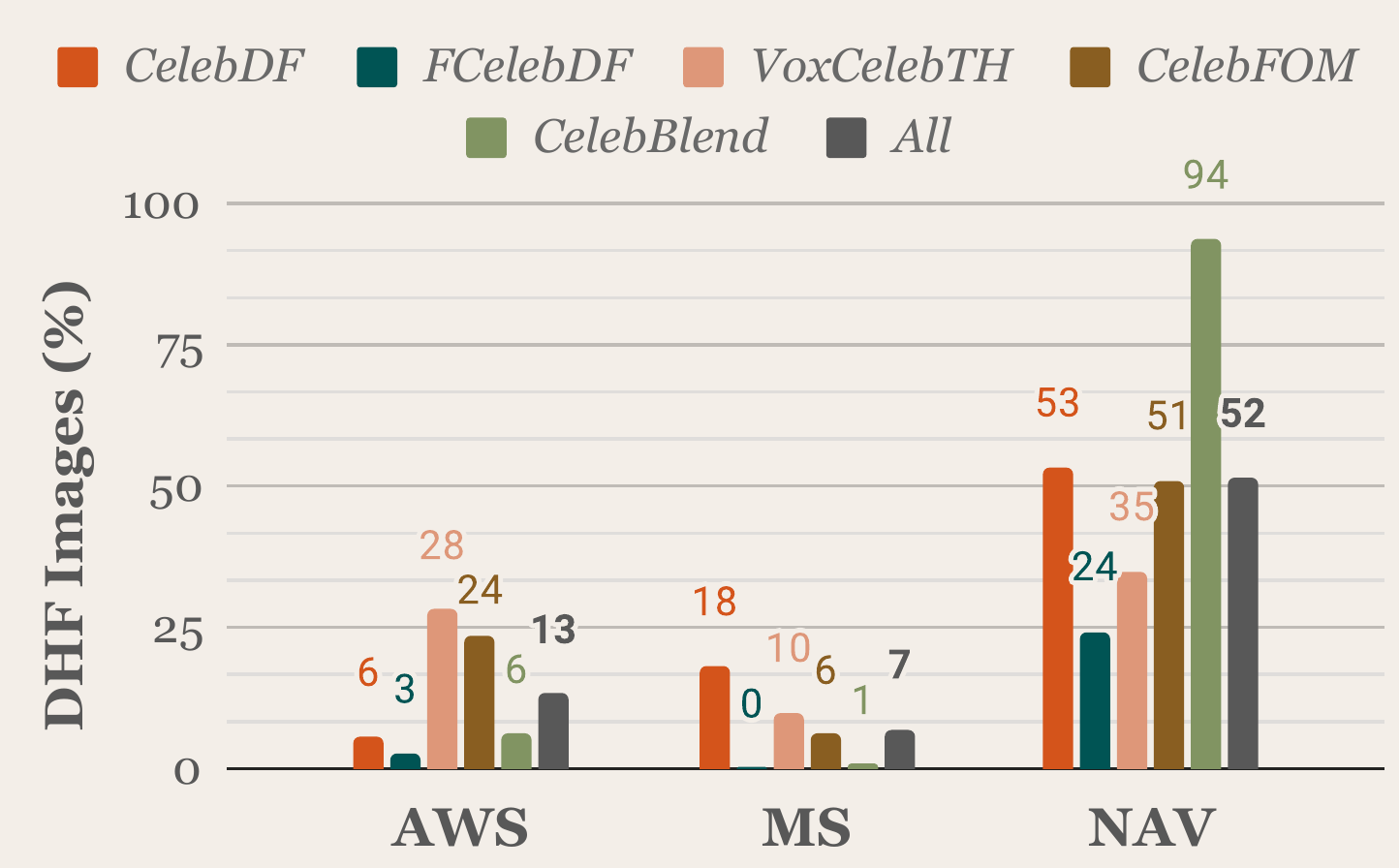}
        \caption{Deepfakes with High Fidelity: Confidence of reference (real) is less than Target (fake).}
        \label{fig:LD}
    \end{subfigure}
    \begin{subfigure}{0.32\linewidth}
        \centering
        \includegraphics[clip, trim=0pt 0pt 0pt 0pt,width=1\linewidth]{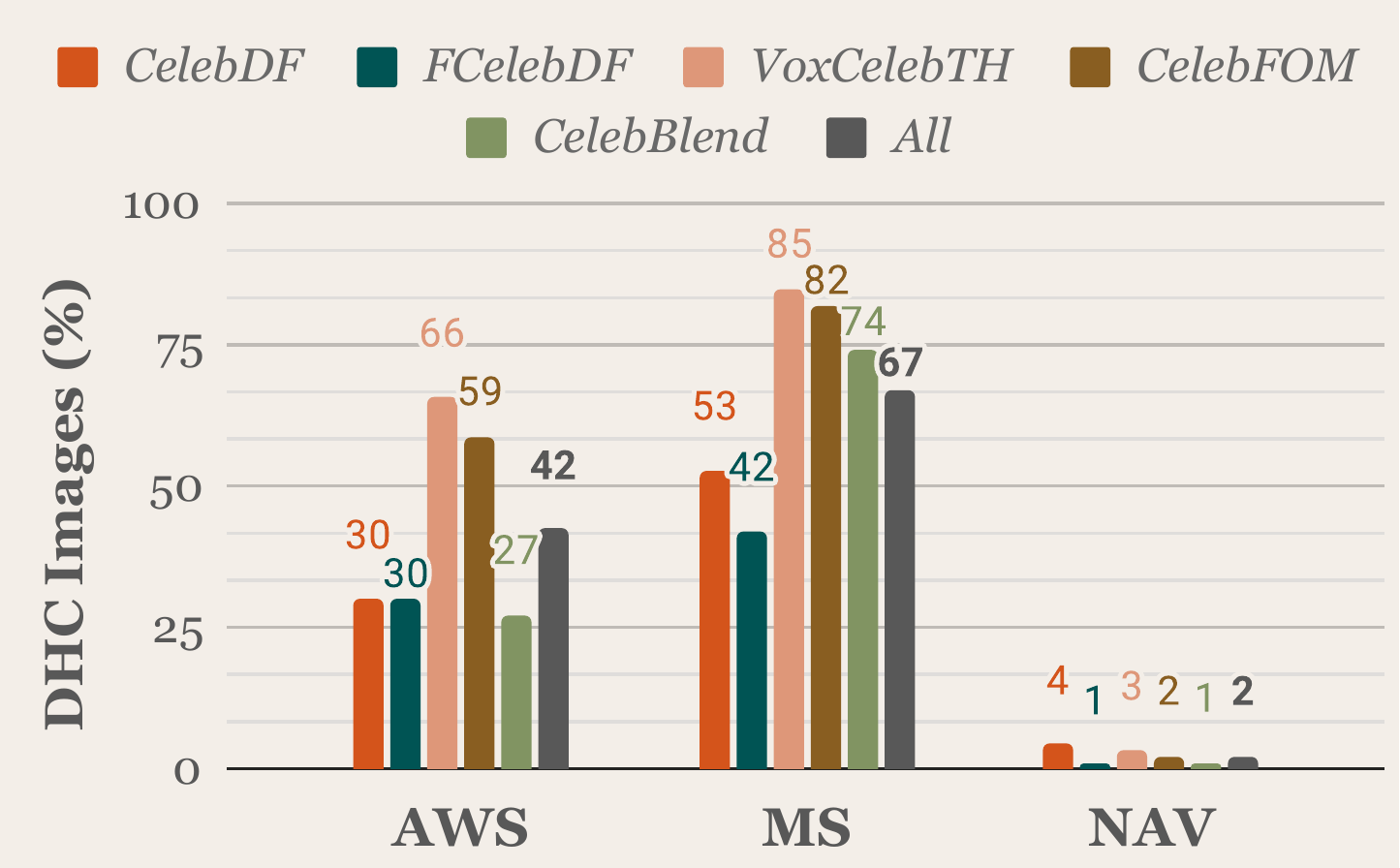}
        \caption{Deepfakes with High Confidence: The prediction confidence of a successful deepfake impersonation attack is higher than 90\%.}
        \label{fig:DHC}
    \end{subfigure}\hfill
    \begin{subfigure}{0.32\linewidth}
        \centering
        \includegraphics[clip, trim=0pt 0pt 0pt 0pt,width=1\linewidth]{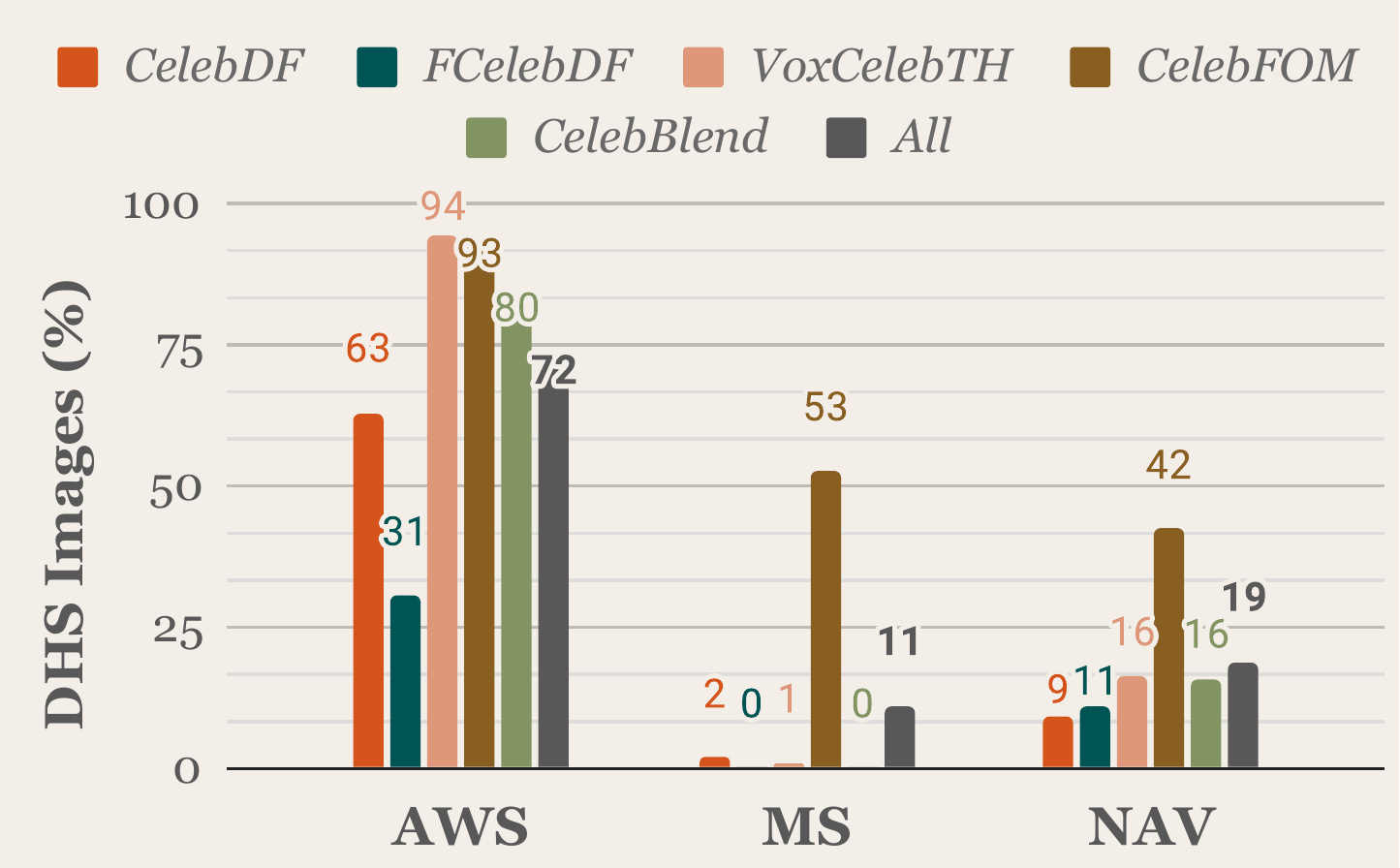}
        \caption{Deepfakes with High Similarity: Attack is unsuccessful but API shows high similarity (>80\%) between real and deepfake image.}
     \label{fig:DHS}
    \end{subfigure}\hfill
    \begin{subfigure}{0.32\linewidth}
        \centering
        \includegraphics[clip, trim=0pt 0pt 0pt 0pt,width=1\linewidth]{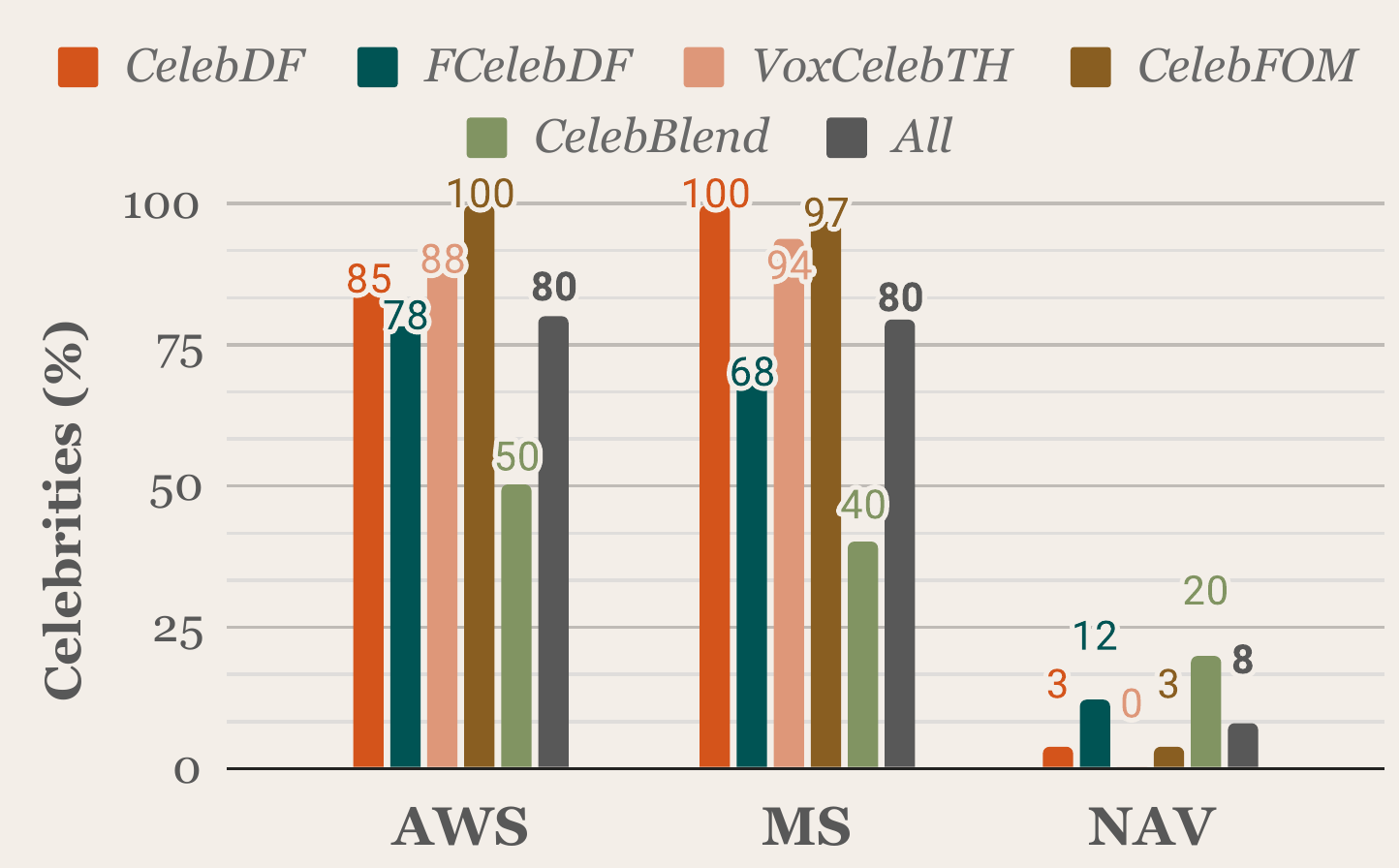}
        \caption{Successful Impersonation of Celebrities: Deepfake of a celebrity is recognized by the API as the same celebrity.}
        \label{fig:SIC}
    \end{subfigure}
    \caption{Results of the two attack scenarios and four attack analysis metrics. Note: The numbers are round to the nearest integer to increase readability (see Appendix~\ref{Appendix: Detailed results for each Dataset} for detailed results).}
    \label{fig:Results}
    %  \vspace{-10pt}
\end{figure*}

%In Table~\ref{tab:DATASET}, we describe the numbers of the celebrities and images in datasets we experimented with. For Celeb-DF, we tested 6,184 images extracted from 5,643 videos of 58 Celebrities. We had 50 Celebrities, a total of 4,952 images from 200 videos in FCelebDF. We extracted the frames from the video and selected the clear images for an accurate experiment. For Vox1CelebTH and Vox2CelebTH, we tested 1,600 images of 50 Celebrities. For CelebFOM, we extracted and used images from 2,176 videos of 58 celebrities, and we tested 180 images of 20 celebrities for CelebBlend. Each celebrity has a different number of images. 
\subsection{\textbf{Price of Commercial Web APIs}}
% \simon{we can move the price to appendix, if no space}
%\simon{not sure this information is essential...}
We perform the DI attack on Celebrity Recognition and Face Similarity APIs of AWS, MS, and NAV, where AWS, MS, and NAV API costs \$1 per 1,000 transactions under a million (M) transactions per month. However, each company has a slightly different pricing policy for over 1M transactions. NAV charges \$0.50 per 1,000 transactions for under 100M transactions, and \$0.30 for over 100M transactions. MS provides 30,000 transactions per month for free instance at 0.3 transactions per second (TPS). For standard instances provided at 10 TPS, MS imposes \$0.80 per 1,000 transactions under 5M transactions, \$0.60 for 5M to 100M transactions, and \$0.40 for over 100M transactions. AWS charges \$0.80 per 1,000 transactions for under 10M transactions, \$0.60 for under 100M transactions, and \$0.40 for over 100M transactions.

\subsection{\textbf{Evaluation Setup}}
To evaluate the web API, we first load an image from the dataset or online source. To enhance the web APIs' face recognition performance without any considerations of background and resizing~\cite{SeeingnotBelieving}, we implement MTCNN to extract face from each image on an Intel i7-9700 CPU @ 3GHz, with a CPU-based TensorFlow 1.4.1. We request the REST API of celebrity recognition web API from each company using Python 3 with the face cropped image. As a result, API returns a predicted celebrity name with confidence and coordinates of face bounding box into a JSON data type. We integrate the output into a CSV file with filenames, the predicted celebrity names of the reference image and target image, confidence, and similarity to analyze the result. We set the confidence threshold of $\beta=90\%$ and use the default similarity threshold of $\gamma=80\%$, as mentioned in the AWS API guide~\cite{AWSSimilarity} to measure DHC and DHS metric.
\textit{See Table~\ref{tab:AppendixCelebDFDataset}--\ref{tab:AppendixCelebFOMDataset} in Appendix~\ref{Appendix: Detailed results for each Dataset} for detailed results on each Celebrity.}

\subsection{DF Impersonation Attack Performance}
\textbf{Targeted Attack Performance. } Figure~\ref{fig:STA} presents the targeted attack success rate across all datasets. VoxCelebTH-based DI attack achieves the highest success against MS (78.0\%) and AWS (68.7\%) web APIs. % while it never fools NAV APIs. 
%the targeted attack, achieving the highest of only 13.3\% for the CelebBlend dataset.
From Fig.~\ref{fig:STA}, we can also observe that facial reenactment-based DI attacks such as VoxCelebTH and CelebFOM achieve the highest attack success rate. We believe this high success rate attributes to the fact that facial reenactment deepfakes tend to preserve the same identity as the target video, only manipulating the facial expressions and movements using the reference video. %\simon{this should be also mentioned and emphasized eariler when describe this method...this is hypothesis right?}. 
Therefore, AWS and MS APIs still perceive them as the real celebrity. Interestingly, we observe the lowest targeted attack success rate with NAV API. We find that this low success rate is not because NAV API is robust or strong against attacks, but because NAV API is always predicting other celebrity, which directly translates to high non-targeted attack success rate (see Fig.~\ref{fig:SNA}), we will discuss in the next section. 

On the other hand, CelebDF and CelebBlend based DI attacks achieve a success rate of less than 22\% against MS and AWS APIs. %However, generally other types of deepfakes such as CelebDF, FCelebDF, and CelebBlend yield the low success rates.}
Compared to facial reenactment, it is harder to generate high-quality deepfakes using facial replacement methods. Therefore, we find CelebDF and FCelebDF showing relatively lower performance. Furthermore, we observe the low performance of the targeted attack using the CelebBlend dataset. We believe this is because it only contains synthetic identities made by morphing two celebrities, which are more difficult to perceive as a target celebrity.  However, overall we find that state-of-the-art commercial face recognition APIs can be deceived to classify deepfakes as real ones.

%Figure~\ref{fig:STA} shows the success rate of the targeted attack on all dataset. The highest targeted Attack success rate was under the attack of MS using Vox1CelebTH, which was 78.0\%. TalkingHead datasets achieved a high attack success rate relatively. MS was the most vulnerable API to a targeted attack, scoring the highest attack success rate for four datasets. Naver had the lowest success rate with a 0\% success rate for Vox1CelebTH and Vox2CelebTH in targeted attacks.

\noindent
\textbf{Non-Targeted Attack Performance. } We present non-targeted deepfake impersonation attack success rate in Fig.~\ref{fig:SNA}. The non-targeted DI attack achieves nearly 98.6 to 99.9\% success rate for NAV API with all deepfake datasets. This high success rate attributes to the fact that NAV API always tries its best to predict the celebrity, even though it has low confidence. Therefore, NAV API always outputs the celebrity, resulting in a high non-targeted attack success rate.  %despite of the low confidence value.   
In addition, the non-targeted attack is also more successful on AWS and MS APIs, compared to those of the targeted attack, as shown in Fig.~\ref{fig:STA} and \ref{fig:SNA}. In particular, VoxCelebTH and CelebFOM achieve 77.8\% and 65.0\% attack success rate on AWS, respectively, which is nearly 10\% and 17\% higher than that of a targeted attack, respectively. Also, a similar higher performance (VoxCelebTH: 79\% and CelebFOM: 60\%) is obtained with MS API with these two datasets. We believe these two datasets using reenactment have relatively fewer changes from target video. Therefore, they also achieve an even higher success rate for the non-targeted attack.  

Moreover, we observe that AWS API is not robust against the non-targeted DI attack using Celeb-DF, FCelebDF, and CelebBlend dataset, achieving 42.5\%, 34.3\%, and 42.2\% attack success rate, respectively, as shown in Fig~\ref{fig:SNA}. These results indicate that about 40\% of the time, AWS will misclassify deepfakes as some other real celebrities, which can be problematic if deployed in a real-world case. Except for VoxCelebTH deepfake datasets, MS API demonstrates to be more robust against non-targeted DI attack, showing the lowest success rates among three APIs across four datasets. We believe MS API tends to be conservative in reporting results, where it does not over-predict when the confidence is low and only produces results with high confidence prediction. This would be the desired property to reduce false predictions. In summary, we observe that the non-targeted attack is more successful across all web APIs than the targeted attack.

%Figure~\ref{fig:SNA} shows the success rate of a non-targeted attack. Naver achieved an almost 100\% success rate in most of the datasets for non-targeted attacks. We assume it is because Naver API returns the prediction even it has low confidence. The non-targeted attack was also successful in AWS and MS API, especially with Vox1CelebTH, Vox2CelebTH, and CelebFOM. These three datasets have relatively fewer changes from refereneces. However, AWS was not robust enough to the non-targeted attack with Celeb-DF, FCelebDF, and CelebBlend, scoring 42.5\%, 34.3\%, 42.2\% of success rate, respectively.

%\subsection{Empirical Analysis}

\noindent
\textbf{Deepfakes with High Fidelity (DHF) Performance. }
In Fig.~\ref{fig:LD}, we analyze the high fidelity deepfake results, which measure the percentage of cases, where deepfake prediction confidence $\mathcal{P}^\mathtt{conf}_\mathcal{D}$ is higher than $\mathcal{P}^\mathtt{conf}_\mathcal{T}$ (i.e., the target image prediction confidence). In fact, this is a very fascinating case (counterintuitive but very interesting), resulting in the higher confidence to deepfakes over the real celebrity image. It also shows the significant problems of APIs, where they perceive deepfakes as more real.

Generally, the confidence level of real celebrities' images is indeed higher than deepfakes. However, in Fig.~\ref{fig:LD}, we highlight the results, when it is not the case. For example, AWS has a 28.2\% DHF-score on VoxCelebTH, which means out of 3,200 fake images, 902 deepfakes have higher confidence than the same celebrity's real image. NAV API shows the highest percent of DHF (93.7\%) across all five datasets, indicating that NAV API provides much higher misprediction confidence to deepfakes over real, compared to MS and AWS APIs. We find that this aggressive prediction strategy of NAV API makes it significantly vulnerable to DI attacks, as shown in Fig.~\ref{fig:LD}. 

Moreover, we observe that AWS and MS show a lower percentage of high fidelity cases, especially with FCelebDF and CelebBlend DI attacks.
%, which correlates with the relatively low targeted attack success rate.} \simon{why it is related with low target attack success rate?}  
%Based on our empirical analysis, we find that the prediction confidence for reference (real) images for Naver API is relatively higher than MS and AWS API. 
However, they are also fooled by VoxCelebTH and CelebFOM DI attack, assigning higher confidence for deepfakes $\mathcal{X_D}$ than the real celebrity image $\mathcal{X_T}$ with 28\% and 24\% for AWS and 10\% and 6\% for MS API, respectively. This behavior is consistent with targeted and non-targeted attack results, where VoxCelebTH and CelebFOM are the most effective to fool celebrity recognition APIs. MS API performs reasonably well among the three APIs but shows its worst performance on the CelebDF dataset with 18.2\%.

%In Fig.~\ref{fig:LD}, we further analyze the percentage of cases when deepfake prediction's confidence is higher than the reference image's prediction confidence. Naver API shows the highest rate for all five datasets, which means Naver API recognizes deepfake as real more than the MS and AWS APIs. Moreover, we observed that AWS and MS show a lower percentage of such cases, especially with FCelebDF and CelebBlend datasets, which were also relatively less effective for the targeted attack. Based on our empirical analysis, we found that the prediction confidence for reference (real) images for Naver API is relatively higher than MS and AWS API. However, we also observed that this aggressive prediction strategy of Naver API weakens it against deepfake impersonation attacks, as shown in Fig.~\ref{fig:LD}.
%Figure~\ref{fig:LD} shows the ratio of the case when the confidence of fake images prediction is higher than predicted confidence in references. Naver shows the highest rate for all six datasets, which means Naver API recognizes deepfake into real more than the references. However, AWS and MS show lower accuracy, especially with FCelebDF and CelebBlend, which were relatively less effective for the targeted and non-targeted attack. 
\begin{figure*}
    \centering
    \includegraphics[width=0.99\linewidth]{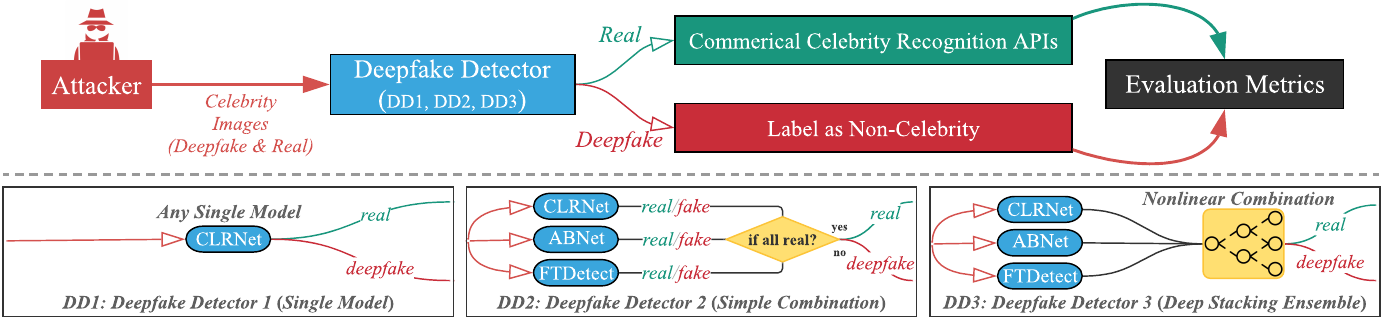}
    \caption{We propose 3 defense settings (DD1, DD2, and DD3) against the Deepfake Impersonation attack on web APIs.}
    \label{fig:Defense}
    % \vspace{-5pt}
\end{figure*}

\noindent
\textbf{Deepfakes with High Confidence (DHC) Performance. }
%Next, we analyzed cases when prediction confidence of the deepfake $\mathcal{P}^\mathcal{D}_{conf}$, predicted as a celebrity, is more than a specific threshold ($\beta=90\%$) to observe the percentage of the deepfakes classified with very high confidence in Fig.~\ref{fig:DHC}.% We set the threshold $\beta$ to be 90\% t %This metric can capture the unreliability of predictions for each web API.
In Fig.~\ref{fig:DHC}, we show the cases when the prediction confidence of the deepfake $\mathcal{P}^\mathtt{conf}_\mathcal{D}$, predicted as a celebrity, is more than a specific threshold ($\beta=90\%$). This allows analyzing the percentage of the deepfakes classified with very high confidence to capture the vulnerability of the recognition system. 
Interestingly, MS API, which achieves higher robustness against the non-targeted attacks, shows a noticeably high deepfake prediction confidence percentage, ranging from 42\% to 86.5\%. This represents that when MS API predicts deepfakes as a celebrity, it does so with very high confidence. Similarly, AWS has the second-highest DHC for all datasets, from 27.3\% to 73.4\%. We can also examine that VoxCelebTH and CelebFOM have a higher rate than other datasets.

% \simon{i am not entirely understand. u need to explain this to me in person...}
On the other hand, the NAV API exhibits very few cases in all datasets where the prediction confidence of deepfakes $\mathcal{P}^\mathtt{conf}_\mathcal{D}$ is more than 90\%, achieving a DHC of only 4.3\% in the worst case. This result demonstrates that NAV API rarely provides high confidence predictions to deepfakes $\mathcal{X_D}$. In fact, this aspect can be integrated as a broader defense of NAV API, where it can reject the low confidence value of input value, including deepfakes. 
% and therefore is robust against them in this manner.
%Even though Naver API is vulnerable to deepfakes, as shown in Fig.~\ref{fig:STA}--~\ref{fig:LD}. However, NAV provides a low confidence score. \simon{this means what?}

Previously, the impersonation attack using the CelebBlend dataset showed a relatively lower attack success rate for MS API. However, we observe that CelebBlend achieves a high DHC with 74.4\% for MS API. Therefore, VoxCelebTH, CelebFOM, and CelebBlend datasets are useful for fooling MS API and performing impersonation attack, with a high confidence score. In summary, MS and AWS APIs appear to be more robust. However, DHC results demonstrate that if MS and AWS APIs make mistakes, they tend to do so with high confidence.

%In Fig.~\ref{fig:DHC}, we explain the case when the confidence is more than 90\%. This metric also shows how the outputs of each API are reliable. In Naver's case, confidence is generally not high, under 5\% for all datasets. On the other hand, MS shows noticeably the highest rate of about 42\% to 86.5\%. AWS has the second-highest rate for all datasets, which was 27.3\% to 73.4\%. We can also notice that Vox1CelebTH, Vox2CelebTH, and CelebFOM have a higher rate than other datasets. It means that these three datasets are highly useful to fool celebrity recognition APIs.

\noindent
\textbf{Deepfakes with High Similarity (DHS) Performance. }
%Figure~\ref{fig:DHS} presents the case when the attack is unsuccessful, but the similarity score between deepfake $\mathcal{X_D}$ and target celebrity image $\mathcal{X_T}$ is above a specific threshold ( which determines the percentage of deepfakes $\mathcal{X_D}$ that are not recognized as a celebrity but are highly similar to target celebrity image $\mathcal{X_T}$.
 We pre\-sent the case when the attack is unsuccessful in Fig.~\ref{fig:DHS}. However, the Face Similarity (FS) score between deepfake $\mathcal{X_D}$ and target celebrity image $\mathcal{X_T}$ is above a specific threshold ($\gamma=80\%$). This metric can be still useful, determining the percentage of $\mathcal{X_D}$ not recognized as a celebrity but are still highly similar to $\mathcal{X_T}$.

Surprisingly, the AWS API achieves the highest DHS percentage for all datasets, ranging from 30\% to 94.4\%, which indicates that the AWS face similarity API is considerably more susceptible to DI attacks than MS and NAV APIs. %Suppose Celebrity Recognition (CR) fails, then we use Face Similarity (FS) as the second evaluation step. In that case, based on the results from Fig~\ref{fig:Results}, there is high certainty that the AWS Face Similarity API will return a high similarity score between $\mathcal{X_D}$ and $\mathcal{X_T}$.
It is also interesting to observe that AWS API, which demonstrates a low success rate for CelebDF, FCelebDF, and CelebBlend datasets in targeted attack settings, shows a relatively high DHS percentage for the same datasets. 
%The AWS API shows the highest \hl{DHS percentage} for all datasets, ranging from 30\% to 94.4\%, i.e., it is more vulnerable against deepfakes \st{in this analysis.} \simon{I am not sure how we can convince it is more vulnerable with similarity...how can we show? add your justification}
Generally, MS and NAV APIs offer a low DHS percentage for most datasets and recognizes deepfakes well but shows a high DHS of 52.6\% (MS) and 42.3\% (NAV) for deepfakes from the CelebFOM dataset, as shown in Fig.~\ref{fig:DHS}.

%\hl{Since NAV API does not provide face similarity function API, we assumed that the deepfake and reference image had high similarity when the reference prediction and target prediction were the same.} \simon{unclear, explain to me.}
%\simon{I really dont know the key point here...u need to pitch well on the usefulness of this metric and result...}
%In Fig.~\ref{fig:DHS}, we describe the case when the attack was unsuccessful, but the similarity was high, which is above 80\%. Since Naver does not provide face similarity API, we assumed that the deepfake and reference image have high similarity when the reference prediction and target prediction were same. As a result, AWS shows the highest success rates of all datasets, i.e., it is more vulnerable against deepfakes. MS API offers a lower rate for most datasets and recognizes deepfakes well. In CelebFOM, however, the 52.6\% of the output showed high similarity.

\noindent
\textbf{Successful Impersonation of Celebrities (SIC) Performance. }
%\simon{not sure how this differs from target vs. non-targeted attack.}
%We also analyzed the cases when the deepfake impersonation attack fools the web APIs with at least one deepfake of a celebrity (e.g., Obama) being recognized as the real celebrity (e.g., Obama).  \simon{I dont know how this works...how u measure...}
Lastly, in Fig.~\ref{fig:SIC}, we calculate the total percentage of unique and successful celebrity impersonations, i.e.,  each celebrity is counted just once for a dataset when any of the deepfakes in a DI attack dataset can successfully deceive the API. We find that MS API shows the highest SCI of 100\%, 96.6\%, and 94\% for deepfakes from CelebDF, CelebFOM, and  VoxCelebTH datasets, which means that the attacker can successfully impersonate 58 out of 58 celebrities in the CelebDF dataset, 55 out of 58 celebrities in CelebFOM, and 94 out of 100 celebrities in the VoxCelebTH dataset, respectively. We can also observe similar results for the AWS API in Fig.~\ref{fig:SIC}. 

However, the NAV API shows a relatively low SCI percentage of less than 21\% for all datasets. Upon further investigation, we find that NAV API is biased toward Asian and mostly Korean celebrities. Therefore, all real Asian celebrities $\mathcal{X_T}$ are accurately recognized, and their deepfakes $\mathcal{X_D}$ are also recognized as $\mathcal{X_T}$ by the NAV API. However, many non-Asian celebrities are not. We suspect that NAV's celebrity database is highly optimized for local Asian celebrities. In addition, the low ratio on the CelebBlend dataset across all APIs in Fig.~\ref{fig:SIC} is because CelebBlend contains some new celebrities. All three APIs recognize neither the real images of these new celebrities nor their deepfakes.

%and MS API demonstrate a very high SIC (up to 99.9\%) for all datasets except for CelebBlend, meaning that we can impersonate nearly all celebrities using deepfakes from these datasets.

%Other than the CelebBlend dataset, a considerable ratio of celebrities have at least one deepfake recognized as themselves by the MS (68.0\% to 99.9\%) and AWS (78.0\% to 99.99\%) API (see Fig~\ref{fig:SIC}). %\simon{i dont understand}

%This result shows that the deepfakes $\mathcal{X_D}$ were impersonation attack effectively fool the MS and AWS APIs for nearly all celebrities in our dataset. \simon{i dont understand}

%Figure~\ref{fig:SIC} shows the case when the web APIs are successfully fooled so that the APIs recognized deepfake images of celebrity as original celebrity. AWS and MS showed 100\% accuracy for CelebFOM and Celeb-DF, respectively. In other word, the deepfake of celebrity fooled AWS and MS with high performance, making the APIs to recognize the deepfake as reference celebrity. Naver had low accuracy, under 20\% to even 0\% for every dataset. However, it is hard to say that deepfakes failed to fool Naver because we have to consider the prediction accuracy of Naver API with original images.

\subsection{\textbf{Summary of DI Attack Results}}
\textbf{Attack Performance of Different APIs. } 
We observe that all three APIs perform uniquely for targeted and non-targeted attack scenarios. To summarize:  1) NAV API shows the lowest success rate on the targeted attack and the highest success rate for the non-targeted attack, 2) AWS API is the most robust against the targeted attack, and 3) MS API is the most robust against non-targeted attack. Therefore, there is no clear winner among the three APIs.

\noindent
\textbf{Vulnerabilities of Different APIs. } 
Each DI attack evaluation metric reveals distinct vulnerabilities in the APIs. For instance, NAV API is the most vulnerable to deepfakes with high fidelity (DHF), AWS is highly susceptible to deepfakes with high similarity (DHS), and MS is most defenseless against deepfakes with high confidence (DHC). Furthermore, MS and AWS APIs both performed poorly for successful impersonation of celebrities (SCI) analysis.

%First, we explain the differences between API providers. The two attack scenarios (targeted and non-targeted) results are slightly different between AWS and MS but generally showed similar patterns (see Fig~\ref{fig:STA} and \ref{fig:SNA}). However, we learned new characteristics of the two APIs (MS and AWS) through the attack analysis metrics. For example, the two APIs have a difference in the confidence of the output when the reference image's confidence is lower than that of the fake, as shown in Fig.~\ref{fig:LD}, where AWS has a higher ratio than MS for all datasets except Celeb-DF. Whereas, in Fig.~\ref{fig:DHC} MS API has a higher rate than AWS for all datasets results when deepfakes have high prediction confidence of being real. \simon{need to rewrite whole}

\noindent
\textbf{Performance of Different DI Attack Datasets and Generation methods. }
% \textbf{Different Deepfake Attack Generation Methods (Datasets).}
Each DI attack dataset also performs differently for each API. However, we find some interesting patterns in the results. For example, 1) facial reenactment based methods such as VoxCelebTH and CelebFOM generally demonstrate the most success, and 2) in facial replacement based methods, CelebDF always shows more success in attack and analysis than FCelebDF. 
These can be used for deciding the type of deepfake generation methods to exploit specific APIs to carry out more effective DI attacks.

\begin{table*}
\centering
\caption{Performance of deepfake detection methods and updated attack success rates after deploying our defense. Bold indicates the best defense result for each dataset. CLRNet is used as DD1, and (CLRNet, ABNet, FTDetect) for DD2 and DD3.}
\label{tab:defense}
\resizebox{1\linewidth}{!}{%
\begin{tabular}{l||lc||c|c|c||c|c|c} 
\toprule
\multirow{3}{*}{ \textbf{Dataset} } & \multicolumn{2}{c||}{\textbf{Defense}} & \multicolumn{6}{c}{\textbf{Updated \textit{attack success rate} after employing proposed \textit{defense} method} } \\ 
\cline{2-9}
 & \multirow{2}{*}{\textbf{Method}} & \multirow{2}{*}{\textbf{Acc.}} & \multicolumn{3}{c||}{\textbf{Targeted Attack}} & \multicolumn{3}{c}{\textbf{Non-targeted Attack}} \\ 
\cline{4-9}
 &  &  & \textit{AWS}  & \textit{MS}  & \textit{NAV}  & \textit{AWS}  & \textit{MS}  & \textit{NAV}  \\ 
\hhline{=::==::===::===}
\multirow{3}{*}{\begin{tabular}[c]{@{}l@{}}FCelebDF\\(ours) \end{tabular}} & Xception & 89.7\% & 11.3\%$\rightarrow$1.16\%  & 5.90\%$\rightarrow$0.60\%  & 5.20\%$\rightarrow$0.54\%  & 34.4\%$\rightarrow$3.54\%  & 7.20\%$\rightarrow$0.74\%  & 98.6\%$\rightarrow$10.2\%  \\
 & ShallowNet & 90.5\% & 11.3\%$\rightarrow$1.07\%  & 5.90\%$\rightarrow$0.56\%  & 5.20\%$\rightarrow$0.50\%  & 34.4\%$\rightarrow$3.27\%  & 7.20\%$\rightarrow$0.68\%  & 98.6\%$\rightarrow$9.37\%  \\
 & \textbf{CLRNet}  & \textbf{97.9\%} & 11.3\%$\rightarrow$\textbf{0.23\%}  & 5.90\%$\rightarrow$\textbf{0.12}\%  & 5.20\%$\rightarrow$\textbf{0.10}\%  & 34.4\%$\rightarrow$\textbf{0.69}\%  & 7.20\%$\rightarrow$\textbf{0.14}\%  & 98.6\%$\rightarrow$\textbf{1.98}\%  \\ 
\hhline{=::==::===::===}
\multirow{3}{*}{CelebDF} & Xception & 89.9\% & 10.7\% $\rightarrow$1.08\%  & 21.3\% $\rightarrow$2.15\%  & 2.10\% $\rightarrow$0.22\%  & 42.5\% $\rightarrow$4.29\%  & 24.4\% $\rightarrow$2.46\%  & 99.8\% $\rightarrow$10.1\%  \\
 & CLRNet & 98.2\% & 10.7\%~$\rightarrow$0.19\% & 21.3\%~$\rightarrow$0.38\% & 2.10\%~$\rightarrow$0.04\% & 42.5\%~$\rightarrow$0.77\% & 24.4\%~$\rightarrow$0.44\% & 99.8\%~$\rightarrow$1.80\% \\
 & \textbf{ABNet}  & \textbf{99.9\%} & 10.7\% $\rightarrow$\textbf{0.01\%}  & 21.3\% $\rightarrow$\textbf{0.02}\%  & 2.10\% $\rightarrow$\textbf{0.00}\%  & 42.5\% $\rightarrow$\textbf{0.04}\%  & 24.4\% $\rightarrow$\textbf{0.02}\%  & 99.8\% $\rightarrow$\textbf{0.10}\%  \\ 
\hhline{=::==::===::===}
\multirow{3}{*}{VoxCelebTH} & ShallowNet & 90.2\% & 68.7\%$\rightarrow$5.56\%  & 78.0\%$\rightarrow$5.98\%  & 0.00\%$\rightarrow$0.00\%  & 77.8\%$\rightarrow$6.76\%  & 78.8\%$\rightarrow$7.05\%  & 99.9\%$\rightarrow$9.79\%  \\
 & CLRNet & 96.8\% & 68.7\%$\rightarrow$1.81\% & 78.0\%$\rightarrow$1.95\% & 0.00\%$\rightarrow$0.00\%~ & 77.8\%$\rightarrow$2.21\% & 78.8\%$\rightarrow$2.30\% & 99.9\%$\rightarrow$3.20\% \\
 & \textbf{FTDetect}  & \textbf{98.4\%}  & 68.7\%$\rightarrow$\textbf{0.88\%}  & 78.0\%$\rightarrow$\textbf{0.95}\%  & 0.00\%$\rightarrow$\textbf{0.00}\%  & 77.8\%$\rightarrow$\textbf{1.12}\%  & 78.8\%$\rightarrow$\textbf{1.23}\%  & 99.9\%$\rightarrow$\textbf{1.56}\%  \\ 
\hhline{=::==::===::===}
\multirow{3}{*}{\begin{tabular}[c]{@{}l@{}}CelebFOM\\(ours) \end{tabular}} & ShallowNet & 85.5\% & 47.5\%$\rightarrow$6.89\%  & 58.3\%$\rightarrow$8.45\%  & 2.60\%$\rightarrow$0.38\%  & 65.0\%$\rightarrow$9.43\%  & 60.3\%$\rightarrow$8.74\%  & 99.8\%$\rightarrow$14.5\%  \\
 & MesoNet & 92.2\% & 47.5\%$\rightarrow$3.71\% & 58.3\%$\rightarrow$4.55\% & 2.60\%$\rightarrow$0.20\% & 65.0\%$\rightarrow$5.07\% & 60.3\%$\rightarrow$4.70\% & 99.8\%$\rightarrow$7.78\% \\
 & \textbf{CLRNet}  & \textbf{95.8\%} & 47.5\%$\rightarrow$\textbf{2.01\%}  & 58.3\%$\rightarrow$\textbf{2.47}\%  & 2.60\%$\rightarrow$\textbf{0.11}\%  & 65.0\%$\rightarrow$\textbf{2.76}\%  & 60.3\%$\rightarrow$\textbf{2.56}\%  & 99.8\%$\rightarrow$\textbf{4.23}\%  \\ 
\hhline{=::==::===::===}
\multirow{3}{*}{\begin{tabular}[c]{@{}l@{}}CelebBlend\\(ours)\end{tabular}} & Xception & 98.2\% & 13.9\%$\rightarrow$0.25\% & 19.0\%$\rightarrow$0.34\% & 13.3\%$\rightarrow$0.24\% & 42.2\%$\rightarrow$0.76\% & 21.2\%$\rightarrow$0.38\% & 99.9\%$\rightarrow$1.80\% \\
 & MesoNet & 98.5\% & 13.9\%$\rightarrow$0.21\% & 19.0\%$\rightarrow$0.29\% & 13.3\%$\rightarrow$0.20\% & 42.2\%$\rightarrow$0.63\% & 21.2\%$\rightarrow$0.32\% & 99.9\%$\rightarrow$1.50\% \\
 & \textbf{CLRNet} & \textbf{98.8\%} & 13.9\%$\rightarrow$\textbf{0.17\%} & 19.0\%$\rightarrow$\textbf{0.23\%} & 13.3\%$\rightarrow$\textbf{0.16\%} & 42.2\%$\rightarrow$\textbf{0.51\%} & 21.2\%$\rightarrow$\textbf{0.25\%} & 99.9\%$\rightarrow$\textbf{1.20\%} \\
 \hhline{=::==::===::===}
 \multirow{3}{*}{\begin{tabular}[c]{@{}l@{}}\textbf{All Datasets}\\\textbf{Combined}\end{tabular}} & DD1 & 97.5\% & 28.0\%$\rightarrow$0.70\% & 33.1\%$\rightarrow$0.83\% & 4.70\%$\rightarrow$0.12\% & 50.6\%$\rightarrow$1.27\% & 37.0\%$\rightarrow$0.92\% & 99.6\%$\rightarrow$2.49\% \\
 & DD2 & 85.8\% & 28.0\%$\rightarrow$3.98\% & 33.1\%$\rightarrow$4.70\% & 4.70\%$\rightarrow$0.66\% & 50.6\%$\rightarrow$7.19\% & 37.0\%$\rightarrow$5.25\% & 99.6\%$\rightarrow$14.1\% \\
 & \textbf{DD3} & \textbf{98.2\%} & 28.0\%$\rightarrow$\textbf{0.50\%} & 33.1\%$\rightarrow$\textbf{0.60\%} & 4.70\%$\rightarrow$\textbf{0.08\%} & 50.6\%$\rightarrow$\textbf{0.91\%} & 37.0\%$\rightarrow$\textbf{0.67\%} & 99.6\%$\rightarrow$\textbf{1.79\%} \\
\bottomrule
\end{tabular}
}
% \vspace{-5pt}
\end{table*}
\section{Proposed Defense Mechanism}
\label{sec:defense}
% \hl{Explain and extend this section}

As previously discussed, state-of-the-art commercial APIs are vulnerable to deepfake-based impersonation attacks. Therefore, we need a defense mechanism to enhance the face recognition model's robustness and defend against DI attacks. We set two requirements on the system: first, it should be reliable and easily deployable through existing software development efforts; second, it should be integrated with minimal changes to the existing infrastructure.
%For example, AWS may want to offer a ``deepfake robust'' version of their face recognition API, \sh{but they do not want to retrain their entire face recognition model.}\simon{unclear}.
We address the first requirement by using high performing, off-the-shelf deepfake detectors, which were empirically validated to perform well against deepfakes~\cite{ABNet,FakeTalkerDetect,tariq2020convolutional,Xception,MesoNet,FakeSpotter,ShallowNet1,ShallowNet2} (see Appendix~\ref{Appendix: Experimental Settings} for the details on training of these baselines). Meeting the second requirement is crucial for the sake of adaptability; we design our own defense mechanism, such that it wraps the face recognition API, resulting in a black-box, yet deepfake-robust version. In other words, any image which is to be evaluated by the web APIs will first need to pass a deepfake filter check. Our proposed defense approach is presented in Fig.~\ref{fig:Defense}. 

\noindent
\textbf{Defense Setting. } We also design and compare the defense mechanism using three deepfake detector (DD) settings, as shown in Fig.~\ref{fig:Defense}.
\begin{enumerate}[leftmargin=15pt]
    \item \textit{Single Model $(DD1)$}: We train a single model separately on each dataset and evaluate its performance against DI attacks.
    \item \textit{Simple Combination $(DD2)$}: We select the three best performing models (CLRNet, ABNet, FTDetect) over all datasets. Suppose all three detection models classifies an image as real (i.e., ``if all real?'' condition), as shown in Fig.~\ref{fig:Defense}. In this case, it passes on to the celebrity recognition API; otherwise, it is classified as a deepfake and labeled as as non-celebrity.
    \item \textit{Deep Stacking Ensemble $(DD3)$}: The main difference from a simple combination approach is that we use a stacking ensemble instead of a simple ``if all real?'' condition. We use the weights from all three models and form a stacked ensemble with non-linear ensemble functions that take both model scores and data as the input to predict an output.
\end{enumerate}
In a real-world environment, we can never know which attack generation method has been used (e.g., if the attack has been generated using CelebDF or CelebFOM). Hence, a model trained on one dataset may not lead to competitive performance. Consequently, it is critical to have a detector that performs reasonably well on various generation methods, which is the reason we design the DD2 and DD3 settings.

\noindent
\textbf{Defense Evaluation Details. } The number of deepfakes and real images used for the evaluation of the defense mechanism remains constant to prevent any data imbalance issues, which can significantly affect the detection accuracy. %\st{We up-sample real images if they are less than Deepfakes or vice versa.} \simon{do we need?} 
The training and testing sets do not overlap and we only use the test images (see Table~\ref{tab:DATASET}, column--4) to evaluate the attack and defense performances.

\noindent
\textbf{Defense Performance under Single Model Setting. } We evaluate the performances of a number of deepfake detection models~\cite{ABNet,FakeTalkerDetect,tariq2020convolutional,Xception,MesoNet,FakeSpotter,ShallowNet1,ShallowNet2} and present the results of the top-3 defense methods for each dataset in Table~\ref{tab:defense}. ABNet and FTDetect achieve the best performance for CelebDF (99.9\%) and VoxCelebTH (98.4\%). On the other hand, CLRNet yields the best accuracy for FCelebDF (97.9\%), CelebFOM (95.8\%), and CelebBlend (98.8\%). CLRNet is also the second-best method for Celeb-DF (98.2\%) and VoxCelebTH(96.8\%), making it the most robust defense mechanism against all DI attacks. Therefore, we select CLRNet as the representative of Single Model Setting (DD1) and use it for further experiments.

\noindent
\textbf{Comparison between DD1, DD2 and DD3. } Since we find that ABNet, FTDetect, and CLRNet are the top three best methods based on the Single Model setting results, we use them to carry out further experiments. We train DD1 (CLRNet), DD2 (ABNet, FTDetect, and CLRNet), and DD3 (ABNet, FTDetect, and CLRNet) collectively on all datasets (see Table~\ref{tab:defense} ``All Dataset Combined'' row). We observe that DD1 (CLRNet), being a Single model, can perform relatively well for all datasets (97.5\%), while using the DD2 setting resulted in the worst performance (85.8\%). This drop in performance is mainly attributed to the simple ``if all real?" condition in DD2 (see Fig.~\ref{fig:Defense}), which results in many false-positives. We achieve the best performance (98.2\%) using the DD3, where the complex non-linear combinations of the deep stacking ensemble setting enabled DD3 to yield far less false-positives compared to the simple DD2. %\simon{what the holy cow is this?}

\noindent
\textbf{Significant Reduction of Attack Success Rate after Defense. }
%Table~\ref{tab:defense} shows the result 
%The CLRNet performed the best for FCelebDF and CelebFOM datasets. Furthermore, ABNet and FTDetect achieved the highest accuracy for Celeb-DF and VoxCelebTH datasets, respectively, as shown in Table~\ref{tab:defense} column 1-3.
We observe that using any of the top-3 deepfake detection methods, the attack success rate is significantly decreased for a given dataset. Based on the results in Table~\ref{tab:defense}, we choose the best three models (CLRNet, ABNet, and FTDetect) to build our defense mechanism, as described in Fig.~\ref{fig:Defense}.
Similar to the attack scenario presented in Fig.~\ref{fig:Overview}, the attacker will provide celebrities' deepfake to the web APIs. However, this time, they have to first pass through the deepfake detector wrapper. In column 3 of Table~\ref{tab:defense}, we present the updated success rates of targeted and non-targeted attacks using our proposed defense mechanisms. The attack success rates for all APIs dropped significantly to as low as 0.0\% (see Table~\ref{tab:defense}).

% ~\sowon{The attack success rates for all APIs reduce significantly to as low as 0.0\% (see Table~\ref{tab:defense}).} %The attack success rate for all APIs reduces significantly to as low as 0.0\% (see Table~\ref{tab:defense}.
For example, the NAV API, which had a 99.8\% non-targeted attack success rate, has decreased to 0.10\% using our defense mechanism (see Table~\ref{tab:defense}). We can observe similar trends in Table~\ref{tab:defense}, clearly indicating that our defense mechanism performs effectively without causing any change in the API's underlying infrastructure.

\section{Discussion}
\label{sec:discussion}

\textbf{Confidence Threshold vs. Prediction. } Based on our experiments with the three commercial celebrity recognition APIs, we find that the NAV API always returns some celebrity's name, as it returns predictions even when its confidence is as low as 0.01\%. In contrast, the AWS and MS API do not return any celebrity's name as their prediction unless their confidence in it is equal to or more than approximately 55\% and 70\%, respectively, based on our empirical observations. Therefore, we can infer that NAV API is the most proactive at reporting, and the MS API is the most conservative at predicting the results. Nevertheless, whenever the MS API predicts the deepfake as a celebrity, it predicts with high confidence on all attack datasets (i.e., greater than 87.25\% on average, even in the worst case), and MS API is also significantly vulnerable against deepfake images. 
There is a clear trade-off of having higher or lower confidence for producing a prediction result. We can indeed set a threshold and future work should explore for determining an optimum threshold. However, the current work's main objective is to compare and examine the APIs in their current state as it is, observing differing API behaviors, without adding any constraints (by ourselves), because this is how most end-users will be using and experiencing. In this way, we can demonstrate the aspects and areas where each API needs to improve.

\noindent
\textbf{API Prediction Bias. }
We found that the MS and AWS APIs being developed by western countries have a bias towards western celebrities. They often failed to identify or misidentify the celebrities from the Asian countries. For example, the real photo of famous Asian K-pop singer IU is not recognized as a celebrity by the MS API. Likewise, famous Asian/Indian actress Madhuri Dixit is not recognized by both AWS and MS APIs. Similarly, NAV API developed by a South Korean (Asian) company shows a similar kind of bias as most of the time, incorrectly predicting the photos of western celebrities with Asian celebrities. Especially, we counted the repeated celebrities from the NAV API result. Overall, we tested 16,239 images, and 125 Asian celebrities were repeated over ten times for a total of 9,936 images (See Fig.~\ref{fig:Racial_Distribution_appendix} in Appendix~\ref{Appendix: Racial Distribution of Datasets and Prediciton from API} for detailed results on racial distribution). %\simon{maybe we should connect with demorgaphics.}

\noindent
\textbf{Multiple Images for Recognition. } Our work only looks at stills  (i.e., single images) in face recognition. Other face recognition software, such as the ones used to unlock mobile devices, can take multiple pictures and require movement. It would be interesting to examine how these APIs will react to videos or a set of consecutive images. However, it is not currently possible because commercial APIs do not provide this capability. They take only a single image for the prediction. Therefore, we performed a small experiment by providing several consecutive video frames to the API, one by one, and calculated the average prediction score. Interestingly, we found that the results are very similar to the single image results (See Fig.~\ref{fig:Angelina_appendix} and \ref{fig:Alexis_appendix} in Appendix~\ref{Appendix: Additional Discussion on Results}).

\noindent
\textbf{Generic Defense Method. } The proposed defense method can provide excellent results. And, to some extent, it can be an effective defense mechanism. However, these off-the-shelf models may not be optimal against each DI attack, and false positives can play a vital role in increasing the attack's success rate. In addition, due to the rise of new deepfakes, existing detection models are not guaranteed to work well against them. Therefore, a more generic and effective defense method against different types of existing and new DI attacks is urgently required. And more research is needed in that direction, exploring transfer learning, domain adaptation, and meta transfer learning to better cope with new DI attacks.

%\noindent
%\textbf{Transferable to Other Face API. }
%\textbf{Transferable to Other Face API. }
%We consider AWS, MS, and NAV API as a good representative of all other face recognition APIs. Therefore, ~\hl{by successfully demonstrating the high performance of DI attacks on those three APIs in a black-box setting,} we can make this conjecture that similar performance can be achieved on other face recognition APIs. Furthermore, based on black-box success, we hypothesize that our deepfake impersonation attack can be equally or more effective in a white-box setting.

\noindent
\textbf{Limitations and Future Work. } The current approach does not consider the hand-manipulated fakes and face swaps of celebrities generated by the sophisticated photoshop editing tools~\cite{ShallowNet1}. We plan to experiment with them in the future to measure the robustness of these APIs further. Also, we have not tried all types of deepfake generation methods, though we tried to cover the major ones. Nevertheless, to the best of our knowledge, we are the first to evaluate deepfakes as impersonation attacks to evaluate the state-of-the-art commercial face recognition APIs' robustness to reveal the potential vulnerability in such systems.
%Initially, we also planned to evaluate Google's celebrity recognition API as well. However, it is under restricted access as the feature is intended to be used by media and entertainment companies, or partners approved by those companies, on professionally-produced media content. Therefore, our request to access this API was rejected. We also experimented with the Alibaba cloud API. However, it only provides limited functionality that can only be used for content moderation, and therefore, it did not apply to our use cases.
Future work will involve including more celebrity recognition APIs in our pool. Further, we plan to develop a more generic and robust defense method to incorporate into the commercial systems.  

%\simon{sounds like we did not do much and there are many things to do...we need to write this more carefully.}

\noindent
\textbf{Ethical and Privacy Issues. }
%Deepfake videos have recently surfaced, the vast majority of which use face pictures of female celebrities to create sexually explicit videos with no consent. Such videos circulate quickly across the Internet and cause severe problems as innocent people are abused in cyberspace. We have done this work to evaluate celebrity recognition APIs with the Deepfake  videos due to the seriousness of the issue and the lack of useful tools for detecting these videos. 
%As deepfakes are evolving, there are no concrete and firm    
In addition to celebrity deepfake benchmark datasets, we collected publicly available deepfake videos from the Internet, and all researchers have been informed about the comprehensive research protocols. We also met with our institution's Institutional Review Board (IRB), and they acknowledged that approval is not required because the videos have already been posted on the Internet. We only used one computer for the preprocessing and experiments to prevent ourselves from even accidentally distributing these. Also, we did not test any deepfakes for children/minors. 
We would also like to emphasize that there are no ethical or privacy concerns with our use of photos from FCelebDF videos. We cropped only the faces from deepfake videos to evaluate the vulnerabilities of web APIs against realistic FCelebDF. We tested 5,000 FCelebDF images with 50 celebrities and public figures. As final proof, we checked the recently proposed deepfake related laws in different countries from Asia, the US, and Europe, consulting a law professor in our institution. We found that we did not breach any laws in this study.
Nevertheless, we believe that it is of utmost importance and responsibility to create a technology that can avoid the fraudulent use of deepfakes, which can go beyond celebrities and target the public in the long term with more harmful intents such as creating revenge porn, etc. As scholars, through our work, we should aim to mitigate any possible harm to ethics and privacy. We plan to anonymize the celebrity photographs used and blur them in the final version of our paper to minimize privacy risks.

\noindent
\textbf{Deployment. } We plan to build a REST API using our proposed defense mechanism that can be used with the commercial celebrity recognition APIs to classify incoming requests as real or deepfake. Also, we suggest combining our deepfake defense mechanism with AWS, MS, and NAV APIs. In this way, we can remove the vulnerability of face recognition APIs against deepfakes, which can mitigate the DI attacks.

\section{Conclusion}
\label{sec:conclusion}
%\simon{conclusion sounds like a summary...can u write better? SHA: u have written ample of conclusions...? can you rewrite?}

In this work, we warn about the potential danger of web-based commercial face recognition systems against Deepfake Impersonation (DI) attacks, demonstrating how unprepared these state-of-the-art systems are for the threats caused by deepfakes. We reveal the vulnerabilities of popular celebrity recognition APIs through impersonation attacks and present the results of extensive analyses of attack success rates and API robustness, using five deepfake datasets generated through three distinct methods. From experiments, we find that some deepfake generation methods are of greater threat to recognition systems than others and that each system reacts to DI attacks differently; moreover, we propose a practical defense mechanism to mitigate such attacks effectively. We believe our research findings can shed light on better designing robust web-based APIs as well as appropriate defense mechanisms, which are urgently needed to fight against malicious use of deepfakes.

\begin{acks}
We thank Sunwoo Jung and Siho Han for carefully reviewing this manuscript, which greatly improved the paper.
This work was supported by Institute for Information \& Communication Technology Planning \& Evaluation (IITP) grant funded by the Korea government (MSIT) (No. 2019-0-01343, Regional strategic industry convergence security core talent training business) and the Basic Science Research Program through National Research Foundation of Korea grant funded by MSIT (No.2020R1C1C1006004). Lastly, this research was supported by the Ministry of Science, ICT (MSIT), Korea, under the High-Potential Individuals Global Training Program (No. 2019-0-01579) supervised by the Institute for Information \& Communications Technology Planning \& Evaluation (IITP).
\end{acks}

\bibliographystyle{ACM-Reference-Format}
% \bibliography{ccs-sample}
\bibliography{references}
% \clearpage
\appendix
\section{Appendix: Experimental Settings}
\label{Appendix: Experimental Settings}
This appendix provides the training details of the deepfake detection methods that we used for the defense mechanism.

\subsection{Training Details}
In this section, we include the training details of each deepfake detection model used in our paper. The results in the paper are reported for the top-3 deepfake detectors for each dataset. Also, the images in the training and testing set have no overlap. To be consistent, we use the images inside the test set (see Table ~\ref{tab:DATASET}, column--4) for measuring the attack and defense performance.

\subsection{Baselines Methods for Defense}
We experiment with seven state-of-the-art deepfake detection methods. The description of these models is provided as:

\subsubsection{Xception} Fran{\c{c}}ois Chollet~\cite{Xception} proposes Xception, and it is one of the best models for face recognition and classification tasks. We use the Keras implementation of Xception.
    
\subsubsection{FakeSpotter} Wang et al.~\cite{FakeSpotter} propose FakeSpotter based on monitoring neuron behaviors to spot AI synthesized fake faces. We have tried our best to implement FakeSpotter according to the description from the original paper.
    
\subsubsection{CLRNet} Convolutional LSTM based methods have shown tremendous success in the time-series domain~\cite{tariq2019detecting,tariq2020cantransfer}. Tariq et al.~\cite{tariq2020convolutional,tariq2021web} propose a Convolutional LSTM based Residual Neural Network architecture to detect deepfake by utilizing the temporal and spatial information from videos. We have acquired the code for CLRNet from the authors and used it for training and testing.
    
\subsubsection{ABNet} Agarwal et al.~\cite{ABNet} proposed a neural network that uses appearance and behavioral biometrics to detect deepfakes. We tried our best to implement ABNet as the code is not publicly available online.
    
\subsubsection{ShallowNet} Tariq et al.~\cite{ShallowNet1,ShallowNet2} showed that it is possible to achieve near state-of-the-art performance with shallow and simple yet effective neural network architecture. We implemented ShallowNet according to the description from the original paper using Python v3.6.8 using TensorFlow v1.13.1 and used the Keras v2.2.4.
    
\subsubsection{MesoNet} Afchar et al.~\cite{MesoNet} proposed that using a low number of layers to focus on images' mesoscopic properties can achieve high performance on deepfake detection.  We used the MesoInception4 model, which is their best performer. We used the code provided by authors from their GitHub repository to implement MesoInception4. For simplicity, we will refer to MesoInception4 as MesoNet.
    
\subsubsection{FTDetect} Jeon et al.~\cite{FakeTalkerDetect} proposed reusable neural network architecture using pre-trained models with only a few real images for fine-tuning in Siamese networks to effectively detect the fake images. We used the code provided by authors from their GitHub repository to implement FTDetect. For simplicity, we will refer to FakeTalkerDetect as FTDetect.

\subsection{Machine Configuration}
We use Intel(R) Xeon(R) Silver 4114 CPU @ 2.20 GHz with 256.0 GB RAM and NVIDIA GeForce Titan RTX for training and inference of the deepfake detection models used in our defense mechanism.

\subsection{Evaluation}
The models predict the probability of a single or a group of consecutive frames being real or fake. We use Precision, Recall, and F1-Score for the evaluation. Due to space limitations, we are reporting only the F1-Scores in the main text. In some deepfake videos, only a few frames contain fake content. Therefore, we evaluate all the video frames. Furthermore, we kept the same number of real and fake images in the training, validation, and test sets to minimize the influence of data imbalance during evaluation.

\subsection{Preprocessing and Data Augmentation}
From each real and fake video, we extract a minimum of 16 samples. Each sample contains five consecutive frames ($16\times5=80$ images per video). We do this because some methods, such as CLRNet, require consecutive frames to extract temporal information and spatial information.  We use multi-task CNN (MTCNN)~\cite{MTCNN} to detect the face landmark information inside the extracted frames. Afterward, we use this landmark information to crop the face from the image and aligned it to the center. The average image size for all datasets after cropping the faces is $240\times240$. Therefore, we resize all the frames to a $240\times240$ resolution. We also utilize data augmentation methods to diversify the training data as follows: 

\begin{itemize}
    \item Rotation (-30$^{\circ}$ degrees to 30$^{\circ}$), Horizontal flip (50\% probability), Zoom (-20\% to 20\%), Brightness (-30\% to 30\%) and Channel shift (-50 to 50).
\end{itemize}

\section{Appendix: Additional Discussion on Results}
\label{Appendix: Additional Discussion on Results}
%In this section, we provide more discussion on the results and also present some interesting findings.
\subsection{Transferable to Other Face APIs}
We consider AWS, MS, and NAV API as a good representative of all other face recognition APIs. Therefore, by successfully demonstrating the high performance of DI attacks on those three APIs in a black-box setting, we can make this conjecture that similar performance can be achieved on other face recognition APIs. Furthermore, based on black-box success, we hypothesize that our deepfake impersonation attack can be equally or more effective in a white-box setting.

\subsection{Evaluating Other APIs}
Initially, we also have planned to evaluate Google's celebrity recognition API as well. However, it is under restricted access. The feature is intended to be used by media and entertainment companies or partners approved by those companies on professionally-produced media content. Therefore, our request to access this API was rejected. We also have experimented with the Alibaba cloud API. However, it only provides limited functionality that can only be used for content moderation. Therefore, we do not include Alibaba API to our use cases.

\subsection{Alternative Explanation to DI Attack vs. Adversarial Example Attack}
The goal of adversarial attacks is to increase \textit{false negatives} by misclassifying, for example, `real Obama' ($A$) to someone else ($B$), denoted as $C(A)\to C(B)$, where $C(\cdot)$ is the class label function. In contrast, the goal of DI attack is to misclassify `fake Obama' ($A^*$) to be `real Obama ($A$)', denoted as $C(A^*)\to C(A)$, where $A^*$ is created with images of A and a target video ($X$), producing false positives. It is important to note that $A^*=A \bigotimes X$ and $A^*!=A$,  where $\bigotimes$ is the abstraction of some deepfake generation method. Therefore, both attacks are different and can be used to evaluate and measure the model's robustness from different perspectives, i.e., false positive or false negative.

\begin{figure}[t!]
    \begin{subfigure}[t]{0.125\linewidth}
    \centering
    \includegraphics[width=1\linewidth]{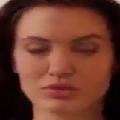}
    % \caption{}
    \end{subfigure}\hfill
    \begin{subfigure}[t]{0.125\linewidth}
    \centering
    \includegraphics[width=1\linewidth]{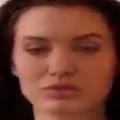}
    % \caption{}
    \end{subfigure}\hfill
    \begin{subfigure}[t]{0.125\linewidth}
    \centering
    \includegraphics[width=1\linewidth]{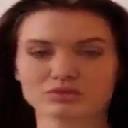}
    % \caption{}
    \end{subfigure}\hfill
    \begin{subfigure}[t]{0.125\linewidth}
    \centering
    \includegraphics[width=1\linewidth]{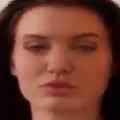}
    % \caption{}
    \end{subfigure}\hfill
    \begin{subfigure}[t]{0.125\linewidth}
    \centering
    \includegraphics[width=1\linewidth]{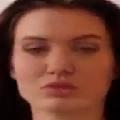}
    % \caption{}
    \end{subfigure}\hfill
    \begin{subfigure}[t]{0.125\linewidth}
    \centering
    \includegraphics[width=1\linewidth]{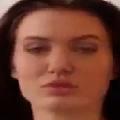}
    % \caption{}
    \end{subfigure}\hfill
    \begin{subfigure}[t]{0.125\linewidth}
    \centering
    \includegraphics[width=1\linewidth]{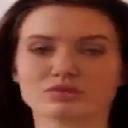}
    % \caption{}
    \end{subfigure}\hfill
    \begin{subfigure}[t]{0.125\linewidth}
    \centering
    \includegraphics[width=1\linewidth]{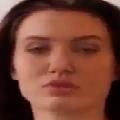}
    % \caption{}
    \end{subfigure}%\hfill
    % \begin{subfigure}[t]{0.20\linewidth}
    % \centering
    % \includegraphics[width=1\linewidth]{Appendix_Figures/Consecutive/00131_0.jpg}
    % \caption{}
    % \end{subfigure}%
    \caption{Eight consecutive frames from a deepfake video of Angelina Jolie, each given one by one to the AWS API, resulted in avg. prediction confidence of 85.4\%. This score is very close to the avg. prediction score of 85\% for Angelina Jolie on her stills (i.e., single images), as shown in Table~\ref{tab:AppendixCelebDFDataset}.}
    \label{fig:Angelina_appendix}
    \vspace{-10pt}
\end{figure}

\begin{figure}[t!]
    \begin{subfigure}[t]{0.2\linewidth}
    \centering
    \includegraphics[width=1\linewidth]{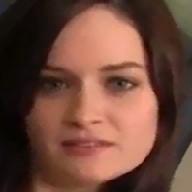}
    % \caption{}
    \end{subfigure}%\hfill
    \begin{subfigure}[t]{0.2\linewidth}
    \centering
    \includegraphics[width=1\linewidth]{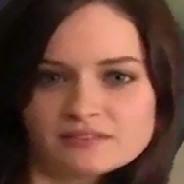}
    % \caption{}
    \end{subfigure}%\hfill
    \begin{subfigure}[t]{0.2\linewidth}
    \centering
    \includegraphics[width=1\linewidth]{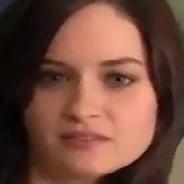}
    % \caption{}
    \end{subfigure}%\hfill
    \begin{subfigure}[t]{0.2\linewidth}
    \centering
    \includegraphics[width=1\linewidth]{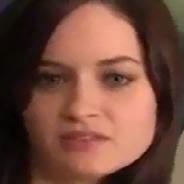}
    % \caption{}
    \end{subfigure}
    % \hfill
    % \begin{subfigure}[t]{0.20\linewidth}
    % \centering
    % \includegraphics[width=1\linewidth]{Appendix_Figures/Consecutive/00040_0.jpg}
    % \caption{}
    % \end{subfigure}\hfill
    % \begin{subfigure}[t]{0.20\linewidth}
    % \centering
    % \includegraphics[width=1\linewidth]{Appendix_Figures/Consecutive/00042_0.jpg}
    % \caption{}
    % \end{subfigure}\hfill
    % \begin{subfigure}[t]{0.20\linewidth}
    % \centering
    % \includegraphics[width=1\linewidth]{Appendix_Figures/Consecutive/00044_0.jpg}
    % \caption{}
    % \end{subfigure}\hfill
    % \begin{subfigure}[t]{0.20\linewidth}
    % \centering
    % \includegraphics[width=1\linewidth]{Appendix_Figures/Consecutive/00046_0.jpg}
    % \caption{}
    % \end{subfigure}%
    \caption{Four consecutive frames from a deepfake video of Alexis Bledel, each given one by one to the MS API, resulted in avg. prediction of 86.3\%. This score is very close to the avg. prediction score of 85\% for Alexis Bledel on her stills.}
    \label{fig:Alexis_appendix}
    \vspace{-10pt}
\end{figure}

% \clearpage

\onecolumn
\section{Appendix: Interesting Scenario}
\label{Appendix: Interesting Scenario}

\begin{figure*}[b!]
    \begin{subfigure}[t]{0.75\linewidth}
    \centering
    \includegraphics[width=1\linewidth]{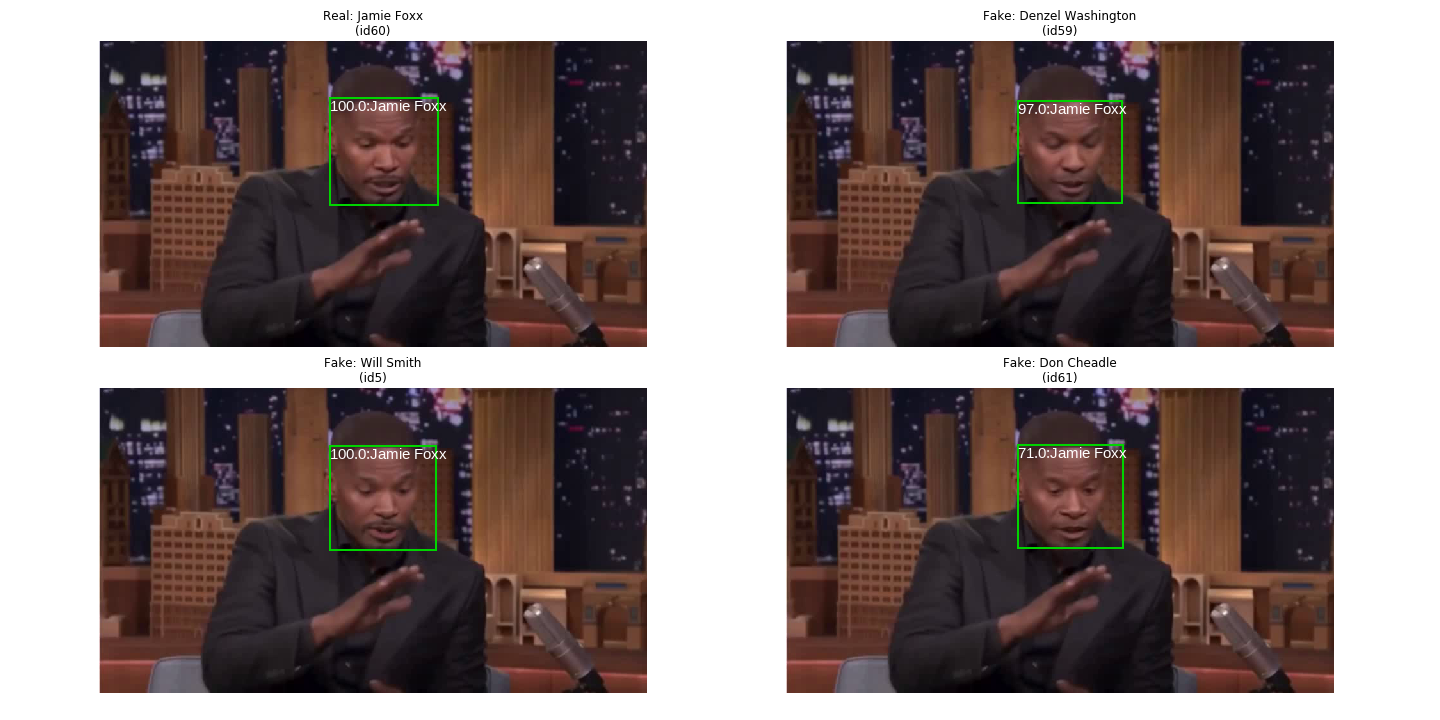}
    \caption{\underline{Top-left:} Image from a real video of ``Jamie Foxx'' is predicted as ``\textit{Jamie Foxx}'' with 100\% prediction confidence. \underline{Top-right:} Deepfake of ``Denzel Washington'' generated from the original video of ``Jamie Foxx'' is predicted as \textit{``Jamie Foxx''} with 97\% prediction confidence. \underline{Bottom-left:} Deepfake of ``Will Smith'' generated from the original video of ``Jamie Foxx'' is predicted as \textit{``Jamie Foxx''} with 100\% prediction confidence. \underline{Bottom-right:} Deepfake of ``Don Cheadle'' generated from the original video of ``Jamie Foxx'' is predicted as \textit{``Jamie Foxx''} with 71\% prediction confidence.}
    \end{subfigure}\hfill
    \begin{subfigure}[t]{0.75\linewidth}
    \centering
    \includegraphics[width=1\linewidth]{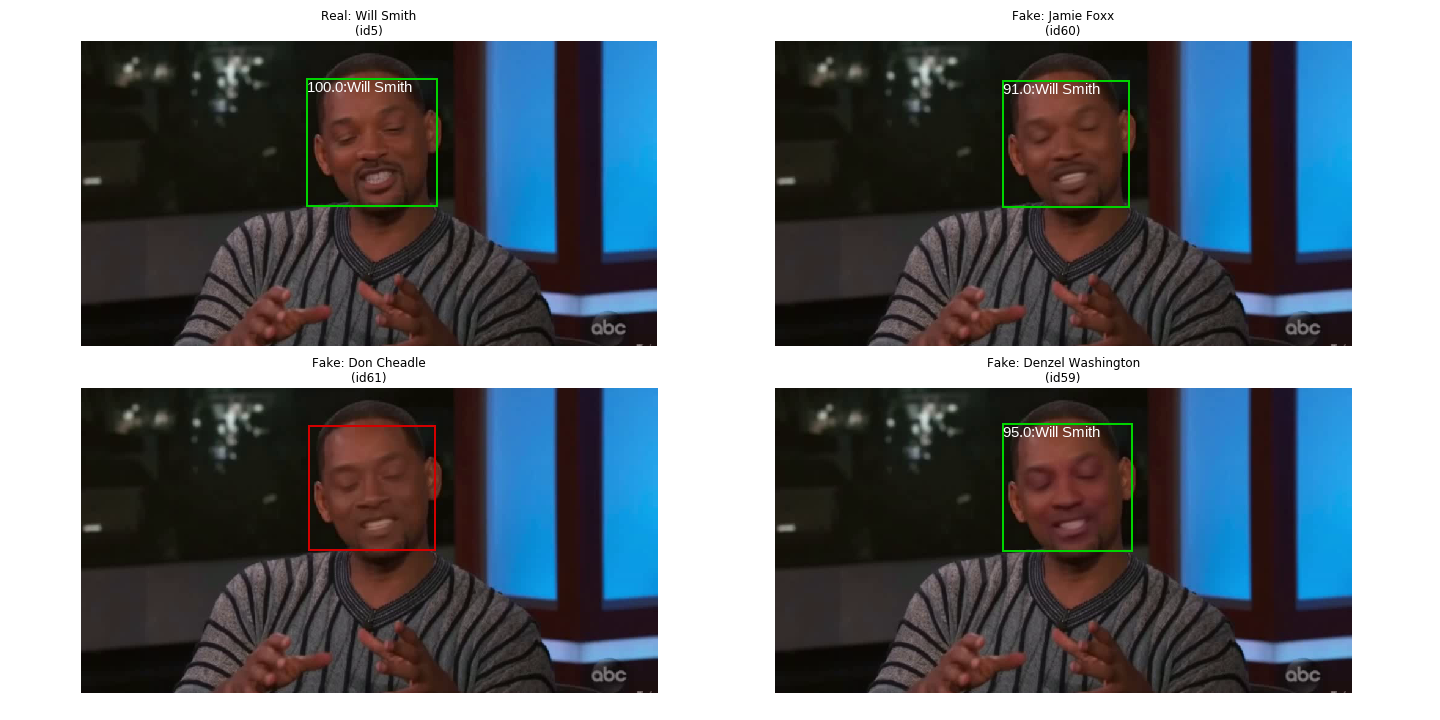}
    \caption{\underline{Top-left:} Image from a real video of ``Will Smith'' is predicted as ``\textit{Will Smith}'' with 100\% prediction confidence. \underline{Top-right:} Deepfake of ``Jamie Foxx'' generated from the original video of ``Will Smith'' is predicted as \textit{``Will Smith''} with 91\% prediction confidence. \underline{Bottom-left:} Deepfake of ``Don Cheadle'' generated from the original video of ``Will Smith'' is predicted as \textit{``Non-celebrity''}. \underline{Bottom-right:} Deepfake of ``Denzel Washingtion'' generated from the original video of ``Will Smith'' is predicted as \textit{``Will Smith''} with 95\% prediction confidence.}
    \end{subfigure}%
    \caption{\underline{The Curious Case of Jamie Foxx and Will Smith:} We found that the deepfakes for most of the ``Black Celebrities'' in the CelebDF dataset are predicted as the same celebrity from the original video (i.e., Jamie Foxx for the figure (a) and Will Smith for the figure (b), respectively). We are not sure about the exact reason for such behavior. However, we hypothesize that this could be possible because of either or both these two reasons: 1) Racial bias in these face recognition APIs and 2) The generation capability of the deepfake generation method for CelebDF has a bias towards
    ``White Celebrities'' as they are the majority group in that dataset. Nevertheless, it would be interesting to perform a thorough study on this issue.}
    \label{fig:JamieFoxx_WillSmith_appendix_2}
\end{figure*}

\section{Appendix: Additional Results for Amazon API}
\label{Appendix: Additional Results for Amazon API}

\begin{figure*}[hbt!]
    \begin{subfigure}[t]{0.75\linewidth}
    \centering
    \includegraphics[width=1\linewidth]{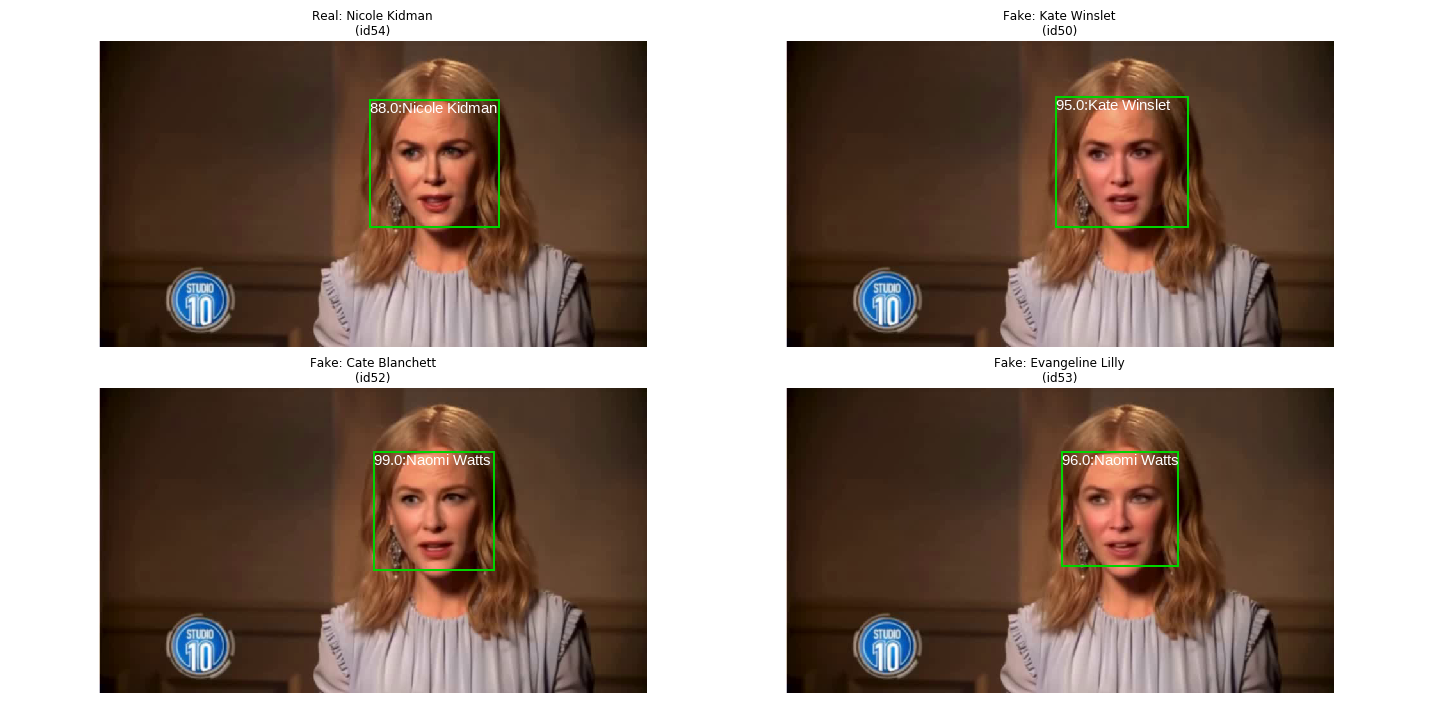}
    \caption{\underline{Top-left:} Image from a real video of ``Nicole Kidman'' is predicted as ``\textit{Nicole Kidman}'' with 88\% prediction confidence. \underline{Top-right:} Deepfake of ``Kate Winslet'' generated from the original video of ``Nicole Kidman'' is predicted as \textit{``Kate Winslet''} with 95\% prediction confidence (Successful Targeted Attack i.e., when $C(\mathcal{X}_\mathcal{D})=C(\mathcal{X}_\mathcal{T})$). \underline{Bottom-left:} Deepfake of ``Cate Blanchett'' generated from the original video of ``Nicole Kidman'' is predicted as \textit{``Naomi Watts''} with 99\% prediction confidence (Successful Non-targeted Attack i.e., when $C(\mathcal{X}_\mathcal{D})\in \mathbb{C}$). \underline{Bottom-right:} Deepfake of ``Evangeline Lilly'' generated from the original video of ``Nicole Kidman'' is predicted as \textit{``Naomi Watts''} with 96\% prediction confidence (Successful Non-targeted Attack i.e., when $C(\mathcal{X}_\mathcal{D})\in \mathbb{C}$). \underline{NOTE:} ``\textit{All these deepfakes have higher prediction confidence than the real image of Nicole Kidman}''.}
    \end{subfigure}\hfill
    \begin{subfigure}[t]{0.75\linewidth}
    \centering
    \includegraphics[width=1\linewidth]{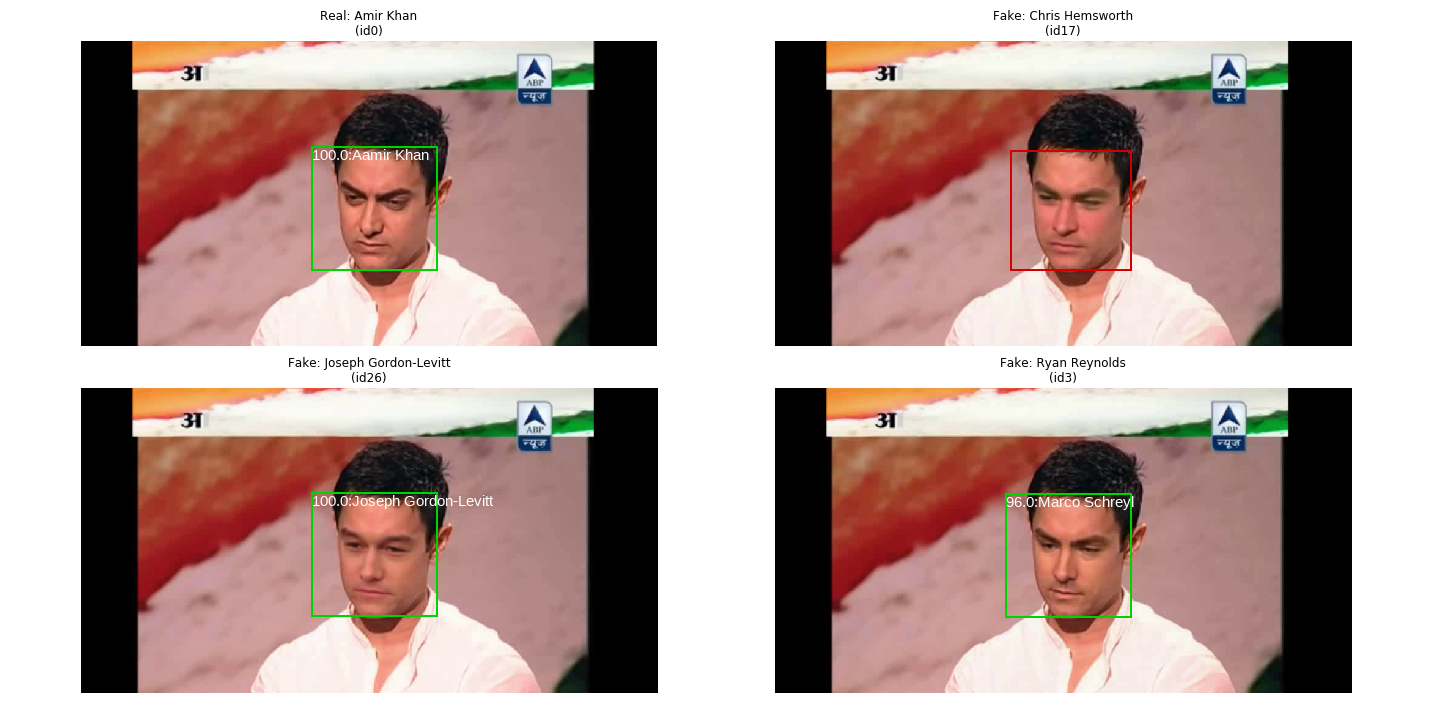}
    \caption{\underline{Top-left:} Image from a real video of ``Aamir Khan'' is predicted as ``\textit{Aamir Khan}'' with 100\% prediction confidence. \underline{Top-right:} Deepfake of ``Chris Hemsworth'' generated from the original video of ``Aamir Khan'' is predicted as \textit{``Non-Celebrity''} by the API. \underline{Bottom-left:} Deepfake of ``Joseph Gordon-Levitt'' generated from the original video of ``Aamir Khan'' is predicted as \textit{``Joseph Gordon-Levitt''} with 100\% prediction confidence (Successful Targeted Attack i.e., when $C(\mathcal{X}_\mathcal{D})=C(\mathcal{X}_\mathcal{T})$). \underline{Bottom-right:} Deepfake of ``Ryan Reynolds'' generated from the original video of ``Aamir Khan'' is predicted as \textit{``Ryan Reynolds''} with 96\% prediction confidence (Successful Non-targeted Attack i.e., when $C(\mathcal{X}_\mathcal{D})\in \mathbb{C}$).}
    \end{subfigure}%
    \caption{Additional results showing the ``\underline{success of deepfake impersonation attack}'' on \textit{Amazon Rekognition API} using CelebDF dataset.}
    \label{fig:CelebDF_appendix_1}
\end{figure*}

\section{Appendix: Additional Results for Microsoft API}
\label{Appendix: Additional Results for Microsoft API}

\begin{figure*}[hbt!]
    \begin{subfigure}[t]{0.75\linewidth}
    \centering
    \includegraphics[width=1\linewidth]{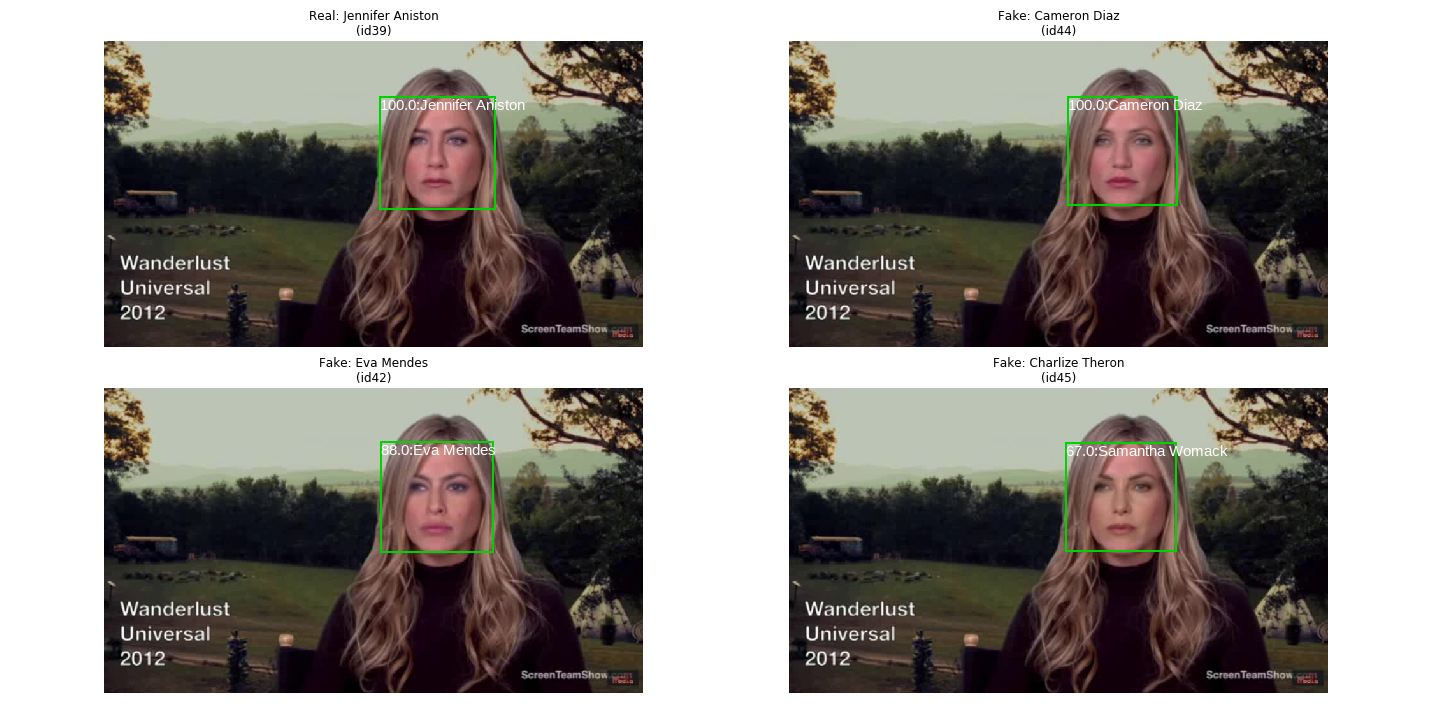}
    \caption{\underline{Top-left:} Image from a real video of ``Jennifer Aniston'' is predicted as ``\textit{Jennifer Aniston}'' with 100\% prediction confidence. \underline{Top-right:} Deepfake of ``Cameron Diaz'' generated from the original video of ``Jennifer Aniston'' is predicted as \textit{``Cameron Diaz''} with 100\% prediction confidence (Successful Targeted Attack i.e., when $C(\mathcal{X}_\mathcal{D})=C(\mathcal{X}_\mathcal{T})$). \underline{Bottom-left:} Deepfake of ``Eva Mendes'' generated from the original video of ``Jennifer Aniston'' is predicted as \textit{``Eva Mendes''} with 88\% prediction confidence (Successful Targeted Attack i.e., when $C(\mathcal{X}_\mathcal{D})=C(\mathcal{X}_\mathcal{T})$). \underline{Bottom-right:} Deepfake of ``Charlize Theron'' generated from the original video of ``Jennifer Aniston'' is predicted as \textit{``Samantha Womack''} with 67\% prediction confidence (Successful Non-targeted Attack i.e., when $C(\mathcal{X}_\mathcal{D})\in \mathbb{C}$).}
    \end{subfigure}\hfill
    \begin{subfigure}[t]{0.75\linewidth}
    \centering
    \includegraphics[width=1\linewidth]{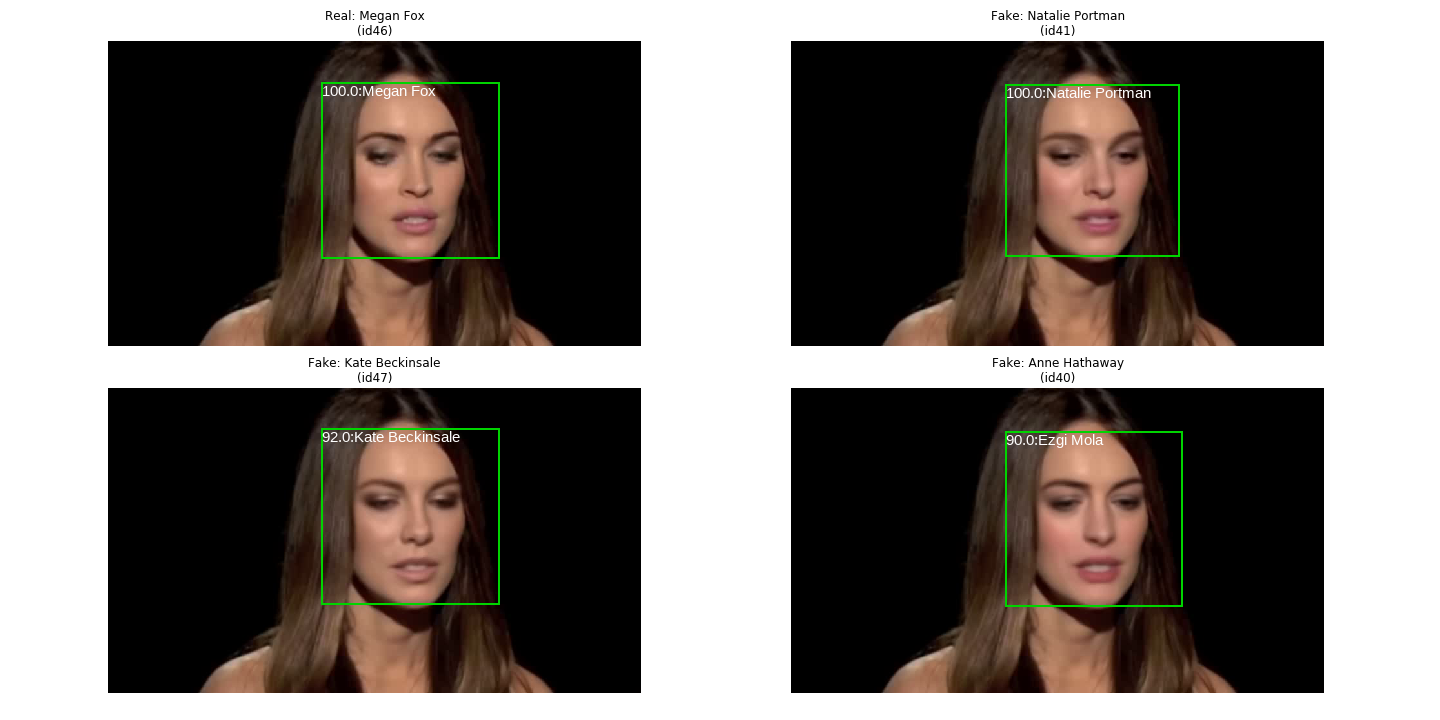}
    \caption{\underline{Top-left:} Image from a real video of ``Megan Fox'' is predicted as ``\textit{Megan Fox}'' with 100\% prediction confidence. \underline{Top-right:} Deepfake of ``Natalie Portman'' generated from the original video of ``Megan Fox'' is predicted as \textit{``Natalie Portman''} with 100\% prediction confidence (Successful Targeted Attack i.e., when $C(\mathcal{X}_\mathcal{D})=C(\mathcal{X}_\mathcal{T})$). \underline{Bottom-left:} Deepfake of ``Kate Beckinsale'' generated from the original video of ``Megan Fox'' is predicted as \textit{``Kate Beckinsale''} with 92\% prediction confidence (Successful Targeted Attack i.e., when $C(\mathcal{X}_\mathcal{D})=C(\mathcal{X}_\mathcal{T})$). \underline{Bottom-right:} Deepfake of ``Anne Hathaway'' generated from the original video of ``Megan Fox'' is predicted as \textit{``Ezgi Mola''} with 90\% prediction confidence (Successful Non-targeted Attack i.e., when $C(\mathcal{X}_\mathcal{D})\in \mathbb{C}$). \underline{NOTE:} ``\textit{All these deepfakes have a very high prediction confidence}''.}
    \end{subfigure}%
    \caption{Additional results showing the ``\underline{success of deepfake impersonation attack}'' on \textit{Microsoft Celebrity Recognition API} using CelebDF dataset.}
    \label{fig:CelebDF_appendix_2}
\end{figure*}

\section{Appendix: Comparing same reference with multiple Targets in CelebBlend}
\label{Appendix: Comparing same reference}

\begin{figure*}[hbt!]
    \begin{subfigure}[t]{0.24\linewidth}
        \centering
        \includegraphics[width=\linewidth]{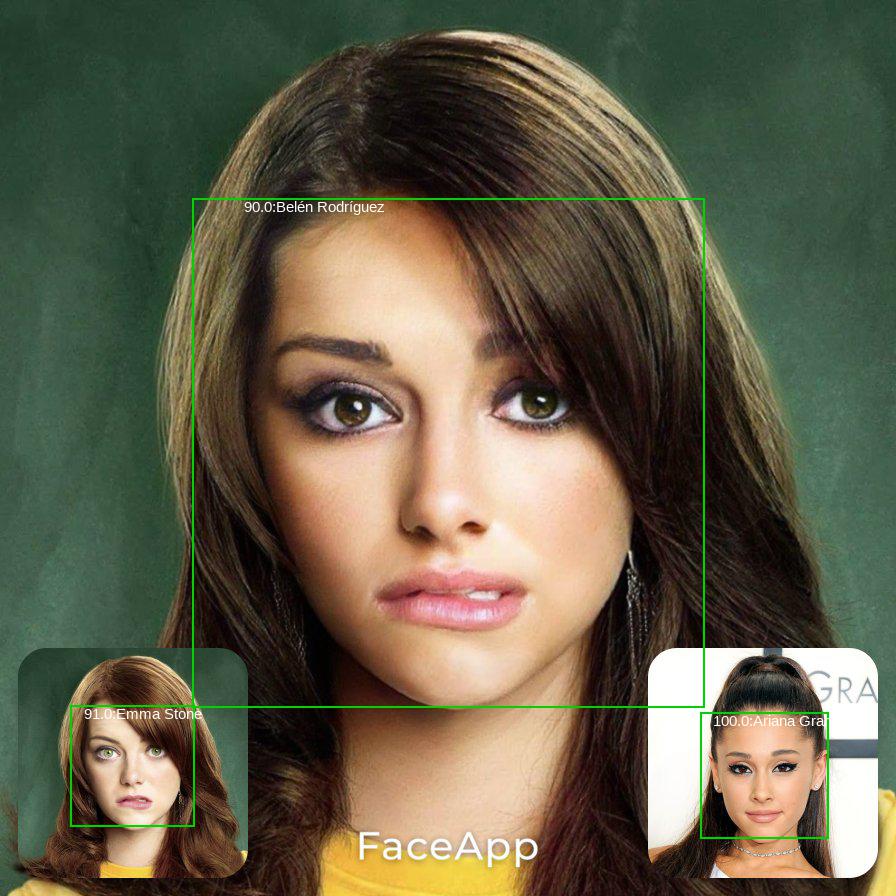}
        \caption{Emma Stone blend with Ariana Grande is predicted as Belén Rodríguez with 90\% prediction confidence (Successful Non-targeted Attack i.e., when $C(\mathcal{X}_\mathcal{D})\in \mathbb{C}$).}
    \end{subfigure}\hfill
    \begin{subfigure}[t]{0.24\linewidth}
        \centering
        \includegraphics[width=\linewidth]{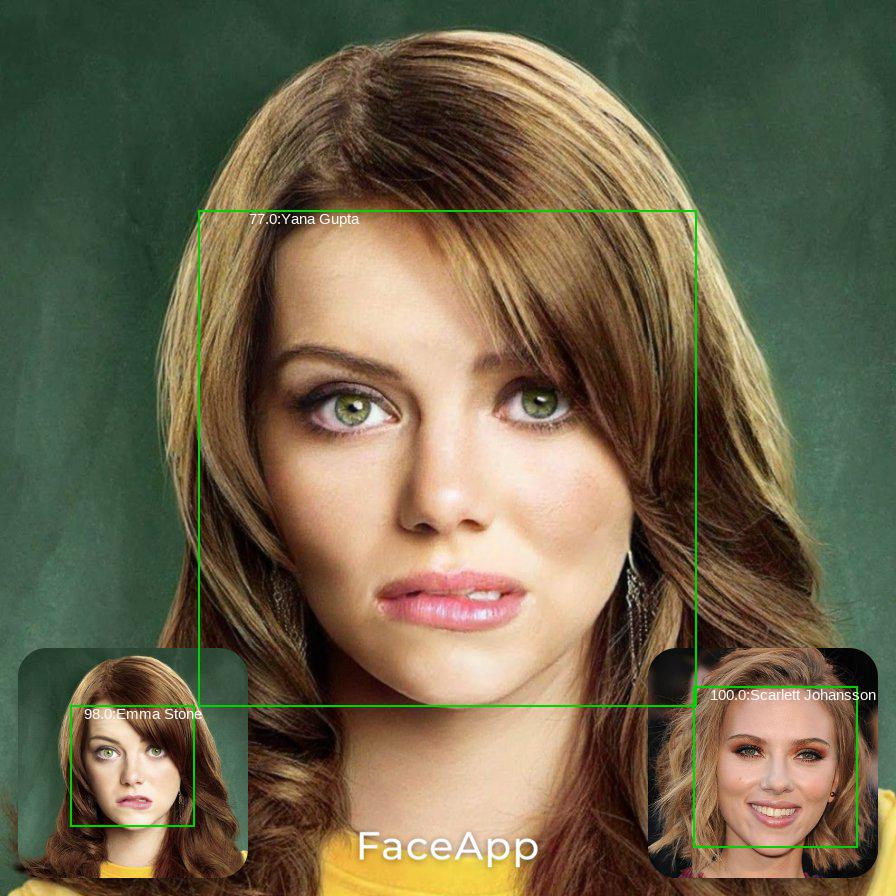}
        \caption{Emma Stone blend with Scarlett Johansson is predicted as Yana Gupta with 77\% prediction confidence (Successful Non-targeted Attack i.e., when $C(\mathcal{X}_\mathcal{D})\in \mathbb{C}$).}
    \end{subfigure}\hfill
    \begin{subfigure}[t]{0.24\linewidth}
        \centering
        \includegraphics[width=\linewidth]{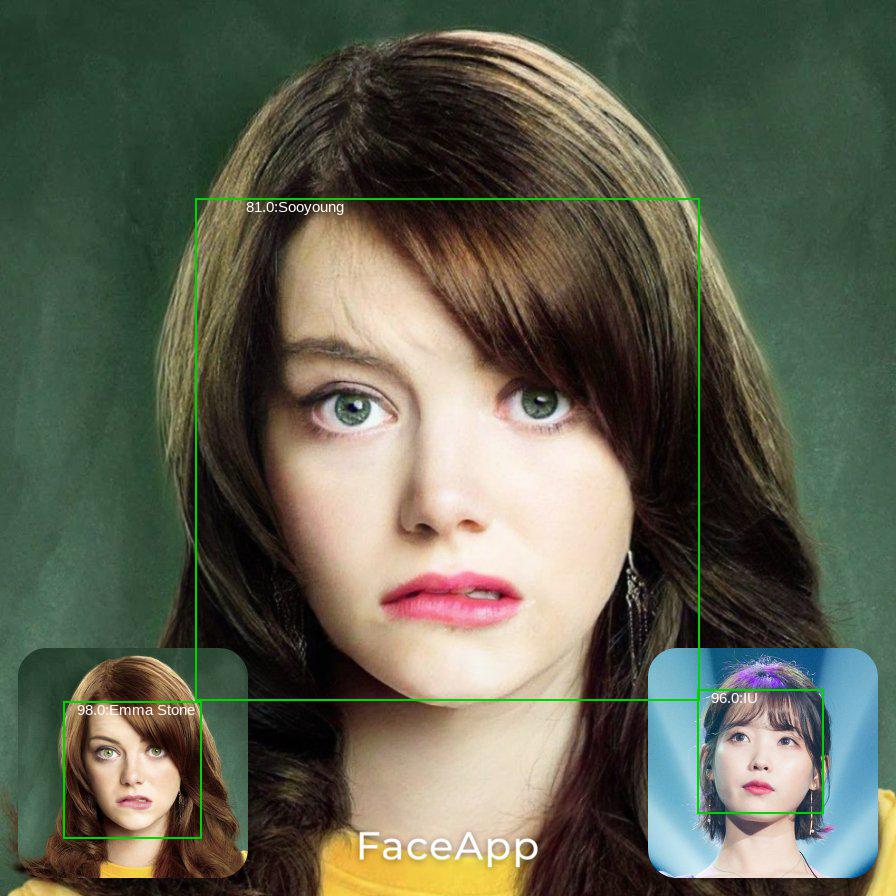}
        \caption{Emma Stone blend with Lee Ji-eun (IU) is predicted as Choi Sooyoung with 81\% prediction confidence (Successful Non-targeted Attack i.e., when $C(\mathcal{X}_\mathcal{D})\in \mathbb{C}$).}
    \end{subfigure}\hfill
    \begin{subfigure}[t]{0.24\linewidth}
        \centering
        \includegraphics[width=\linewidth]{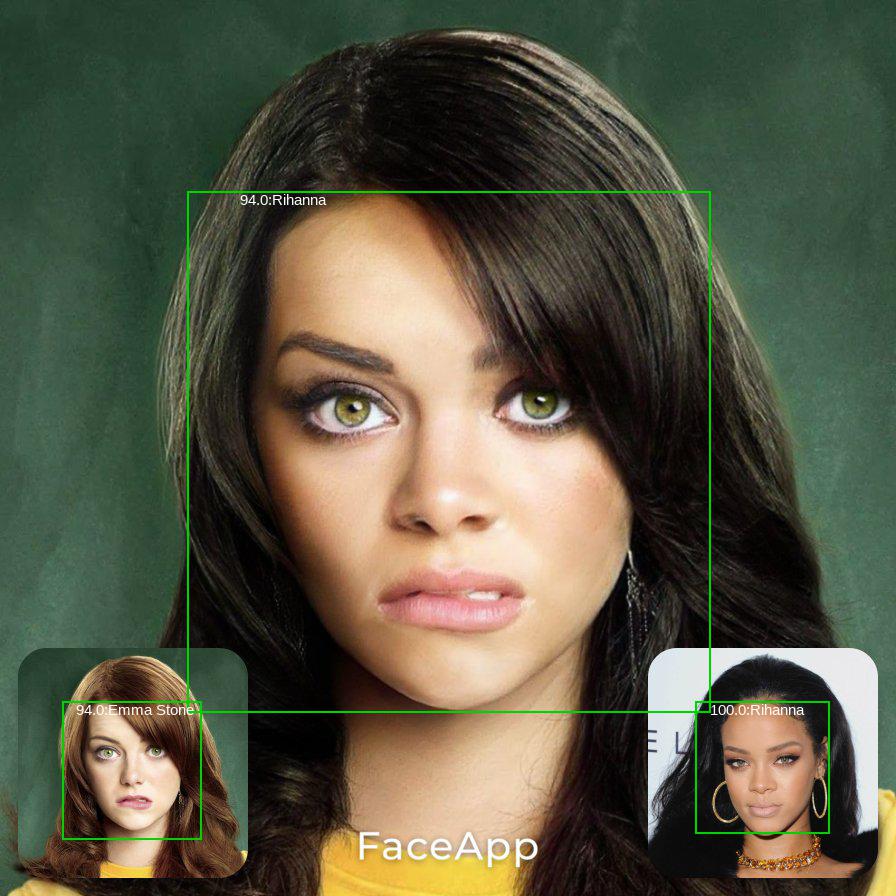}
        \caption{Emma Stone blend with Robyn Rihanna is predicted as Robyn Rihanna with 94\% prediction confidence (Successful Targeted Attack i.e., when $C(\mathcal{X}_\mathcal{D})=C(\mathcal{X}_\mathcal{T})$).}
    \end{subfigure}\hfill
    
    \begin{subfigure}[t]{0.24\linewidth}
        \centering
        \includegraphics[width=\linewidth]{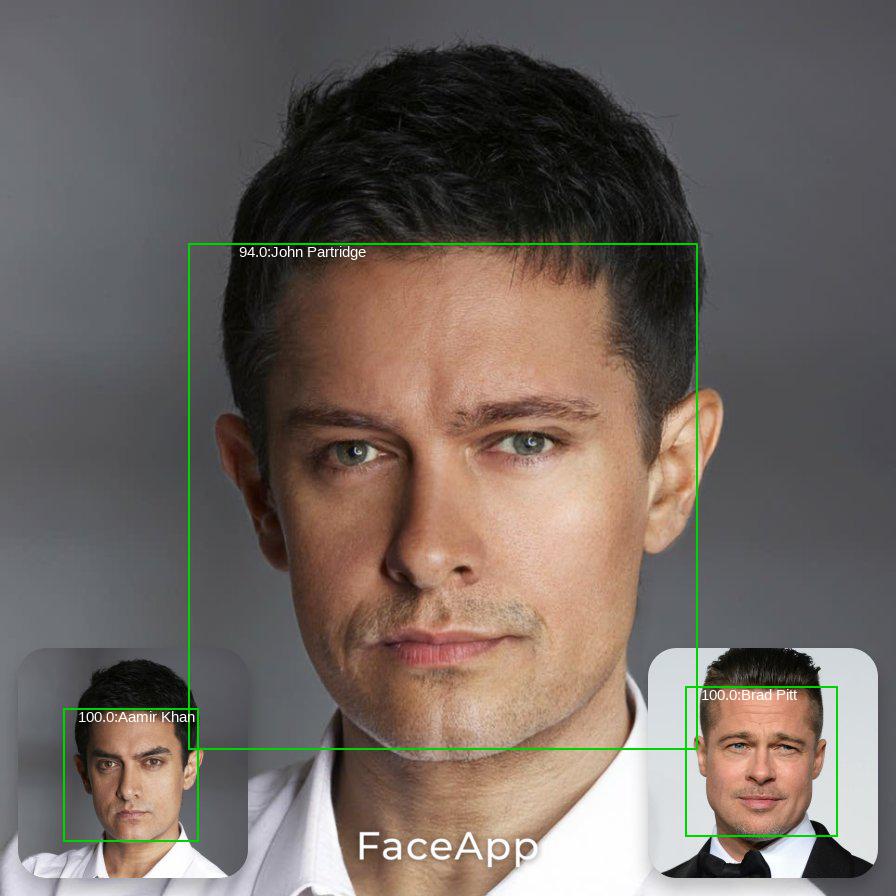}
        \caption{Aamir Khan blend with Brad Pitt is predicted as John Partridge with 94\% prediction confidence (Successful Non-targeted Attack i.e., when $C(\mathcal{X}_\mathcal{D})\in \mathbb{C}$).}
    \end{subfigure}\hfill
    \begin{subfigure}[t]{0.24\linewidth}
        \centering
        \includegraphics[width=\linewidth]{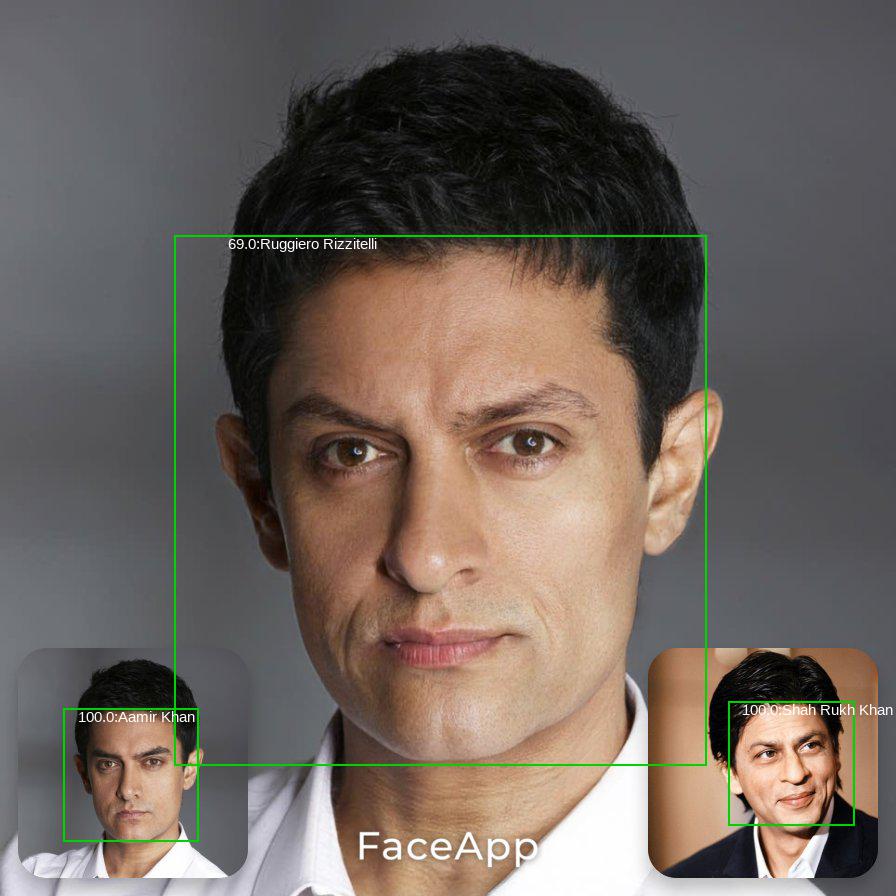}
        \caption{Aamir Khan blend with Shahrukh Khan is predicted as Ruggiero Rizzitelli with 69\% confidence (Successful Non-targeted Attack i.e., when $C(\mathcal{X}_\mathcal{D})\in \mathbb{C}$).}
    \end{subfigure}\hfill
    \begin{subfigure}[t]{0.24\linewidth}
        \centering
        \includegraphics[width=\linewidth]{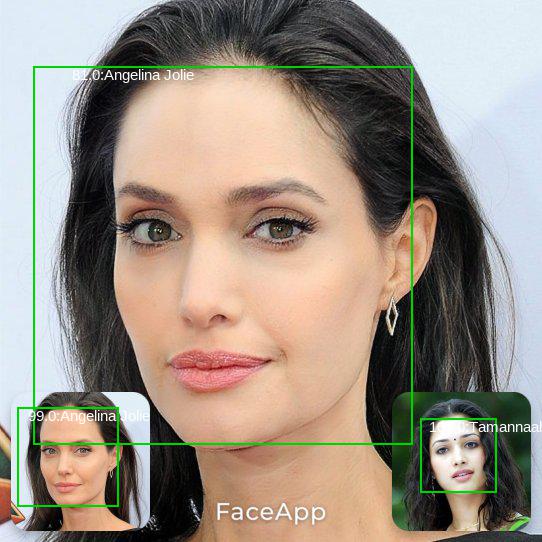}
        \caption{Angelina Jolie blend with Tamannaah is predicted as Angelina Jolie with 81\% confidence (Successful Non-targeted Attack i.e., when $C(\mathcal{X}_\mathcal{D})\in \mathbb{C}$).}
    \end{subfigure}\hfill
    \begin{subfigure}[t]{0.24\linewidth}
        \centering
        \includegraphics[width=\linewidth]{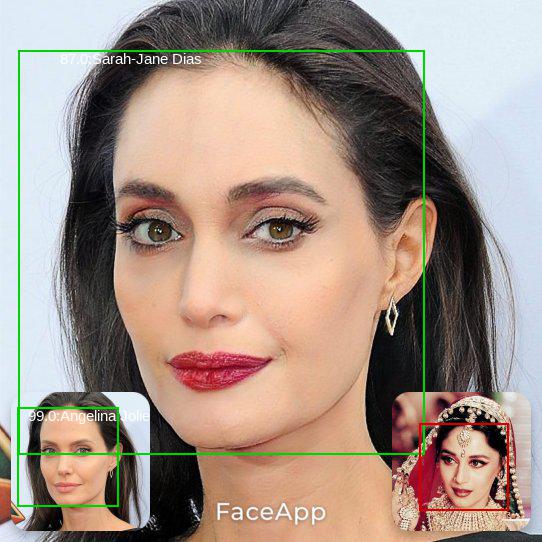}
        \caption{Angelina Jolie blend with Madhuri Dixit is predicted as SarahJane Dias with 87\% prediction confidence (Successful Non-targeted Attack i.e., when $C(\mathcal{X}_\mathcal{D})\in \mathbb{C}$).}
    \end{subfigure}\hfill
    
    \begin{subfigure}[t]{0.24\linewidth}
        \centering
        \includegraphics[width=\linewidth]{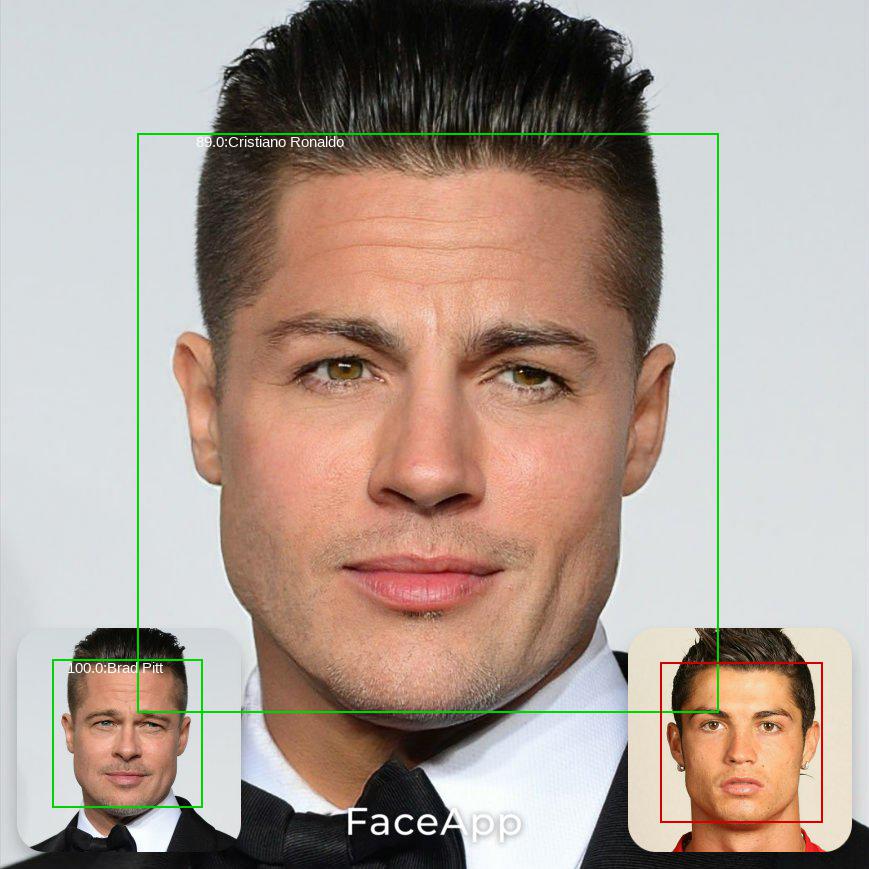}
        \caption{Brad Pitt blend with Cristiano Ronaldo  is predicted as Cristiano Ronaldo with 89\% prediction confidence (Successful Targeted Attack i.e., when $C(\mathcal{X}_\mathcal{D})=C(\mathcal{X}_\mathcal{T})$).}
    \end{subfigure}\hfill
    \begin{subfigure}[t]{0.24\linewidth}
        \centering
        \includegraphics[width=\linewidth]{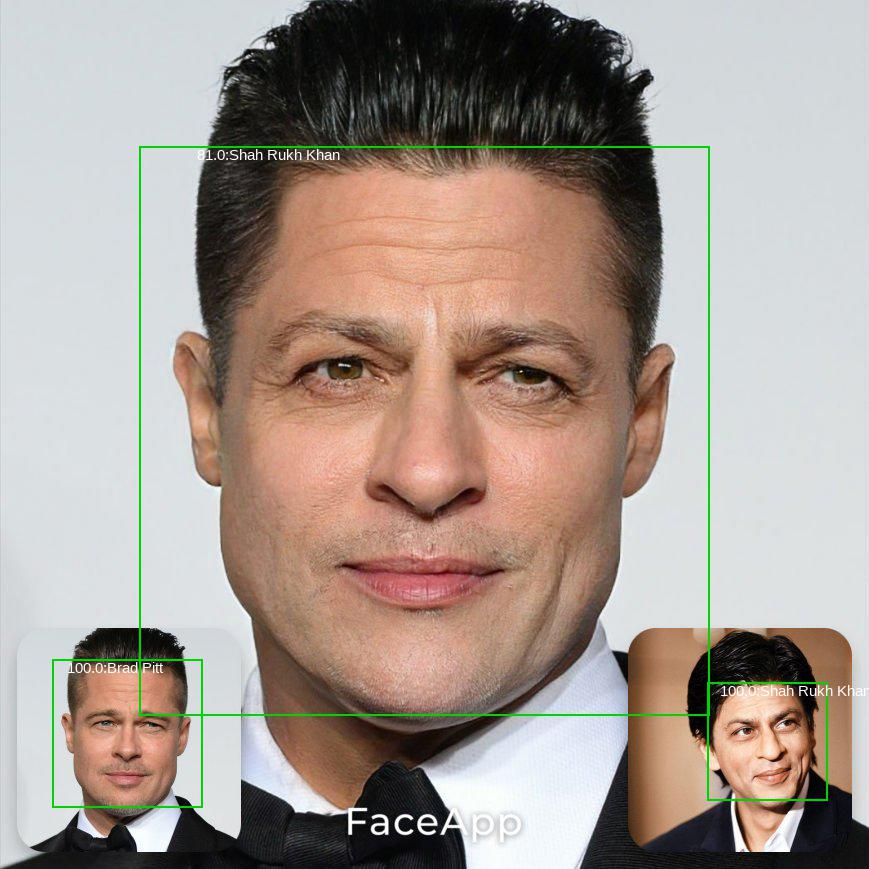}
        \caption{Brad Pitt blend with Shahrukh Khan is predicted as Shahrukh Khan with 81\% prediction confidence (Successful Targeted Attack i.e., when $C(\mathcal{X}_\mathcal{D})=C(\mathcal{X}_\mathcal{T})$).}
    \end{subfigure}\hfill
    \begin{subfigure}[t]{0.24\linewidth}
        \centering
        \includegraphics[width=\linewidth]{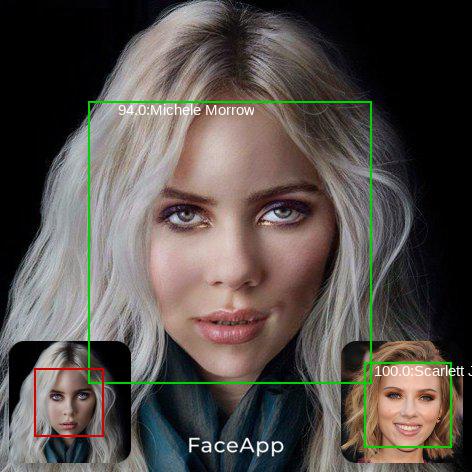}
        \caption{Billie Eilish blend with Scarlett Johansson is predicted as Michele Morrow with 94\% prediction confidence (Successful Non-targeted Attack i.e., when $C(\mathcal{X}_\mathcal{D})\in \mathbb{C}$).}
    \end{subfigure}\hfill
    \begin{subfigure}[t]{0.24\linewidth}
        \centering
        \includegraphics[width=\linewidth]{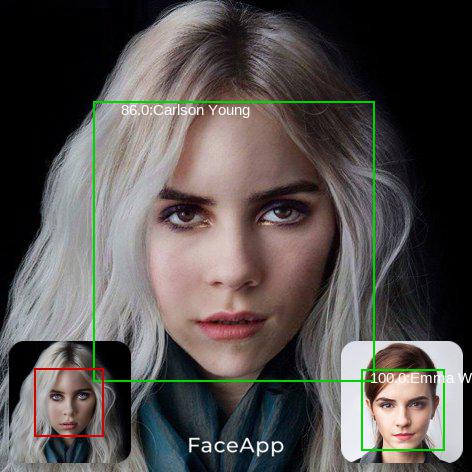}
        \caption{Billie Eilish blend with Emma Watson is predicted as Carlson Young with 86\% prediction confidence (Successful Non-targeted Attack i.e., when $C(\mathcal{X}_\mathcal{D})\in \mathbb{C}$).}
    \end{subfigure}%
    \caption{Additional results comparing similar images from CelebBlend dataset: The blended images have partial facial features from the source and reference celebrities. The DNN used by these APIs may predict the resultant image based on the closeness of its facial features to some celebrity.}
    \label{fig:CelebBlend_appendix_1}
\end{figure*}

\section{Appendix: Additional Results on CelebBlend using All three APIs.}
\label{Appendix: Additional Results on CelebBlend}
\begin{figure*}[hbt!]
    \begin{subfigure}[t]{0.32\linewidth}
        \centering
        \includegraphics[width=0.8\linewidth]{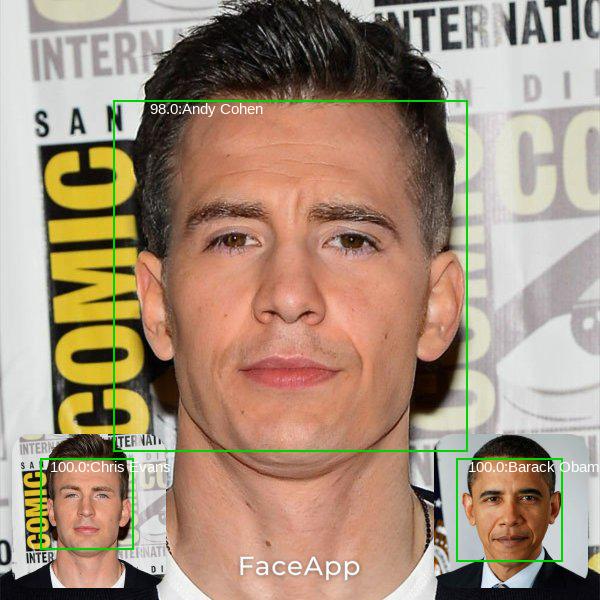}
        \caption{``Chris Evans'' blend with ``Barack Obama'' is predicted as ``Andy Cohen'' with 98\% prediction confidence (Successful Non-targeted Attack i.e., when $C(\mathcal{X}_\mathcal{D})\in \mathbb{C}$).}
    \end{subfigure}\hfill
    \begin{subfigure}[t]{0.32\linewidth}
        \centering
        \includegraphics[width=0.8\linewidth]{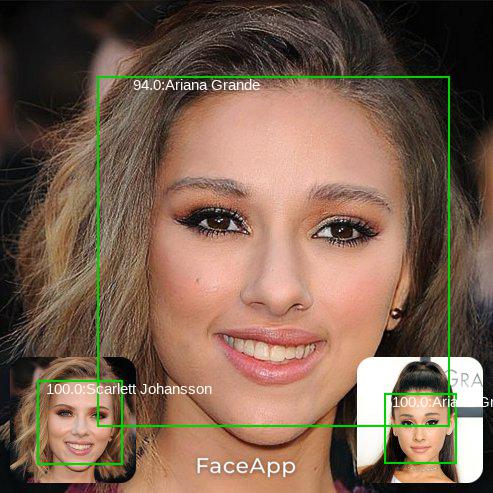}
        \caption{``Scarlett Johansson'' blend with ``Ariana Grande'' is predicted as ``Ariana Grande'' with 94\% prediction confidence (Successful Targeted Attack i.e., when $C(\mathcal{X}_\mathcal{D})=C(\mathcal{X}_\mathcal{T})$).}
    \end{subfigure}\hfill
    \begin{subfigure}[t]{0.32\linewidth}
        \centering
        \includegraphics[width=0.8\linewidth]{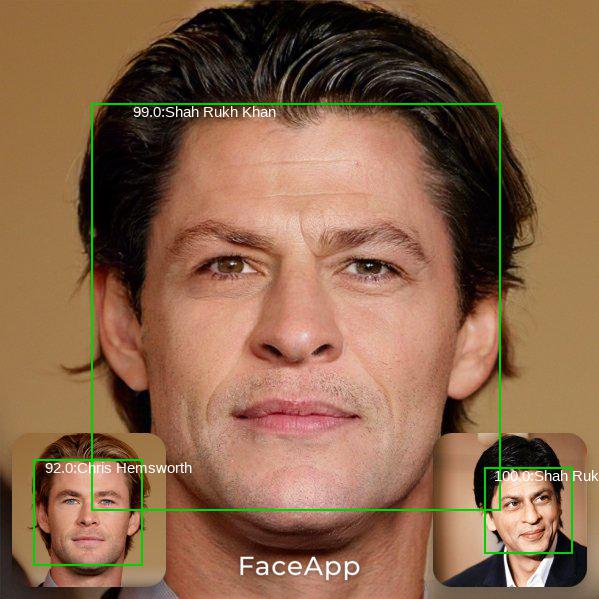}
        \caption{``Chris Hemsworth'' blend with ``ShahRukh Khan'' is predicted as ``ShahRukh Khan'' with 99\% confidence (Successful Targeted Attack i.e., $C(\mathcal{X}_\mathcal{D})=C(\mathcal{X}_\mathcal{T})$).}
    \end{subfigure}\hfill
    
    \begin{subfigure}[t]{0.32\linewidth}
        \centering
        \includegraphics[width=0.8\linewidth]{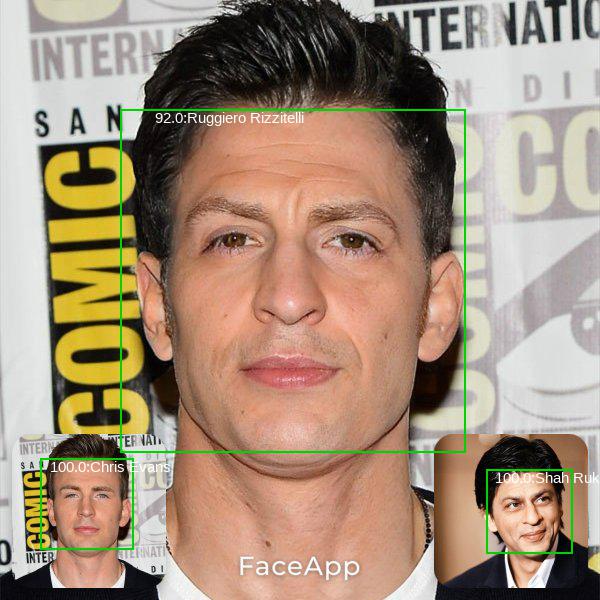}
        \caption{``Chris Evans'' blend with ``ShahRukh Khan'' is predicted as ``Ruggiero Rizzitelli'' with 92\% prediction confidence (Successful Non-targeted Attack i.e., when $C(\mathcal{X}_\mathcal{D})\in \mathbb{C}$).}
    \end{subfigure}\hfill
    \begin{subfigure}[t]{0.32\linewidth}
        \centering
        \includegraphics[width=0.8\linewidth]{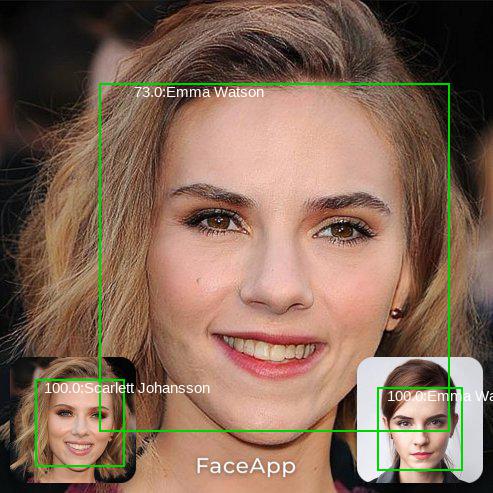}
        \caption{``Scarlett Johansson'' blend with ``Emma Watson'' is predicted as ``Emma Watson'' with 73\% prediction confidence (Successful Targeted Attack i.e., when $C(\mathcal{X}_\mathcal{D})=C(\mathcal{X}_\mathcal{T})$).}
    \end{subfigure}\hfill
    \begin{subfigure}[t]{0.32\linewidth}
        \centering
        \includegraphics[width=0.8\linewidth]{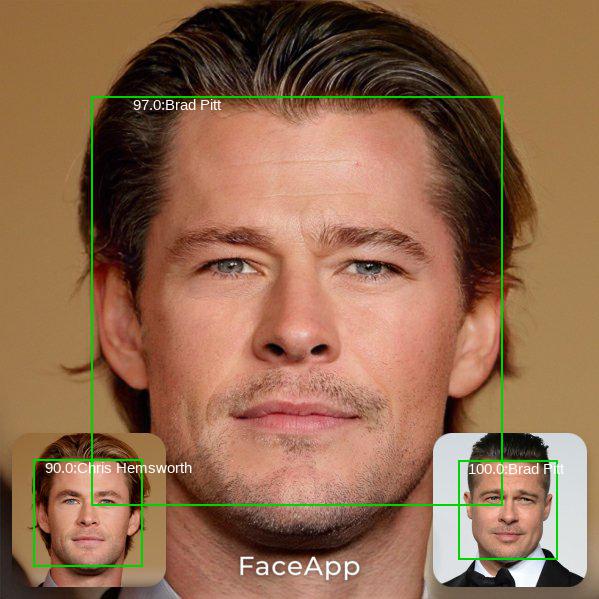}
        \caption{``Chris Hemsworth'' blend with ``Brad Pitt'' is predicted as ``Brad Pitt'' with 97\% prediction confidence (Successful Targeted Attack i.e., when $C(\mathcal{X}_\mathcal{D})=C(\mathcal{X}_\mathcal{T})$).}
    \end{subfigure}\hfill
    
    \begin{subfigure}[t]{0.32\linewidth}
        \centering
        \includegraphics[width=0.8\linewidth]{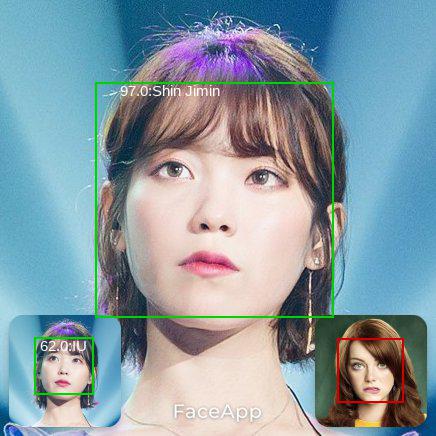}
        \caption{``IU'' blend with ``Emma Stone'' is predicted as ``Shin Jimin'' with 97\% prediction confidence (Successful Non-targeted Attack i.e., when $C(\mathcal{X}_\mathcal{D})\in \mathbb{C}$).}
    \end{subfigure}\hfill
    \begin{subfigure}[t]{0.32\linewidth}
        \centering
        \includegraphics[width=0.8\linewidth]{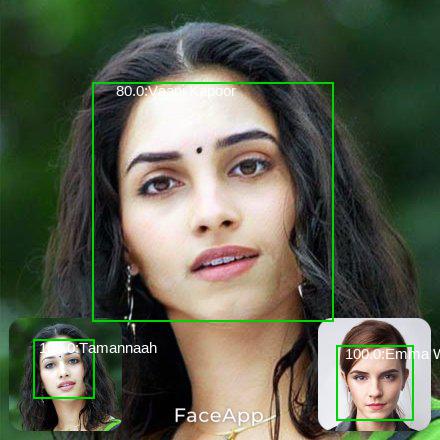}
        \caption{``Tamannaah'' blend with ``Emma Stone'' is predicted as ``Vanni Kapoor'' with 80\% prediction confidence (Successful Non-targeted Attack i.e., when $C(\mathcal{X}_\mathcal{D})\in \mathbb{C}$).}
    \end{subfigure}\hfill
    \begin{subfigure}[t]{0.32\linewidth}
        \centering
        \includegraphics[width=0.8\linewidth]{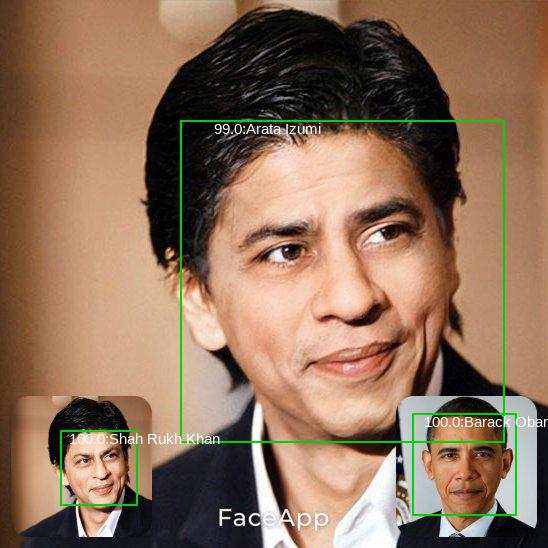}
        \caption{``ShahRukh Khan'' blend with ``Barack Obama'' is predicted as ``Arata Izumi'' with 99\% prediction confidence (Successful Non-targeted Attack i.e., when $C(\mathcal{X}_\mathcal{D})\in \mathbb{C}$).}
    \end{subfigure}%
    \caption{Additional results from \textit{Amazon Rekognition API} ($1^{st}$ row), \textit{Microsoft Azure Cognitive Services API} ($2^{nd}$ row), and \textit{Naver Clova Face Recognition API} ($3^{rd}$ row) on CelebBlend dataset. Note: In this figure, we show the Asian celebrities' results for NAV API because NAV API performs well on mostly Asian Celebrities.}
    \label{fig:CelebBlend_appendix_2}
\end{figure*}

\section{Appendix: Racial Distribution of Datasets and Predictions from API}
\label{Appendix: Racial Distribution of Datasets and Prediciton from API}
\begin{figure*}[hbt!]
    \begin{subfigure}[t]{0.46\linewidth}
        \centering
        \includegraphics[width=\linewidth]{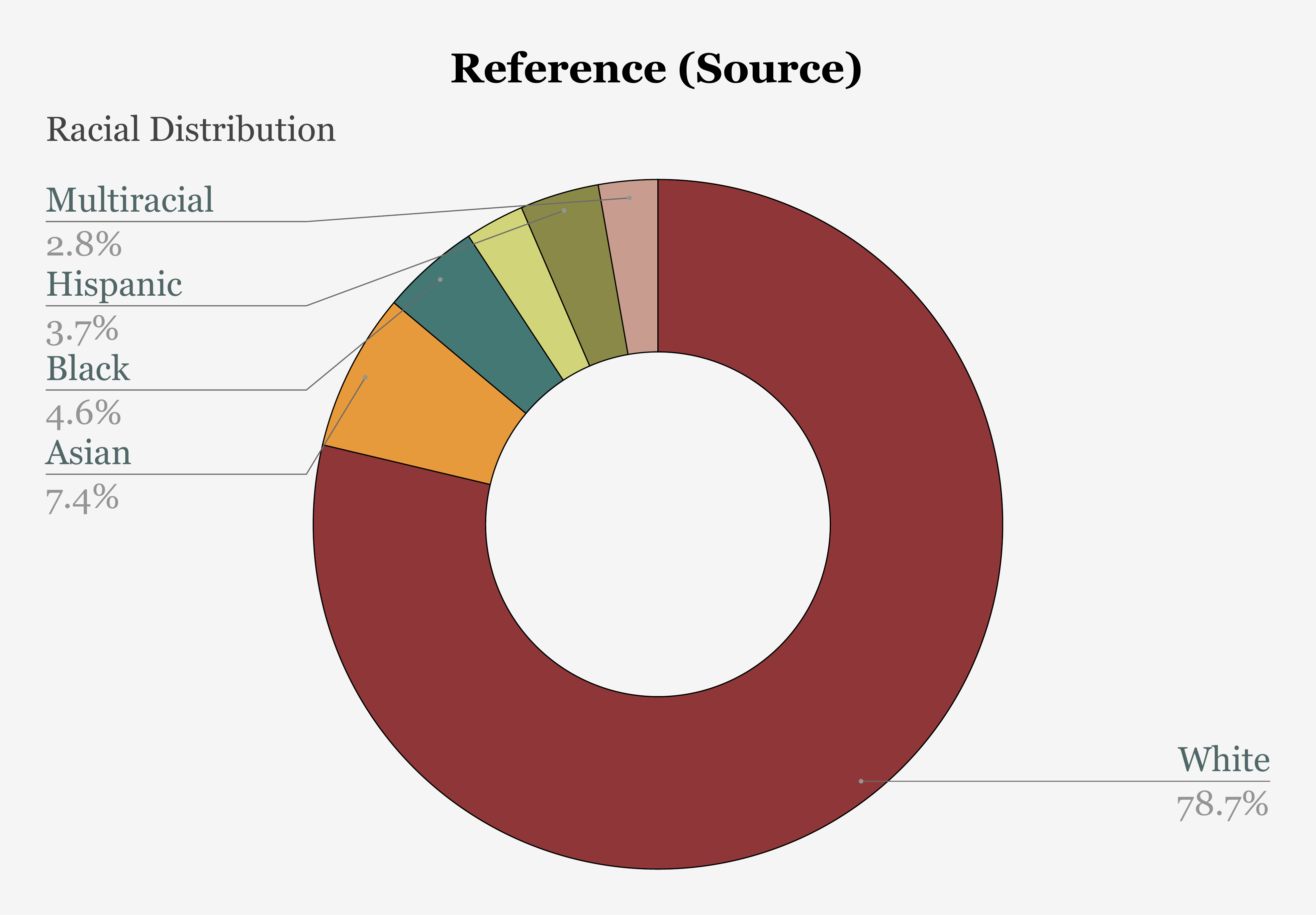}
        \caption{}
    \end{subfigure}\hfill
    \begin{subfigure}[t]{0.46\linewidth}
        \centering
        \includegraphics[width=\linewidth]{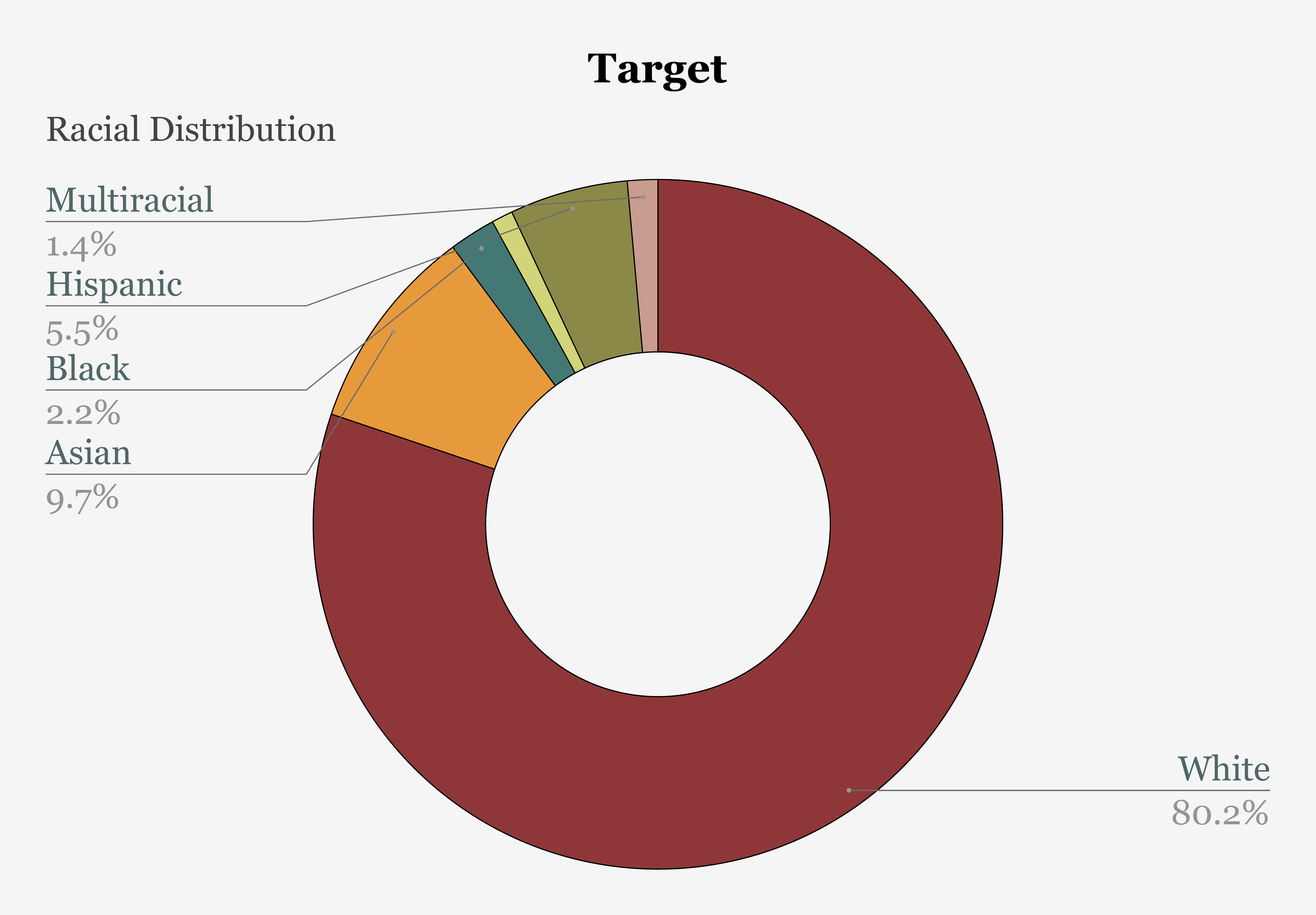}
        \caption{}
    \end{subfigure}\hfill
    \begin{subfigure}[t]{0.46\linewidth}
        \centering
        \includegraphics[width=\linewidth]{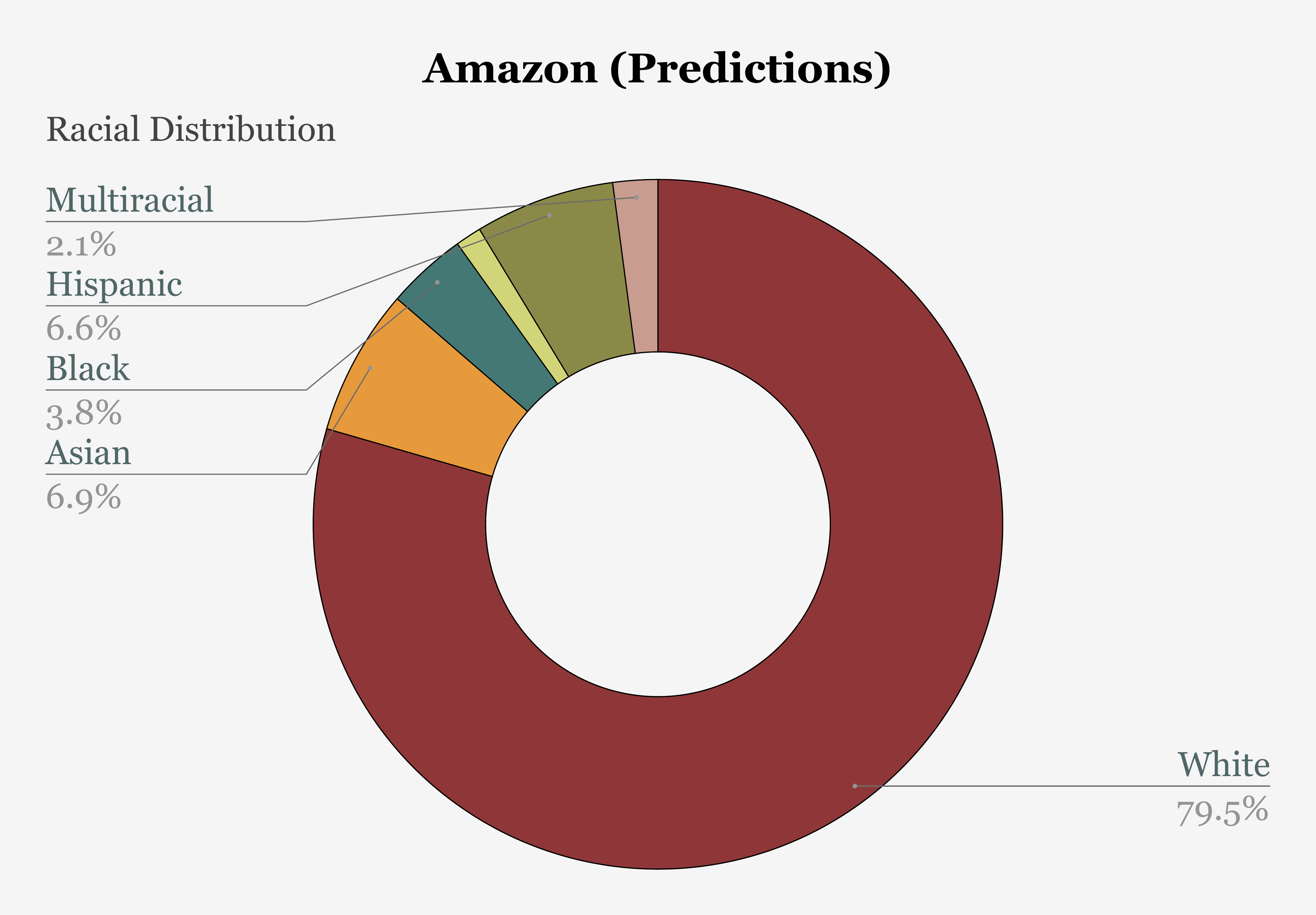}
        \caption{}
    \end{subfigure}\hfill
    \begin{subfigure}[t]{0.46\linewidth}
        \centering
        \includegraphics[width=\linewidth]{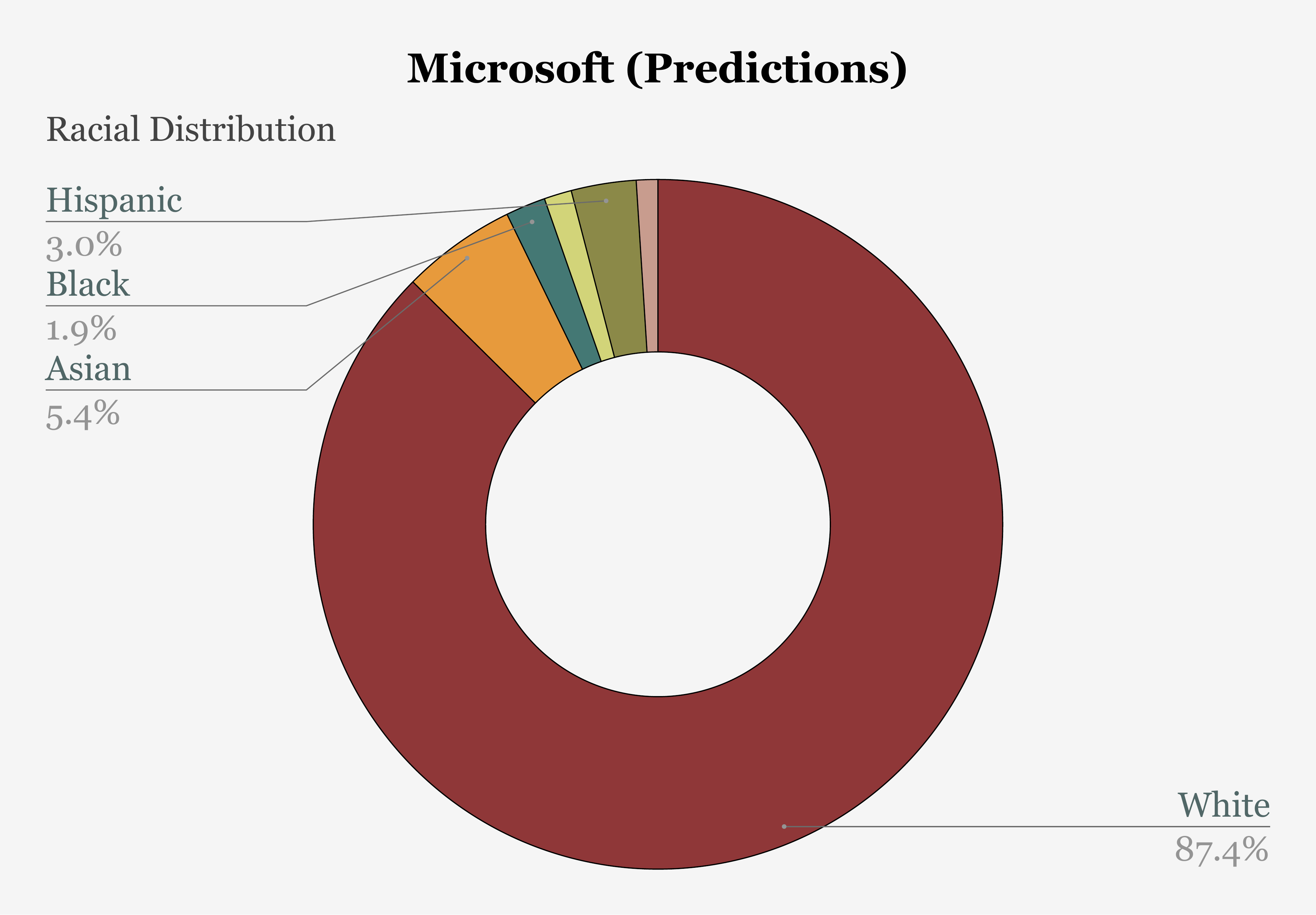}
        \caption{}
    \end{subfigure}%
    \hfill
    \begin{subfigure}[t]{0.46\linewidth}
        \centering
        \includegraphics[width=\linewidth]{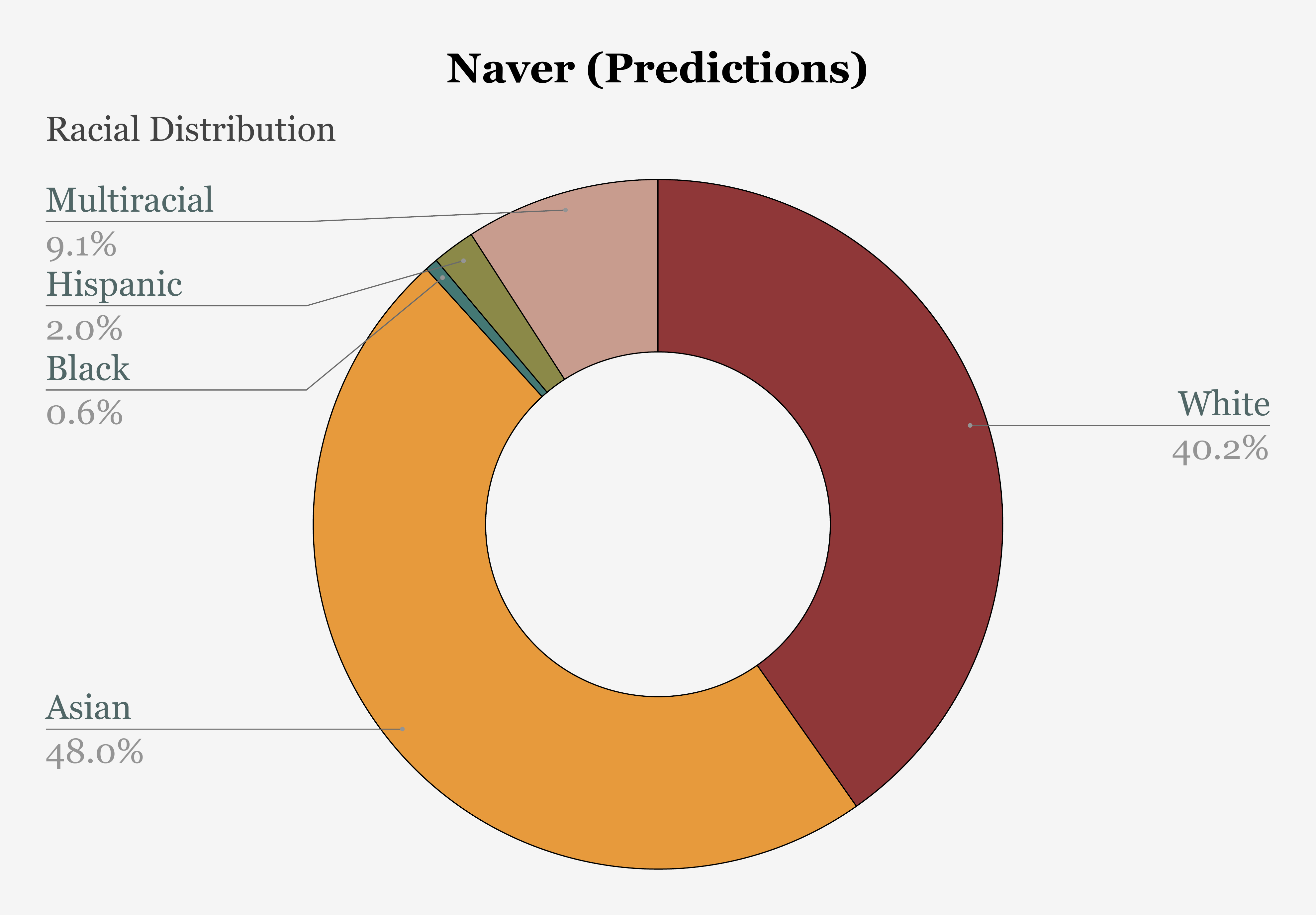}
        \caption{}
    \end{subfigure}%
    \caption{\underline{Racial Distribution of Datasets and Predictions from APIs}: We can observe that the major trend is that most of the datasets are biased toward Whites people (a and b). Moreover, the Western APIs (b and c) vs. Asian API (e) also shows a bias toward their respective races in their predictions.}
    \label{fig:Racial_Distribution_appendix}
\end{figure*}

\onecolumn
\section{Appendix: Detailed results for each Dataset}
\label{Appendix: Detailed results for each Dataset}

% \usepackage[longtable]{multirow}
% \usepackage{colortbl}
% \usepackage{longtable}
% \usepackage{hhline}

% \clearpage
\arrayrulecolor{black}
\begin{longtable}{|l|c|c|c|c|c|c|c|c|} 
\caption{Detailed performance measurements for CelebDF Dataset using Amazon, Microsoft and Naver APIs.}
\label{tab:AppendixCelebDFDataset}\\ 
\hline
\rowcolor[rgb]{0.553,0.388,0.455} \multicolumn{1}{|c|}{{\cellcolor[rgb]{0.553,0.388,0.455}}} & \multicolumn{3}{c|}{\begin{tabular}[c]{@{}>{\cellcolor[rgb]{0.553,0.388,0.455}}c@{}}\textcolor{white}{\textbf{Average Prediction Confidence}}\\\textcolor{white}{\textbf{Amazon API (\%)}}\end{tabular}} & \multicolumn{3}{c|}{\begin{tabular}[c]{@{}>{\cellcolor[rgb]{0.553,0.388,0.455}}c@{}}\textcolor{white}{\textbf{Average Prediction Confidence}}\\\textcolor{white}{\textbf{}}\textcolor{white}{\textbf{Microsoft API}}\textcolor{white}{~(\%)}\end{tabular}} & \multicolumn{2}{c|}{\begin{tabular}[c]{@{}>{\cellcolor[rgb]{0.553,0.388,0.455}}c@{}}\textbf{\textcolor{white}{Average Prediction Confidence}}\\\textbf{\textcolor{white}{}\textcolor{white}{Naver API }}\textcolor{white}{(\%)}\end{tabular}} \\*
\rowcolor[rgb]{0.824,0.749,0.78} \multicolumn{1}{|c|}{\multirow{-2}{*}{{\cellcolor[rgb]{0.553,0.388,0.455}}\begin{tabular}[c]{@{}>{\cellcolor[rgb]{0.553,0.388,0.455}}c@{}}\textcolor{white}{\textbf{Celebrity}}\\\textcolor{white}{\textbf{Name}} \end{tabular}}} & \textit{\textbf{Reference}} & \textit{\textbf{Target}} & \textit{\textbf{Similarity}} & \textit{\textbf{Reference}} & \textit{\textbf{Target}} & \textit{\textbf{Similarity}} & \textit{\textbf{Reference}} & \textit{\textbf{Target}} \endfirsthead 

\hline
\rowcolor[rgb]{0.553,0.388,0.455} \multicolumn{1}{|c|}{{\cellcolor[rgb]{0.553,0.388,0.455}}} & \multicolumn{3}{c|}{\begin{tabular}[c]{@{}>{\cellcolor[rgb]{0.553,0.388,0.455}}c@{}}\textcolor{white}{\textbf{Average Prediction Confidence}}\\\textcolor{white}{\textbf{Amazon API (\%)}}\end{tabular}} & \multicolumn{3}{c|}{\begin{tabular}[c]{@{}>{\cellcolor[rgb]{0.553,0.388,0.455}}c@{}}\textcolor{white}{\textbf{Average Prediction Confidence}}\\\textcolor{white}{\textbf{}}\textcolor{white}{\textbf{Microsoft API}}\textcolor{white}{~(\%)}\end{tabular}} & \multicolumn{2}{c|}{\begin{tabular}[c]{@{}>{\cellcolor[rgb]{0.553,0.388,0.455}}c@{}}\textbf{\textcolor{white}{Average Prediction Confidence}}\\\textbf{\textcolor{white}{}\textcolor{white}{Naver API }}\textcolor{white}{(\%)}\end{tabular}} \\*
\rowcolor[rgb]{0.824,0.749,0.78} \multicolumn{1}{|c|}{\multirow{-2}{*}{{\cellcolor[rgb]{0.553,0.388,0.455}}\begin{tabular}[c]{@{}>{\cellcolor[rgb]{0.553,0.388,0.455}}c@{}}\textcolor{white}{\textbf{Celebrity}}\\\textcolor{white}{\textbf{Name}} \end{tabular}}} & \textit{\textbf{Reference}} & \textit{\textbf{Target}} & \textit{\textbf{Similarity}} & \textit{\textbf{Reference}} & \textit{\textbf{Target}} & \textit{\textbf{Similarity}} & \textit{\textbf{Reference}} & \textit{\textbf{Target}}
\\\hline
\endhead % all the lines above this will be repeated on every page
\hline
\endfoot
\hline
\endlastfoot
\rowcolor[rgb]{0.914,0.871,0.886} {\cellcolor[rgb]{0.553,0.388,0.455}}\textcolor{white}{Aamir Khan} & 96 & 82 & 68 & 98 & 86 & 31 & 28 & 30 \\ 
\hline
\rowcolor[rgb]{0.824,0.749,0.78} {\cellcolor[rgb]{0.553,0.388,0.455}}\textcolor{white}{Angelina Jolie} & 96 & 86 & 90 & 99 & 86 & 63 & 27 & 28 \\ 
\hline
\rowcolor[rgb]{0.914,0.871,0.886} {\cellcolor[rgb]{0.553,0.388,0.455}}\textcolor{white}{Anne Hathaway} & 84 & 78 & 81 & 99 & 92 & 49 & 46 & 40 \\ 
\hline
\rowcolor[rgb]{0.824,0.749,0.78} {\cellcolor[rgb]{0.553,0.388,0.455}}\textcolor{white}{Ben Affleck} & 84 & 79 & 83 & 99 & 88 & 55 & 29 & 28 \\ 
\hline
\rowcolor[rgb]{0.914,0.871,0.886} {\cellcolor[rgb]{0.553,0.388,0.455}}\textcolor{white}{Brad Pitt} & 98 & 80 & 77 & 89 & 91 & 43 & 37 & 39 \\ 
\hline
\rowcolor[rgb]{0.824,0.749,0.78} {\cellcolor[rgb]{0.553,0.388,0.455}}\textcolor{white}{Cameron Diaz} & 100 & 79 & 81 & 97 & 90 & 40 & 14 & 20 \\ 
\hline
\rowcolor[rgb]{0.914,0.871,0.886} {\cellcolor[rgb]{0.553,0.388,0.455}}\textcolor{white}{Carrie-Anne Moss} & 99 & 82 & 81 & 88 & 88 & 36 & 19 & 26 \\ 
\hline
\rowcolor[rgb]{0.824,0.749,0.78} {\cellcolor[rgb]{0.553,0.388,0.455}}\textcolor{white}{Cate Blanchett} & 91 & 79 & 76 & 95 & 82 & 59 & 23 & 28 \\ 
\hline
\rowcolor[rgb]{0.914,0.871,0.886} {\cellcolor[rgb]{0.553,0.388,0.455}}\textcolor{white}{Charlize Theron} & 82 & 87 & 67 & 96 & 91 & 35 & 82 & 37 \\ 
\hline
\rowcolor[rgb]{0.824,0.749,0.78} {\cellcolor[rgb]{0.553,0.388,0.455}}\textcolor{white}{Chloë Grace Moretz} & 88 & 72 & 77 & 89 & 93 & 58 & 21 & 31 \\ 
\hline
\rowcolor[rgb]{0.914,0.871,0.886} {\cellcolor[rgb]{0.553,0.388,0.455}}\textcolor{white}{Chris Evans} & 87 & 75 & 72 & 98 & 94 & 49 & 37 & 26 \\ 
\hline
\rowcolor[rgb]{0.824,0.749,0.78} {\cellcolor[rgb]{0.553,0.388,0.455}}\textcolor{white}{Chris Hemsworth} & 93 & 88 & 74 & 93 & 89 & 59 & 26 & 33 \\ 
\hline
\rowcolor[rgb]{0.914,0.871,0.886} {\cellcolor[rgb]{0.553,0.388,0.455}}\textcolor{white}{Chris Pine} & 96 & 76 & 76 & 100 & 90 & 44 & 29 & 31 \\ 
\hline
\rowcolor[rgb]{0.824,0.749,0.78} {\cellcolor[rgb]{0.553,0.388,0.455}}\textcolor{white}{Colin Farrell} & 99 & 80 & 83 & 100 & 89 & 63 & 25 & 23 \\ 
\hline
\rowcolor[rgb]{0.914,0.871,0.886} {\cellcolor[rgb]{0.553,0.388,0.455}}\textcolor{white}{Denzel Washington} & 90 & 81 & 90 & 93 & 91 & 63 & 32 & 23 \\ 
\hline
\rowcolor[rgb]{0.824,0.749,0.78} {\cellcolor[rgb]{0.553,0.388,0.455}}\textcolor{white}{Don Cheadle} & 99 & 69 & 90 & 100 & 0 & 60 & 45 & 18 \\ 
\hline
\rowcolor[rgb]{0.914,0.871,0.886} {\cellcolor[rgb]{0.553,0.388,0.455}}\textcolor{white}{Edward Norton} & 95 & 83 & 82 & 100 & 92 & 55 & 15 & 24 \\ 
\hline
\rowcolor[rgb]{0.824,0.749,0.78} {\cellcolor[rgb]{0.553,0.388,0.455}}\textcolor{white}{Emma Stone} & 96 & 80 & 63 & 94 & 88 & 55 & 26 & 22 \\ 
\hline
\rowcolor[rgb]{0.914,0.871,0.886} {\cellcolor[rgb]{0.553,0.388,0.455}}\textcolor{white}{Emma Watson} & 93 & 77 & 81 & 100 & 82 & 58 & 54 & 32 \\ 
\hline
\rowcolor[rgb]{0.824,0.749,0.78} {\cellcolor[rgb]{0.553,0.388,0.455}}\textcolor{white}{Ethan Hawke} & 98 & 69 & 86 & 100 & 94 & 59 & 45 & 30 \\ 
\hline
\rowcolor[rgb]{0.914,0.871,0.886} {\cellcolor[rgb]{0.553,0.388,0.455}}\textcolor{white}{Eva Mendes} & 96 & 71 & 66 & 99 & 90 & 33 & 21 & 37 \\ 
\hline
\rowcolor[rgb]{0.824,0.749,0.78} {\cellcolor[rgb]{0.553,0.388,0.455}}\textcolor{white}{Evangeline Lilly} & 89 & 74 & 81 & 97 & 82 & 62 & 58 & 50 \\ 
\hline
\rowcolor[rgb]{0.914,0.871,0.886} {\cellcolor[rgb]{0.553,0.388,0.455}}\textcolor{white}{Famke Janssen} & 99 & 67 & 79 & 100 & 88 & 50 & 57 & 34 \\ 
\hline
\rowcolor[rgb]{0.824,0.749,0.78} {\cellcolor[rgb]{0.553,0.388,0.455}}\textcolor{white}{Gal Gadot} & 92 & 76 & 80 & 95 & 84 & 48 & 37 & 31 \\ 
\hline
\rowcolor[rgb]{0.914,0.871,0.886} {\cellcolor[rgb]{0.553,0.388,0.455}}\textcolor{white}{Gerard Butler} & 100 & 84 & 86 & 99 & 86 & 70 & 35 & 38 \\ 
\hline
\rowcolor[rgb]{0.824,0.749,0.78} {\cellcolor[rgb]{0.553,0.388,0.455}}\textcolor{white}{Gwyneth Paltrow} & 94 & 73 & 82 & 96 & 82 & 51 & 22 & 22 \\ 
\hline
\rowcolor[rgb]{0.914,0.871,0.886} {\cellcolor[rgb]{0.553,0.388,0.455}}\textcolor{white}{Jake Gyllenhaal} & 96 & 76 & 83 & 100 & 89 & 45 & 21 & 25 \\ 
\hline
\rowcolor[rgb]{0.824,0.749,0.78} {\cellcolor[rgb]{0.553,0.388,0.455}}\textcolor{white}{Jamie Foxx} & 91 & 80 & 89 & 99 & 84 & 62 & 20 & 16 \\ 
\hline
\rowcolor[rgb]{0.914,0.871,0.886} {\cellcolor[rgb]{0.553,0.388,0.455}}\textcolor{white}{Jason Statham} & 96 & 79 & 81 & 99 & 83 & 63 & 25 & 45 \\ 
\hline
\rowcolor[rgb]{0.824,0.749,0.78} {\cellcolor[rgb]{0.553,0.388,0.455}}\textcolor{white}{Jennifer Aniston} & 93 & 81 & 82 & 99 & 91 & 42 & 24 & 36 \\ 
\hline
\rowcolor[rgb]{0.914,0.871,0.886} {\cellcolor[rgb]{0.553,0.388,0.455}}\textcolor{white}{Jennifer Lawrence} & 93 & 77 & 83 & 84 & 91 & 38 & 21 & 19 \\ 
\hline
\rowcolor[rgb]{0.824,0.749,0.78} {\cellcolor[rgb]{0.553,0.388,0.455}}\textcolor{white}{Jessica Alba} & 89 & 82 & 83 & 98 & 90 & 21 & 38 & 19 \\ 
\hline
\rowcolor[rgb]{0.914,0.871,0.886} {\cellcolor[rgb]{0.553,0.388,0.455}}\textcolor{white}{Jim Carrey} & 87 & 71 & 83 & 91 & 90 & 57 & 18 & 25 \\ 
\hline
\rowcolor[rgb]{0.824,0.749,0.78} {\cellcolor[rgb]{0.553,0.388,0.455}}\textcolor{white}{John Travolta} & 100 & 77 & 83 & 100 & 90 & 58 & 32 & 33 \\ 
\hline
\rowcolor[rgb]{0.914,0.871,0.886} {\cellcolor[rgb]{0.553,0.388,0.455}}\textcolor{white}{Gordon-Levitt} & 100 & 70 & 84 & 97 & 93 & 53 & 29 & 29 \\ 
\hline
\rowcolor[rgb]{0.824,0.749,0.78} {\cellcolor[rgb]{0.553,0.388,0.455}}\textcolor{white}{Jude Law} & 92 & 80 & 86 & 100 & 90 & 61 & 46 & 42 \\ 
\hline
\rowcolor[rgb]{0.914,0.871,0.886} {\cellcolor[rgb]{0.553,0.388,0.455}}\textcolor{white}{Justin Timberlake} & 97 & 74 & 83 & 100 & 89 & 64 & 23 & 30 \\ 
\hline
\rowcolor[rgb]{0.824,0.749,0.78} {\cellcolor[rgb]{0.553,0.388,0.455}}\textcolor{white}{Kate Beckinsale} & 91 & 79 & 83 & 100 & 91 & 45 & 64 & 51 \\ 
\hline
\rowcolor[rgb]{0.914,0.871,0.886} {\cellcolor[rgb]{0.553,0.388,0.455}}\textcolor{white}{Kate Winslet} & 85 & 80 & 76 & 96 & 84 & 53 & 35 & 27 \\ 
\hline
\rowcolor[rgb]{0.824,0.749,0.78} {\cellcolor[rgb]{0.553,0.388,0.455}}\textcolor{white}{Keira Knightley} & 93 & 76 & 86 & 97 & 90 & 60 & 26 & 30 \\ 
\hline
\rowcolor[rgb]{0.914,0.871,0.886} {\cellcolor[rgb]{0.553,0.388,0.455}}\textcolor{white}{Leonardo DiCaprio} & 99 & 73 & 88 & 97 & 86 & 64 & 20 & 27 \\ 
\hline
\rowcolor[rgb]{0.824,0.749,0.78} {\cellcolor[rgb]{0.553,0.388,0.455}}\textcolor{white}{Liam Neeson} & 97 & 75 & 86 & 100 & 90 & 38 & 28 & 31 \\ 
\hline
\rowcolor[rgb]{0.914,0.871,0.886} {\cellcolor[rgb]{0.553,0.388,0.455}}\textcolor{white}{Mark Wahlberg} & 100 & 71 & 84 & 100 & 89 & 59 & 100 & 90 \\ 
\hline
\rowcolor[rgb]{0.824,0.749,0.78} {\cellcolor[rgb]{0.553,0.388,0.455}}\textcolor{white}{Megan Fox} & 91 & 79 & 86 & 94 & 88 & 38 & 78 & 83 \\ 
\hline
\rowcolor[rgb]{0.914,0.871,0.886} {\cellcolor[rgb]{0.553,0.388,0.455}}\textcolor{white}{Miley Cyrus} & 89 & 87 & 82 & 97 & 91 & 54 & 26 & 25 \\ 
\hline
\rowcolor[rgb]{0.824,0.749,0.78} {\cellcolor[rgb]{0.553,0.388,0.455}}\textcolor{white}{Natalie Portman} & 94 & 81 & 84 & 97 & 90 & 49 & 28 & 31 \\ 
\hline
\rowcolor[rgb]{0.914,0.871,0.886} {\cellcolor[rgb]{0.553,0.388,0.455}}\textcolor{white}{Nicole Kidman} & 98 & 85 & 81 & 90 & 86 & 67 & 25 & 31 \\ 
\hline
\rowcolor[rgb]{0.824,0.749,0.78} {\cellcolor[rgb]{0.553,0.388,0.455}}\textcolor{white}{Randall Park} & 100 & 73 & 77 & 100 & 86 & 32 & 14 & 25 \\ 
\hline
\rowcolor[rgb]{0.914,0.871,0.886} {\cellcolor[rgb]{0.553,0.388,0.455}}\textcolor{white}{Russell Crowe} & 95 & 71 & 87 & 100 & 90 & 57 & 37 & 29 \\ 
\hline
\rowcolor[rgb]{0.824,0.749,0.78} {\cellcolor[rgb]{0.553,0.388,0.455}}\textcolor{white}{Ryan Gosling} & 95 & 78 & 87 & 100 & 90 & 61 & 30 & 39 \\ 
\hline
\rowcolor[rgb]{0.914,0.871,0.886} {\cellcolor[rgb]{0.553,0.388,0.455}}\textcolor{white}{Ryan Reynolds} & 100 & 83 & 78 & 100 & 89 & 55 & 49 & 43 \\ 
\hline
\rowcolor[rgb]{0.824,0.749,0.78} {\cellcolor[rgb]{0.553,0.388,0.455}}\textcolor{white}{Scarlett Johansson} & 95 & 74 & 86 & 95 & 79 & 52 & 40 & 35 \\ 
\hline
\rowcolor[rgb]{0.914,0.871,0.886} {\cellcolor[rgb]{0.553,0.388,0.455}}\textcolor{white}{Sean Penn} & 93 & 70 & 79 & 99 & 90 & 48 & 42 & 43 \\ 
\hline
\rowcolor[rgb]{0.824,0.749,0.78} {\cellcolor[rgb]{0.553,0.388,0.455}}\textcolor{white}{Shia LaBeouf} & 98 & 83 & 83 & 96 & 88 & 63 & 34 & 28 \\ 
\hline
\rowcolor[rgb]{0.914,0.871,0.886} {\cellcolor[rgb]{0.553,0.388,0.455}}\textcolor{white}{Tom Hanks} & 98 & 83 & 82 & 96 & 87 & 52 & 20 & 26 \\ 
\hline
\rowcolor[rgb]{0.824,0.749,0.78} {\cellcolor[rgb]{0.553,0.388,0.455}}\textcolor{white}{Will Ferrell} & 94 & 86 & 83 & 100 & 90 & 38 & 22 & 27 \\ 
\hline
\rowcolor[rgb]{0.914,0.871,0.886} {\cellcolor[rgb]{0.553,0.388,0.455}}\textcolor{white}{Will Smith} & 100 & 83 & 95 & 100 & 90 & 51 & 21 & 16 \\ 
\hline
\rowcolor[rgb]{0.824,0.749,0.78} {\cellcolor[rgb]{0.553,0.388,0.455}}\textcolor{white}{Yang Lan} & 100 & 66 & 78 & 96 & 90 & 3 & 30 & 0 \\ 
\hline
\rowcolor[rgb]{0.914,0.871,0.886} {\cellcolor[rgb]{0.553,0.388,0.455}}\textcolor{black}{\textbf{Average}} & \textbf{94.3} & \textbf{77.8} & \textbf{81.4} & \textbf{97.1} & \textbf{86.9} & \textbf{50.9} & \textbf{33.7} & \textbf{31.7} \\
\hline
\end{longtable}

\arrayrulecolor{black}
\begin{longtable}{|l|c|c|c|c|c|c|c|c|} 
\caption{Detailed performance measurements for FCelebDF Dataset using Amazon, Microsoft and Naver APIs.}
\label{tab:AppendixFCelebDFDataset}\\ 
\hline
\rowcolor[rgb]{0.573,0.663,0.725} \multicolumn{1}{|c|}{{\cellcolor[rgb]{0.573,0.663,0.725}}} & \multicolumn{3}{c|}{\begin{tabular}[c]{@{}>{\cellcolor[rgb]{0.573,0.663,0.725}}c@{}}\textbf{\textcolor{white}{Average Prediction Confidence}}\\\textbf{\textcolor{white}{Amazon API (\%) }}\end{tabular}} & \multicolumn{3}{c|}{\begin{tabular}[c]{@{}>{\cellcolor[rgb]{0.573,0.663,0.725}}c@{}}\textbf{\textcolor{white}{Average Prediction Confidence}}\\\textbf{\textcolor{white}{Microsoft API (\%) }}\end{tabular}} & \multicolumn{2}{c|}{\begin{tabular}[c]{@{}>{\cellcolor[rgb]{0.573,0.663,0.725}}c@{}}\textbf{\textcolor{white}{Average Prediction Confidence}}\\\textbf{\textcolor{white}{Naver API (\%) }}\end{tabular}} \\* 
\hhline{|>{\arrayrulecolor[rgb]{0.573,0.663,0.725}}->{\arrayrulecolor{black}}--------|}
\rowcolor[rgb]{0.827,0.863,0.886} \multicolumn{1}{|c|}{\multirow{-2}{*}{{\cellcolor[rgb]{0.573,0.663,0.725}}\begin{tabular}[c]{@{}>{\cellcolor[rgb]{0.573,0.663,0.725}}c@{}}\textbf{\textcolor{white}{Celebrity}}\\\textbf{\textcolor{white}{Name}} \end{tabular}}} & \textit{\textbf{Reference}} & \textit{\textbf{Target}} & \textit{\textbf{Similarity}} & \textit{\textbf{Reference}} & \textit{\textbf{Target}} & \textit{\textbf{Similarity}} & \textit{\textbf{Reference}} & \textit{\textbf{Target}} \endfirsthead 

\hline
\rowcolor[rgb]{0.573,0.663,0.725} \multicolumn{1}{|c|}{{\cellcolor[rgb]{0.573,0.663,0.725}}} & \multicolumn{3}{c|}{\begin{tabular}[c]{@{}>{\cellcolor[rgb]{0.573,0.663,0.725}}c@{}}\textbf{\textcolor{white}{Average Prediction Confidence}}\\\textbf{\textcolor{white}{Amazon API (\%) }}\end{tabular}} & \multicolumn{3}{c|}{\begin{tabular}[c]{@{}>{\cellcolor[rgb]{0.573,0.663,0.725}}c@{}}\textbf{\textcolor{white}{Average Prediction Confidence}}\\\textbf{\textcolor{white}{Microsoft API (\%) }}\end{tabular}} & \multicolumn{2}{c|}{\begin{tabular}[c]{@{}>{\cellcolor[rgb]{0.573,0.663,0.725}}c@{}}\textbf{\textcolor{white}{Average Prediction Confidence}}\\\textbf{\textcolor{white}{Naver API (\%) }}\end{tabular}} \\* 
\hhline{|>{\arrayrulecolor[rgb]{0.573,0.663,0.725}}->{\arrayrulecolor{black}}--------|}
\rowcolor[rgb]{0.827,0.863,0.886} \multicolumn{1}{|c|}{\multirow{-2}{*}{{\cellcolor[rgb]{0.573,0.663,0.725}}\begin{tabular}[c]{@{}>{\cellcolor[rgb]{0.573,0.663,0.725}}c@{}}\textbf{\textcolor{white}{Celebrity}}\\\textbf{\textcolor{white}{Name}} \end{tabular}}} & \textit{\textbf{Reference}} & \textit{\textbf{Target}} & \textit{\textbf{Similarity}} & \textit{\textbf{Reference}} & \textit{\textbf{Target}} & \textit{\textbf{Similarity}} & \textit{\textbf{Reference}} & \textit{\textbf{Target}}
\endhead % all the lines above this will be repeated on every page
\hline
\endfoot
\hline
\endlastfoot
\hline
\rowcolor[rgb]{0.914,0.933,0.945} {\cellcolor[rgb]{0.573,0.663,0.725}}\textcolor{white}{Bae Suzy} & 100 & 78 & 7 & 100 & 97 & 2 & 100 & 23 \\ 
\hline
\rowcolor[rgb]{0.827,0.863,0.886} {\cellcolor[rgb]{0.573,0.663,0.725}}\textcolor{white}{Alexis Bledel} & 100 & 74 & 59 & 100 & 85 & 1 & 71 & 20 \\ 
\hline
\rowcolor[rgb]{0.914,0.933,0.945} {\cellcolor[rgb]{0.573,0.663,0.725}}\textcolor{white}{Emma Watson} & 100 & 71 & 38 & 100 & 75 & 0 & 54 & 26 \\ 
\hline
\rowcolor[rgb]{0.827,0.863,0.886} {\cellcolor[rgb]{0.573,0.663,0.725}}\textcolor{white}{Megan Fox} & 100 & 87 & 76 & 100 & 89 & 1 & 100 & 96 \\ 
\hline
\rowcolor[rgb]{0.914,0.933,0.945} {\cellcolor[rgb]{0.573,0.663,0.725}}\textcolor{white}{Kaley Cuoco} & 100 & 73 & 79 & 100 & 90 & 5 & 100 & 29 \\ 
\hline
\rowcolor[rgb]{0.827,0.863,0.886} {\cellcolor[rgb]{0.573,0.663,0.725}}\textcolor{white}{Emma Stone} & 100 & 80 & 74 & 100 & 77 & 2 & 39 & 20 \\ 
\hline
\rowcolor[rgb]{0.914,0.933,0.945} {\cellcolor[rgb]{0.573,0.663,0.725}}\textcolor{white}{Margot Robbie} & 100 & 83 & 81 & 100 & 86 & 1 & 74 & 29 \\ 
\hline
\rowcolor[rgb]{0.827,0.863,0.886} {\cellcolor[rgb]{0.573,0.663,0.725}}\textcolor{white}{Kate Mara} & 100 & 74 & 50 & 100 & 77 & 0 & 100 & 38 \\ 
\hline
\rowcolor[rgb]{0.914,0.933,0.945} {\cellcolor[rgb]{0.573,0.663,0.725}}\textcolor{white}{Natalie Portman} & 98 & 76 & 23 & 97 & 86 & 0 & 40 & 26 \\ 
\hline
\rowcolor[rgb]{0.827,0.863,0.886} {\cellcolor[rgb]{0.573,0.663,0.725}}\textcolor{white}{Maisie Williams} & 100 & 88 & 56 & 100 & 82 & 5 & 45 & 25 \\ 
\hline
\rowcolor[rgb]{0.914,0.933,0.945} {\cellcolor[rgb]{0.573,0.663,0.725}}\textcolor{white}{Im Yoon-ah} & 100 & 76 & 59 & 99 & 86 & 12 & 100 & 28 \\ 
\hline
\rowcolor[rgb]{0.827,0.863,0.886} {\cellcolor[rgb]{0.573,0.663,0.725}}\textcolor{white}{Kari Byron} & 90 & 74 & 56 & 100 & 86 & 0 & 44 & 35 \\ 
\hline
\rowcolor[rgb]{0.914,0.933,0.945} {\cellcolor[rgb]{0.573,0.663,0.725}}\textcolor{white}{Seohyun} & 96 & 71 & 4 & 98 & 0 & 0 & 100 & 0 \\ 
\hline
\rowcolor[rgb]{0.827,0.863,0.886} {\cellcolor[rgb]{0.573,0.663,0.725}}\textcolor{white}{iJustine} & 100 & 73 & 78 & 100 & 81 & 0 & 19 & 20 \\ 
\hline
\rowcolor[rgb]{0.914,0.933,0.945} {\cellcolor[rgb]{0.573,0.663,0.725}}\textcolor{white}{Jennifer Lopez} & 100 & 70 & 6 & 100 & 0 & 0 & 21 & 0 \\ 
\hline
\rowcolor[rgb]{0.827,0.863,0.886} {\cellcolor[rgb]{0.573,0.663,0.725}}\textcolor{white}{Katheryn Winnick} & 98 & 78 & 9 & 100 & 75 & 0 & 19 & 17 \\ 
\hline
\rowcolor[rgb]{0.914,0.933,0.945} {\cellcolor[rgb]{0.573,0.663,0.725}}\textcolor{white}{Evangeline Lilly} & 0 & 77 & 86 & 93 & 82 & 0 & 100 & 31 \\ 
\hline
\rowcolor[rgb]{0.827,0.863,0.886} {\cellcolor[rgb]{0.573,0.663,0.725}}\textcolor{white}{Jennifer Lawrence} & 98 & 77 & 58 & 99 & 83 & 0 & 15 & 21 \\ 
\hline
\rowcolor[rgb]{0.914,0.933,0.945} {\cellcolor[rgb]{0.573,0.663,0.725}}\textcolor{white}{Gal Gadot} & 100 & 77 & 73 & 100 & 89 & 1 & 23 & 24 \\ 
\hline
\rowcolor[rgb]{0.827,0.863,0.886} {\cellcolor[rgb]{0.573,0.663,0.725}}\textcolor{white}{Daisy Ridley} & 98 & 79 & 61 & 99 & 95 & 1 & 21 & 28 \\ 
\hline
\rowcolor[rgb]{0.914,0.933,0.945} {\cellcolor[rgb]{0.573,0.663,0.725}}\textcolor{white}{IU} & 97 & 81 & 83 & 88 & 70 & 8 & 100 & 61 \\ 
\hline
\rowcolor[rgb]{0.827,0.863,0.886} {\cellcolor[rgb]{0.573,0.663,0.725}}\textcolor{white}{Mila Kunis} & 99 & 93 & 87 & 100 & 85 & 0 & 42 & 36 \\ 
\hline
\rowcolor[rgb]{0.914,0.933,0.945} {\cellcolor[rgb]{0.573,0.663,0.725}}\textcolor{white}{Bryce Dallas Howard} & 100 & 79 & 0 & 100 & 83 & 0 & 26 & 24 \\ 
\hline
\rowcolor[rgb]{0.827,0.863,0.886} {\cellcolor[rgb]{0.573,0.663,0.725}}\textcolor{white}{Brittany Snow} & 100 & 64 & 55 & 100 & 86 & 0 & 46 & 20 \\ 
\hline
\rowcolor[rgb]{0.914,0.933,0.945} {\cellcolor[rgb]{0.573,0.663,0.725}}\textcolor{white}{Selena Gomez} & 91 & 80 & 45 & 100 & 91 & 1 & 39 & 22 \\ 
\hline
\rowcolor[rgb]{0.827,0.863,0.886} {\cellcolor[rgb]{0.573,0.663,0.725}}\textcolor{white}{Jennifer Aniston} & 99 & 78 & 30 & 100 & 82 & 0 & 77 & 22 \\ 
\hline
\rowcolor[rgb]{0.914,0.933,0.945} {\cellcolor[rgb]{0.573,0.663,0.725}}\textcolor{white}{Gillian Anderson} & 100 & 76 & 33 & 100 & 90 & 1 & 100 & 26 \\ 
\hline
\rowcolor[rgb]{0.827,0.863,0.886} {\cellcolor[rgb]{0.573,0.663,0.725}}\textcolor{white}{Seulgi} & 100 & 75 & 63 & 0 & 0 & 0 & 0 & 0 \\ 
\hline
\rowcolor[rgb]{0.914,0.933,0.945} {\cellcolor[rgb]{0.573,0.663,0.725}}\textcolor{white}{Emilia Clarke} & 86 & 80 & 45 & 86 & 0 & 3 & 14 & 0 \\ 
\hline
\rowcolor[rgb]{0.827,0.863,0.886} {\cellcolor[rgb]{0.573,0.663,0.725}}\textcolor{white}{Taylor Swift} & 100 & 77 & 58 & 100 & 0 & 0 & 45 & 0 \\ 
\hline
\rowcolor[rgb]{0.914,0.933,0.945} {\cellcolor[rgb]{0.573,0.663,0.725}}\textcolor{white}{Natalie Dormer} & 100 & 70 & 51 & 100 & 0 & 1 & 49 & 0 \\ 
\hline
\rowcolor[rgb]{0.827,0.863,0.886} {\cellcolor[rgb]{0.573,0.663,0.725}}\textcolor{white}{Ariana Grande} & 100 & 85 & 63 & 100 & 82 & 3 & 28 & 20 \\ 
\hline
\rowcolor[rgb]{0.914,0.933,0.945} {\cellcolor[rgb]{0.573,0.663,0.725}}\textcolor{white}{Alexandra Daddario} & 99 & 76 & 12 & 100 & 0 & 0 & 15 & 0 \\ 
\hline
\rowcolor[rgb]{0.827,0.863,0.886} {\cellcolor[rgb]{0.573,0.663,0.725}}\textcolor{white}{Katy Perry} & 100 & 79 & 72 & 100 & 84 & 1 & 50 & 26 \\ 
\hline
\rowcolor[rgb]{0.914,0.933,0.945} {\cellcolor[rgb]{0.573,0.663,0.725}}\textcolor{white}{Katherine McNamara} & 97 & 74 & 63 & 99 & 82 & 0 & 0 & 29 \\ 
\hline
\rowcolor[rgb]{0.827,0.863,0.886} {\cellcolor[rgb]{0.573,0.663,0.725}}\textcolor{white}{Kristen Stewart} & 100 & 68 & 50 & 100 & 82 & 1 & 43 & 15 \\ 
\hline
\rowcolor[rgb]{0.914,0.933,0.945} {\cellcolor[rgb]{0.573,0.663,0.725}}\textcolor{white}{Cameron Diaz} & 100 & 80 & 80 & 100 & 98 & 0 & 12 & 21 \\ 
\hline
\rowcolor[rgb]{0.827,0.863,0.886} {\cellcolor[rgb]{0.573,0.663,0.725}}\textcolor{white}{Elizabeth Olsen} & 100 & 75 & 5 & 100 & 92 & 0 & 100 & 40 \\ 
\hline
\rowcolor[rgb]{0.914,0.933,0.945} {\cellcolor[rgb]{0.573,0.663,0.725}}\textcolor{white}{Irene} & 0 & 75 & 54 & 0 & 0 & 11 & 100 & 33 \\ 
\hline
\rowcolor[rgb]{0.827,0.863,0.886} {\cellcolor[rgb]{0.573,0.663,0.725}}\textcolor{white}{Kate Upton} & 100 & 70 & 41 & 100 & 85 & 0 & 57 & 33 \\ 
\hline
\rowcolor[rgb]{0.914,0.933,0.945} {\cellcolor[rgb]{0.573,0.663,0.725}}\textcolor{white}{Emma Roberts} & 100 & 81 & 80 & 100 & 82 & 0 & 91 & 24 \\ 
\hline
\rowcolor[rgb]{0.827,0.863,0.886} {\cellcolor[rgb]{0.573,0.663,0.725}}\textcolor{white}{Avril Lavigne} & 100 & 71 & 62 & 100 & 96 & 0 & 57 & 21 \\ 
\hline
\rowcolor[rgb]{0.914,0.933,0.945} {\cellcolor[rgb]{0.573,0.663,0.725}}\textcolor{white}{Chloe Grace Moretz} & 100 & 83 & 60 & 100 & 85 & 0 & 0 & 0 \\ 
\hline
\rowcolor[rgb]{0.827,0.863,0.886} {\cellcolor[rgb]{0.573,0.663,0.725}}\textcolor{white}{Sofia Vergara} & 100 & 81 & 53 & 100 & 86 & 0 & 0 & 0 \\ 
\hline
\rowcolor[rgb]{0.914,0.933,0.945} {\cellcolor[rgb]{0.573,0.663,0.725}}\textcolor{white}{Kristin Kreuk} & 100 & 79 & 78 & 100 & 79 & 1 & 32 & 33 \\ 
\hline
\rowcolor[rgb]{0.827,0.863,0.886} {\cellcolor[rgb]{0.573,0.663,0.725}}\textcolor{white}{Scarlett Johansson} & 100 & 80 & 78 & 100 & 92 & 0 & 60 & 28 \\ 
\hline
\rowcolor[rgb]{0.914,0.933,0.945} {\cellcolor[rgb]{0.573,0.663,0.725}}\textcolor{white}{Kaitlin Olson} & 99 & 74 & 54 & 100 & 73 & 1 & 43 & 28 \\ 
\hline
\rowcolor[rgb]{0.827,0.863,0.886} {\cellcolor[rgb]{0.573,0.663,0.725}}\textcolor{white}{Angelina Jolie} & 98 & 72 & 60 & 100 & 0 & 1 & 20 & 0 \\ 
\hline
\rowcolor[rgb]{0.914,0.933,0.945} {\cellcolor[rgb]{0.573,0.663,0.725}}\textcolor{white}{Sophie Turner} & 100 & 67 & 33 & 100 & 0 & 2 & 36 & 0 \\ 
\hline
\rowcolor[rgb]{0.827,0.863,0.886} {\cellcolor[rgb]{0.573,0.663,0.725}}\textcolor{white}{June Lapine} & 0 & 78 & 78 & 0 & 0 & 12 & 0 & 0 \\ 
\hline
\rowcolor[rgb]{0.914,0.933,0.945} {\cellcolor[rgb]{0.573,0.663,0.725}}\textcolor{black}{\textbf{Average}} & \textbf{92.9} & \textbf{76.7} & \textbf{53.2} & \textbf{93.2} & \textbf{66.1} & \textbf{1.6} & \textbf{49.3} & \textbf{22.0} \\
\hline
\end{longtable}

\arrayrulecolor{black}
\begin{longtable}{|l|c|c|c|c|c|c|c|c|} 
\caption{Detailed performance measurements for VoxCelebTH Dataset using Amazon, Microsoft and Naver APIs.}
\label{tab:AppendixVoxCelebTHDataset}\\  
\hline
\rowcolor[rgb]{0.608,0.451,0.384} {\cellcolor[rgb]{0.608,0.451,0.384}} & \multicolumn{3}{c|}{\begin{tabular}[c]{@{}>{\cellcolor[rgb]{0.608,0.451,0.384}}c@{}}\textbf{\textcolor{white}{Average Prediction Confidence}}\\\textbf{\textcolor{white}{Amazon API (\%) }}\end{tabular}} & \multicolumn{3}{c|}{\begin{tabular}[c]{@{}>{\cellcolor[rgb]{0.608,0.451,0.384}}c@{}}\textbf{\textcolor{white}{Average Prediction Confidence}}\\\textbf{\textcolor{white}{Microsoft API (\%) }}\end{tabular}} & \multicolumn{2}{c|}{\begin{tabular}[c]{@{}>{\cellcolor[rgb]{0.608,0.451,0.384}}c@{}}\textbf{\textcolor{white}{Average Prediction Confidence}}\\\textbf{\textcolor{white}{Naver API (\%) }}\end{tabular}} \\* 
\hhline{|>{\arrayrulecolor[rgb]{0.608,0.451,0.384}}->{\arrayrulecolor{black}}--------|}
\rowcolor[rgb]{0.843,0.78,0.749} \multirow{-2}{*}{{\cellcolor[rgb]{0.608,0.451,0.384}}\textbf{\textcolor{white}{Celebrity Name }}} & \textbf{\textit{Reference}} & \textbf{\textit{Target}} & \textbf{\textit{Similarity}} & \textbf{\textit{Reference}} & \textbf{\textit{Target}} & \textbf{\textit{Similarity}} & \textbf{\textit{Reference}} & \textbf{\textit{Target}} \endfirsthead 

\hline
\rowcolor[rgb]{0.608,0.451,0.384} {\cellcolor[rgb]{0.608,0.451,0.384}} & \multicolumn{3}{c|}{\begin{tabular}[c]{@{}>{\cellcolor[rgb]{0.608,0.451,0.384}}c@{}}\textbf{\textcolor{white}{Average Prediction Confidence}}\\\textbf{\textcolor{white}{Amazon API (\%) }}\end{tabular}} & \multicolumn{3}{c|}{\begin{tabular}[c]{@{}>{\cellcolor[rgb]{0.608,0.451,0.384}}c@{}}\textbf{\textcolor{white}{Average Prediction Confidence}}\\\textbf{\textcolor{white}{Microsoft API (\%) }}\end{tabular}} & \multicolumn{2}{c|}{\begin{tabular}[c]{@{}>{\cellcolor[rgb]{0.608,0.451,0.384}}c@{}}\textbf{\textcolor{white}{Average Prediction Confidence}}\\\textbf{\textcolor{white}{Naver API (\%) }}\end{tabular}} \\* 
\hhline{|>{\arrayrulecolor[rgb]{0.608,0.451,0.384}}->{\arrayrulecolor{black}}--------|}
\rowcolor[rgb]{0.843,0.78,0.749} \multirow{-2}{*}{{\cellcolor[rgb]{0.608,0.451,0.384}}\textbf{\textcolor{white}{Celebrity Name }}} & \textbf{\textit{Reference}} & \textbf{\textit{Target}} & \textbf{\textit{Similarity}} & \textbf{\textit{Reference}} & \textbf{\textit{Target}} & \textbf{\textit{Similarity}} & \textbf{\textit{Reference}} & \textbf{\textit{Target}} \\\hline
\endhead % all the lines above this will be repeated on every page
\hline
\endfoot
\hline
\endlastfoot
\rowcolor[rgb]{0.922,0.886,0.871} {\cellcolor[rgb]{0.608,0.451,0.384}}\textcolor{white}{Aaron Yoo} & 100 & 100 & 99 & 100 & 100 & 65 & 13 & 26 \\ 
\hline
\rowcolor[rgb]{0.843,0.78,0.749} {\cellcolor[rgb]{0.608,0.451,0.384}}\textcolor{white}{Abigail Spencer} & 100 & 85 & 96 & 100 & 93 & 52 & 56 & 31 \\ 
\hline
\rowcolor[rgb]{0.922,0.886,0.871} {\cellcolor[rgb]{0.608,0.451,0.384}}\textcolor{white}{Alain Delon} & 0 & 78 & 90 & 0 & 99 & 0 & 32 & 27 \\ 
\hline
\rowcolor[rgb]{0.843,0.78,0.749} {\cellcolor[rgb]{0.608,0.451,0.384}}\textcolor{white}{Alex Pettyfer} & 100 & 75 & 88 & 100 & 88 & 43 & 36 & 26 \\ 
\hline
\rowcolor[rgb]{0.922,0.886,0.871} {\cellcolor[rgb]{0.608,0.451,0.384}}\textcolor{white}{Brenda Blethyn} & 100 & 93 & 98 & 100 & 98 & 52 & 46 & 30 \\ 
\hline
\rowcolor[rgb]{0.843,0.78,0.749} {\cellcolor[rgb]{0.608,0.451,0.384}}\textcolor{white}{Chris Hemsworth} & 99 & 98 & 99 & 94 & 90 & 73 & 47 & 31 \\ 
\hline
\rowcolor[rgb]{0.922,0.886,0.871} {\cellcolor[rgb]{0.608,0.451,0.384}}\textcolor{white}{Cote de Pablo} & 0 & 93 & 94 & 100 & 97 & 45 & 74 & 33 \\ 
\hline
\rowcolor[rgb]{0.843,0.78,0.749} {\cellcolor[rgb]{0.608,0.451,0.384}}\textcolor{white}{Cristin Milioti} & 100 & 96 & 97 & 100 & 95 & 63 & 14 & 41 \\ 
\hline
\rowcolor[rgb]{0.922,0.886,0.871} {\cellcolor[rgb]{0.608,0.451,0.384}}\textcolor{white}{Damon Lindelof} & 100 & 89 & 99 & 100 & 98 & 60 & 27 & 63 \\ 
\hline
\rowcolor[rgb]{0.843,0.78,0.749} {\cellcolor[rgb]{0.608,0.451,0.384}}\textcolor{white}{Danny Pino} & 100 & 96 & 99 & 100 & 100 & 69 & 26 & 21 \\ 
\hline
\rowcolor[rgb]{0.922,0.886,0.871} {\cellcolor[rgb]{0.608,0.451,0.384}}\textcolor{white}{Dermot Mulroney} & 100 & 97 & 98 & 100 & 100 & 60 & 47 & 37 \\ 
\hline
\rowcolor[rgb]{0.843,0.78,0.749} {\cellcolor[rgb]{0.608,0.451,0.384}}\textcolor{white}{Dylan Moran} & 100 & 90 & 96 & 100 & 95 & 26 & 45 & 26 \\ 
\hline
\rowcolor[rgb]{0.922,0.886,0.871} {\cellcolor[rgb]{0.608,0.451,0.384}}\textcolor{white}{Edgar Wright} & 100 & 96 & 98 & 100 & 97 & 69 & 61 & 17 \\ 
\hline
\rowcolor[rgb]{0.843,0.78,0.749} {\cellcolor[rgb]{0.608,0.451,0.384}}\textcolor{white}{Heidi Montag} & 100 & 82 & 94 & 98 & 86 & 16 & 1 & 24 \\ 
\hline
\rowcolor[rgb]{0.922,0.886,0.871} {\cellcolor[rgb]{0.608,0.451,0.384}}\textcolor{white}{Izabella Scorupco} & 100 & 92 & 88 & 100 & 95 & 2 & 33 & 35 \\ 
\hline
\rowcolor[rgb]{0.843,0.78,0.749} {\cellcolor[rgb]{0.608,0.451,0.384}}\textcolor{white}{Jake Abel} & 97 & 87 & 96 & 100 & 97 & 53 & 1 & 27 \\ 
\hline
\rowcolor[rgb]{0.922,0.886,0.871} {\cellcolor[rgb]{0.608,0.451,0.384}}\textcolor{white}{Jeremy Shada} & 100 & 78 & 96 & 100 & 79 & 48 & 41 & 23 \\ 
\hline
\rowcolor[rgb]{0.843,0.78,0.749} {\cellcolor[rgb]{0.608,0.451,0.384}}\textcolor{white}{John Carroll Lynch} & 100 & 100 & 99 & 100 & 98 & 66 & 100 & 49 \\ 
\hline
\rowcolor[rgb]{0.922,0.886,0.871} {\cellcolor[rgb]{0.608,0.451,0.384}}\textcolor{white}{John Hurt} & 100 & 88 & 96 & 100 & 91 & 55 & 35 & 30 \\ 
\hline
\rowcolor[rgb]{0.843,0.78,0.749} {\cellcolor[rgb]{0.608,0.451,0.384}}\textcolor{white}{John Lloyd Young} & 100 & 88 & 99 & 100 & 100 & 69 & 42 & 25 \\ 
\hline
\rowcolor[rgb]{0.922,0.886,0.871} {\cellcolor[rgb]{0.608,0.451,0.384}}\textcolor{white}{Jordan Gavaris} & 99 & 91 & 97 & 100 & 98 & 56 & 46 & 15 \\ 
\hline
\rowcolor[rgb]{0.843,0.78,0.749} {\cellcolor[rgb]{0.608,0.451,0.384}}\textcolor{white}{Julian Sands} & 97 & 98 & 98 & 100 & 96 & 53 & 37 & 35 \\ 
\hline
\rowcolor[rgb]{0.922,0.886,0.871} {\cellcolor[rgb]{0.608,0.451,0.384}}\textcolor{white}{Kellan Lutz} & 100 & 95 & 97 & 100 & 100 & 58 & 9 & 28 \\ 
\hline
\rowcolor[rgb]{0.843,0.78,0.749} {\cellcolor[rgb]{0.608,0.451,0.384}}\textcolor{white}{Kenny Rogers} & 56 & 76 & 98 & 100 & 99 & 71 & 17 & 26 \\ 
\hline
\rowcolor[rgb]{0.922,0.886,0.871} {\cellcolor[rgb]{0.608,0.451,0.384}}\textcolor{white}{Kim Raver} & 100 & 77 & 96 & 100 & 91 & 53 & 42 & 66 \\ 
\hline
\rowcolor[rgb]{0.843,0.78,0.749} {\cellcolor[rgb]{0.608,0.451,0.384}}\textcolor{white}{Kristen Stewart} & 100 & 95 & 96 & 100 & 97 & 54 & 53 & 39 \\ 
\hline
\rowcolor[rgb]{0.922,0.886,0.871} {\cellcolor[rgb]{0.608,0.451,0.384}}\textcolor{white}{Lauryn Hill} & 100 & 98 & 96 & 100 & 97 & 45 & 47 & 23 \\ 
\hline
\rowcolor[rgb]{0.843,0.78,0.749} {\cellcolor[rgb]{0.608,0.451,0.384}}\textcolor{white}{Luke Evans} & 100 & 95 & 97 & 100 & 97 & 57 & 61 & 30 \\ 
\hline
\rowcolor[rgb]{0.922,0.886,0.871} {\cellcolor[rgb]{0.608,0.451,0.384}}\textcolor{white}{Mandy Patinkin} & 59 & 75 & 93 & 100 & 87 & 0 & 34 & 26 \\ 
\hline
\rowcolor[rgb]{0.843,0.78,0.749} {\cellcolor[rgb]{0.608,0.451,0.384}}\textcolor{white}{Marion Cotillard} & 100 & 98 & 99 & 100 & 99 & 68 & 14 & 23 \\ 
\hline
\rowcolor[rgb]{0.922,0.886,0.871} {\cellcolor[rgb]{0.608,0.451,0.384}}\textcolor{white}{Max Thieriot} & 100 & 88 & 94 & 100 & 96 & 53 & 21 & 17 \\ 
\hline
\rowcolor[rgb]{0.843,0.78,0.749} {\cellcolor[rgb]{0.608,0.451,0.384}}\textcolor{white}{Maya Rudolph} & 100 & 91 & 97 & 100 & 96 & 50 & 33 & 33 \\ 
\hline
\rowcolor[rgb]{0.922,0.886,0.871} {\cellcolor[rgb]{0.608,0.451,0.384}}\textcolor{white}{Michelle Stafford} & 100 & 90 & 85 & 100 & 81 & 0 & 34 & 22 \\ 
\hline
\rowcolor[rgb]{0.843,0.78,0.749} {\cellcolor[rgb]{0.608,0.451,0.384}}\textcolor{white}{Nathaniel Buzolic} & 100 & 96 & 89 & 100 & 99 & 45 & 54 & 34 \\ 
\hline
\rowcolor[rgb]{0.922,0.886,0.871} {\cellcolor[rgb]{0.608,0.451,0.384}}\textcolor{white}{Niall Horan} & 100 & 97 & 99 & 100 & 99 & 62 & 15 & 31 \\ 
\hline
\rowcolor[rgb]{0.843,0.78,0.749} {\cellcolor[rgb]{0.608,0.451,0.384}}\textcolor{white}{Nick Thune} & 100 & 89 & 91 & 100 & 91 & 2 & 12 & 19 \\ 
\hline
\rowcolor[rgb]{0.922,0.886,0.871} {\cellcolor[rgb]{0.608,0.451,0.384}}\textcolor{white}{Nicole Richie} & 77 & 92 & 96 & 100 & 98 & 54 & 67 & 20 \\ 
\hline
\rowcolor[rgb]{0.843,0.78,0.749} {\cellcolor[rgb]{0.608,0.451,0.384}}\textcolor{white}{Nina Arianda} & 100 & 82 & 97 & 100 & 98 & 34 & 100 & 85 \\ 
\hline
\rowcolor[rgb]{0.922,0.886,0.871} {\cellcolor[rgb]{0.608,0.451,0.384}}\textcolor{white}{Olivia Colman} & 99 & 92 & 98 & 100 & 95 & 57 & 32 & 27 \\ 
\hline
\rowcolor[rgb]{0.843,0.78,0.749} {\cellcolor[rgb]{0.608,0.451,0.384}}\textcolor{white}{Oscar Isaac} & 100 & 94 & 98 & 100 & 100 & 58 & 43 & 28 \\ 
\hline
\rowcolor[rgb]{0.922,0.886,0.871} {\cellcolor[rgb]{0.608,0.451,0.384}}\textcolor{white}{Paul Scheer} & 100 & 90 & 96 & 100 & 95 & 31 & 29 & 32 \\ 
\hline
\rowcolor[rgb]{0.843,0.78,0.749} {\cellcolor[rgb]{0.608,0.451,0.384}}\textcolor{white}{Preity Zinta} & 100 & 77 & 90 & 100 & 95 & 15 & 17 & 23 \\ 
\hline
\rowcolor[rgb]{0.922,0.886,0.871} {\cellcolor[rgb]{0.608,0.451,0.384}}\textcolor{white}{Rhys Darby} & 77 & 76 & 94 & 100 & 92 & 2 & 51 & 33 \\ 
\hline
\rowcolor[rgb]{0.843,0.78,0.749} {\cellcolor[rgb]{0.608,0.451,0.384}}\textcolor{white}{Rory Kennedy} & 100 & 97 & 98 & 100 & 93 & 54 & 100 & 43 \\ 
\hline
\rowcolor[rgb]{0.922,0.886,0.871} {\cellcolor[rgb]{0.608,0.451,0.384}}\textcolor{white}{RZA} & 0 & 74 & 92 & 0 & 91 & 0 & 1 & 18 \\ 
\hline
\rowcolor[rgb]{0.843,0.78,0.749} {\cellcolor[rgb]{0.608,0.451,0.384}}\textcolor{white}{Sara Ramirez} & 100 & 90 & 91 & 100 & 91 & 54 & 53 & 26 \\ 
\hline
\rowcolor[rgb]{0.922,0.886,0.871} {\cellcolor[rgb]{0.608,0.451,0.384}}\textcolor{white}{Tamara Tunie} & 100 & 92 & 93 & 100 & 85 & 50 & 15 & 31 \\ 
\hline
\rowcolor[rgb]{0.843,0.78,0.749} {\cellcolor[rgb]{0.608,0.451,0.384}}\textcolor{white}{Thomas Jane} & 99 & 77 & 94 & 100 & 93 & 31 & 15 & 31 \\ 
\hline
\rowcolor[rgb]{0.922,0.886,0.871} {\cellcolor[rgb]{0.608,0.451,0.384}}\textcolor{white}{Todd Lasance} & 100 & 99 & 99 & 100 & 99 & 67 & 28 & 48 \\ 
\hline
\rowcolor[rgb]{0.843,0.78,0.749} {\cellcolor[rgb]{0.608,0.451,0.384}}\textcolor{white}{Tracy Morgan} & 99 & 97 & 100 & 100 & 100 & 67 & 45 & 32 \\ 
\hline
\rowcolor[rgb]{0.922,0.886,0.871} {\cellcolor[rgb]{0.608,0.451,0.384}}\textcolor{white}{Aaron Rodgers} & 0 & 65 & 99 & 100 & 100 & 59 & 18 & 21 \\ 
\hline
\rowcolor[rgb]{0.843,0.78,0.749} {\cellcolor[rgb]{0.608,0.451,0.384}}\textcolor{white}{Adam Rodriguez} & 100 & 95 & 98 & 0 & 0 & 32 & 26 & 44 \\ 
\hline
\rowcolor[rgb]{0.922,0.886,0.871} {\cellcolor[rgb]{0.608,0.451,0.384}}\textcolor{white}{Aditya Roy Kapur} & 100 & 80 & 98 & 0 & 89 & 0 & 43 & 36 \\ 
\hline
\rowcolor[rgb]{0.843,0.78,0.749} {\cellcolor[rgb]{0.608,0.451,0.384}}\textcolor{white}{Ajay Devgan} & 100 & 98 & 99 & 100 & 99 & 60 & 28 & 34 \\ 
\hline
\rowcolor[rgb]{0.922,0.886,0.871} {\cellcolor[rgb]{0.608,0.451,0.384}}\textcolor{white}{Andy Roddick} & 100 & 85 & 99 & 100 & 99 & 18 & 40 & 45 \\ 
\hline
\rowcolor[rgb]{0.843,0.78,0.749} {\cellcolor[rgb]{0.608,0.451,0.384}}\textcolor{white}{Ashley Tisdale} & 100 & 100 & 99 & 100 & 98 & 68 & 35 & 25 \\ 
\hline
\rowcolor[rgb]{0.922,0.886,0.871} {\cellcolor[rgb]{0.608,0.451,0.384}}\textcolor{white}{Bill Bailey} & 100 & 97 & 98 & 100 & 98 & 66 & 44 & 26 \\ 
\hline
\rowcolor[rgb]{0.843,0.78,0.749} {\cellcolor[rgb]{0.608,0.451,0.384}}\textcolor{white}{Billy Ray Cyrus} & 100 & 64 & 86 & 100 & 0 & 0 & 22 & 16 \\ 
\hline
\rowcolor[rgb]{0.922,0.886,0.871} {\cellcolor[rgb]{0.608,0.451,0.384}}\textcolor{white}{Brie Bella} & 99 & 78 & 98 & 81 & 92 & 0 & 95 & 21 \\ 
\hline
\rowcolor[rgb]{0.843,0.78,0.749} {\cellcolor[rgb]{0.608,0.451,0.384}}\textcolor{white}{Britney Spears} & 99 & 96 & 94 & 100 & 86 & 49 & 40 & 73 \\ 
\hline
\rowcolor[rgb]{0.922,0.886,0.871} {\cellcolor[rgb]{0.608,0.451,0.384}}\textcolor{white}{Katherine Jenkins} & 100 & 99 & 84 & 0 & 0 & 60 & 26 & 30 \\ 
\hline
\rowcolor[rgb]{0.843,0.78,0.749} {\cellcolor[rgb]{0.608,0.451,0.384}}\textcolor{white}{Charlie Sheen} & 100 & 71 & 93 & 100 & 98 & 41 & 69 & 35 \\ 
\hline
\rowcolor[rgb]{0.922,0.886,0.871} {\cellcolor[rgb]{0.608,0.451,0.384}}\textcolor{white}{Chris Christie} & 100 & 86 & 100 & 100 & 100 & 70 & 100 & 82 \\ 
\hline
\rowcolor[rgb]{0.843,0.78,0.749} {\cellcolor[rgb]{0.608,0.451,0.384}}\textcolor{white}{David Beckham} & 100 & 78 & 99 & 100 & 97 & 43 & 100 & 66 \\ 
\hline
\rowcolor[rgb]{0.922,0.886,0.871} {\cellcolor[rgb]{0.608,0.451,0.384}}\textcolor{white}{David Spade} & 100 & 93 & 96 & 100 & 91 & 51 & 100 & 28 \\ 
\hline
\rowcolor[rgb]{0.843,0.78,0.749} {\cellcolor[rgb]{0.608,0.451,0.384}}\textcolor{white}{Debby Ryan} & 100 & 89 & 81 & 0 & 0 & 18 & 21 & 32 \\ 
\hline
\rowcolor[rgb]{0.922,0.886,0.871} {\cellcolor[rgb]{0.608,0.451,0.384}}\textcolor{white}{Elisabeth Hasselbeck} & 100 & 77 & 93 & 100 & 94 & 2 & 71 & 58 \\ 
\hline
\rowcolor[rgb]{0.843,0.78,0.749} {\cellcolor[rgb]{0.608,0.451,0.384}}\textcolor{white}{Shannon Elizabeth} & 100 & 76 & 91 & 0 & 0 & 21 & 30 & 18 \\ 
\hline
\rowcolor[rgb]{0.922,0.886,0.871} {\cellcolor[rgb]{0.608,0.451,0.384}}\textcolor{white}{Emmy Rossum} & 100 & 85 & 91 & 100 & 86 & 16 & 8 & 13 \\ 
\hline
\rowcolor[rgb]{0.843,0.78,0.749} {\cellcolor[rgb]{0.608,0.451,0.384}}\textcolor{white}{Fan Bingbing} & 100 & 99 & 99 & 100 & 98 & 69 & 82 & 29 \\ 
\hline
\rowcolor[rgb]{0.922,0.886,0.871} {\cellcolor[rgb]{0.608,0.451,0.384}}\textcolor{white}{Fawad Khan} & 100 & 70 & 93 & 0 & 0 & 42 & 57 & 50 \\ 
\hline
\rowcolor[rgb]{0.843,0.78,0.749} {\cellcolor[rgb]{0.608,0.451,0.384}}\textcolor{white}{Gisele Bündchen} & 100 & 79 & 88 & 0 & 0 & 15 & 65 & 40 \\ 
\hline
\rowcolor[rgb]{0.922,0.886,0.871} {\cellcolor[rgb]{0.608,0.451,0.384}}\textcolor{white}{Harry Connick Jr.} & 100 & 93 & 96 & 0 & 0 & 5 & 32 & 26 \\ 
\hline
\rowcolor[rgb]{0.843,0.78,0.749} {\cellcolor[rgb]{0.608,0.451,0.384}}\textcolor{white}{Heather Morris} & 100 & 68 & 85 & 0 & 0 & 0 & 24 & 27 \\ 
\hline
\rowcolor[rgb]{0.922,0.886,0.871} {\cellcolor[rgb]{0.608,0.451,0.384}}\textcolor{white}{Jack Bauer} & 98 & 86 & 95 & 89 & 96 & 0 & 75 & 47 \\ 
\hline
\rowcolor[rgb]{0.843,0.78,0.749} {\cellcolor[rgb]{0.608,0.451,0.384}}\textcolor{white}{Jenna Dewan} & 100 & 80 & 82 & 100 & 72 & 0 & 94 & 47 \\ 
\hline
\rowcolor[rgb]{0.922,0.886,0.871} {\cellcolor[rgb]{0.608,0.451,0.384}}\textcolor{white}{Jennifer Love Hewitt} & 100 & 76 & 71 & 100 & 0 & 0 & 69 & 13 \\ 
\hline
\rowcolor[rgb]{0.843,0.78,0.749} {\cellcolor[rgb]{0.608,0.451,0.384}}\textcolor{white}{Jermaine Dupri} & 100 & 76 & 99 & 100 & 91 & 53 & 39 & 28 \\ 
\hline
\rowcolor[rgb]{0.922,0.886,0.871} {\cellcolor[rgb]{0.608,0.451,0.384}}\textcolor{white}{Jessie J} & 100 & 71 & 68 & 0 & 0 & 0 & 13 & 9 \\ 
\hline
\rowcolor[rgb]{0.843,0.78,0.749} {\cellcolor[rgb]{0.608,0.451,0.384}}\textcolor{white}{Jordan Henderson} & 100 & 73 & 95 & 100 & 95 & 5 & 100 & 13 \\ 
\hline
\rowcolor[rgb]{0.922,0.886,0.871} {\cellcolor[rgb]{0.608,0.451,0.384}}\textcolor{white}{Justin Bieber} & 98 & 81 & 98 & 0 & 93 & 0 & 11 & 18 \\ 
\hline
\rowcolor[rgb]{0.843,0.78,0.749} {\cellcolor[rgb]{0.608,0.451,0.384}}\textcolor{white}{Karen Gillan} & 84 & 84 & 97 & 100 & 91 & 53 & 63 & 54 \\ 
\hline
\rowcolor[rgb]{0.922,0.886,0.871} {\cellcolor[rgb]{0.608,0.451,0.384}}\textcolor{white}{Katy Perry} & 100 & 83 & 94 & 0 & 0 & 13 & 67 & 10 \\ 
\hline
\rowcolor[rgb]{0.843,0.78,0.749} {\cellcolor[rgb]{0.608,0.451,0.384}}\textcolor{white}{Katie Price} & 100 & 80 & 97 & 100 & 88 & 33 & 32 & 38 \\ 
\hline
\rowcolor[rgb]{0.922,0.886,0.871} {\cellcolor[rgb]{0.608,0.451,0.384}}\textcolor{white}{Kobe Bryant} & 84 & 70 & 99 & 100 & 96 & 62 & 62 & 43 \\ 
\hline
\rowcolor[rgb]{0.843,0.78,0.749} {\cellcolor[rgb]{0.608,0.451,0.384}}\textcolor{white}{Kyra Sedgwick} & 100 & 86 & 96 & 100 & 88 & 44 & 48 & 32 \\ 
\hline
\rowcolor[rgb]{0.922,0.886,0.871} {\cellcolor[rgb]{0.608,0.451,0.384}}\textcolor{white}{LeBron James} & 100 & 71 & 86 & 96 & 0 & 0 & 25 & 22 \\ 
\hline
\rowcolor[rgb]{0.843,0.78,0.749} {\cellcolor[rgb]{0.608,0.451,0.384}}\textcolor{white}{Mario Lopez} & 100 & 98 & 99 & 0 & 0 & 74 & 34 & 26 \\ 
\hline
\rowcolor[rgb]{0.922,0.886,0.871} {\cellcolor[rgb]{0.608,0.451,0.384}}\textcolor{white}{Bear Grylls} & 100 & 81 & 94 & 0 & 0 & 31 & 40 & 40 \\ 
\hline
\rowcolor[rgb]{0.843,0.78,0.749} {\cellcolor[rgb]{0.608,0.451,0.384}}\textcolor{white}{Mike Brey} & 100 & 71 & 97 & 100 & 99 & 50 & 65 & 42 \\ 
\hline
\rowcolor[rgb]{0.922,0.886,0.871} {\cellcolor[rgb]{0.608,0.451,0.384}}\textcolor{white}{Nick Offerman} & 100 & 83 & 92 & 100 & 88 & 37 & 43 & 20 \\ 
\hline
\rowcolor[rgb]{0.843,0.78,0.749} {\cellcolor[rgb]{0.608,0.451,0.384}}\textcolor{white}{Oprah Winfrey} & 100 & 96 & 98 & 100 & 96 & 64 & 23 & 22 \\ 
\hline
\rowcolor[rgb]{0.922,0.886,0.871} {\cellcolor[rgb]{0.608,0.451,0.384}}\textcolor{white}{Neil Patrick Harris} & 100 & 97 & 84 & 0 & 0 & 61 & 60 & 32 \\ 
\hline
\rowcolor[rgb]{0.843,0.78,0.749} {\cellcolor[rgb]{0.608,0.451,0.384}}\textcolor{white}{Roger Federer} & 100 & 100 & 97 & 100 & 98 & 18 & 100 & 13 \\ 
\hline
\rowcolor[rgb]{0.922,0.886,0.871} {\cellcolor[rgb]{0.608,0.451,0.384}}\textcolor{white}{Sebastian Stan} & 100 & 86 & 98 & 100 & 89 & 36 & 1 & 27 \\ 
\hline
\rowcolor[rgb]{0.843,0.78,0.749} {\cellcolor[rgb]{0.608,0.451,0.384}}\textcolor{white}{Sebastián Piñera} & 100 & 93 & 96 & 0 & 0 & 57 & 60 & 26 \\ 
\hline
\rowcolor[rgb]{0.922,0.886,0.871} {\cellcolor[rgb]{0.608,0.451,0.384}}\textcolor{white}{Tico Torres} & 98 & 77 & 99 & 100 & 99 & 59 & 61 & 29 \\ 
\hline
\rowcolor[rgb]{0.843,0.78,0.749} {\cellcolor[rgb]{0.608,0.451,0.384}}\textcolor{white}{Trey Songz} & 100 & 68 & 89 & 100 & 79 & 46 & 34 & 15 \\ 
\hline
\rowcolor[rgb]{0.922,0.886,0.871} {\cellcolor[rgb]{0.608,0.451,0.384}}\textcolor{white}{Triple H} & 66 & 97 & 96 & 100 & 98 & 50 & 57 & 47 \\ 
\hline
\rowcolor[rgb]{0.843,0.78,0.749} {\cellcolor[rgb]{0.608,0.451,0.384}}\textcolor{white}{Vincent Perez} & 100 & 73 & 91 & 0 & 0 & 8 & 46 & 43 \\ 
\hline
\rowcolor[rgb]{0.922,0.886,0.871} {\cellcolor[rgb]{0.608,0.451,0.384}}\textbf{Average} & \textbf{93.8} & \textbf{86.4} & \textbf{94.4} & \textbf{80.6} & \textbf{77.3} & \textbf{39.5} & \textbf{44.0} & \textbf{32.0} \\
\hline
\end{longtable}

\arrayrulecolor{black}
\begin{longtable}{|l|c|c|c|c|c|c|c|c|} 
\caption{Detailed performance measurements for CelebFOM Dataset using Amazon, Microsoft and Naver APIs.}
\label{tab:AppendixCelebFOMDataset}\\  
\hline
\rowcolor[rgb]{0.655,0.718,0.537} {\cellcolor[rgb]{0.655,0.718,0.537}} & \multicolumn{3}{c|}{\begin{tabular}[c]{@{}>{\cellcolor[rgb]{0.655,0.718,0.537}}c@{}}\textbf{\textcolor{white}{Average Prediction Confidence}}\\\textbf{\textcolor{white}{Amazon API (\%) }}\end{tabular}} & \multicolumn{3}{c|}{\begin{tabular}[c]{@{}>{\cellcolor[rgb]{0.655,0.718,0.537}}c@{}}\textbf{\textcolor{white}{Average Prediction Confidence}}\\\textbf{\textcolor{white}{Microsoft API (\%) }}\end{tabular}} & \multicolumn{2}{c|}{\begin{tabular}[c]{@{}>{\cellcolor[rgb]{0.655,0.718,0.537}}c@{}}\textbf{\textcolor{white}{Average Prediction Confidence}}\\\textbf{\textcolor{white}{Naver API (\%) }}\end{tabular}} \\* 
\hhline{|>{\arrayrulecolor[rgb]{0.655,0.718,0.537}}->{\arrayrulecolor{black}}--------|}
\rowcolor[rgb]{0.863,0.886,0.812} \multirow{-2}{*}{{\cellcolor[rgb]{0.655,0.718,0.537}}\textbf{\textcolor{white}{Celebrity Name }}} & \textbf{\textit{Reference}} & \textbf{\textit{Target}} & \textbf{\textit{Similarity}} & \textbf{\textit{Reference}} & \textbf{\textit{Target}} & \textbf{\textit{Similarity}} & \textbf{\textit{Reference}} & \textbf{\textit{Target}} \endfirsthead 

\hline
\rowcolor[rgb]{0.655,0.718,0.537} {\cellcolor[rgb]{0.655,0.718,0.537}} & \multicolumn{3}{c|}{\begin{tabular}[c]{@{}>{\cellcolor[rgb]{0.655,0.718,0.537}}c@{}}\textbf{\textcolor{white}{Average Prediction Confidence}}\\\textbf{\textcolor{white}{Amazon API (\%) }}\end{tabular}} & \multicolumn{3}{c|}{\begin{tabular}[c]{@{}>{\cellcolor[rgb]{0.655,0.718,0.537}}c@{}}\textbf{\textcolor{white}{Average Prediction Confidence}}\\\textbf{\textcolor{white}{Microsoft API (\%) }}\end{tabular}} & \multicolumn{2}{c|}{\begin{tabular}[c]{@{}>{\cellcolor[rgb]{0.655,0.718,0.537}}c@{}}\textbf{\textcolor{white}{Average Prediction Confidence}}\\\textbf{\textcolor{white}{Naver API (\%) }}\end{tabular}} \\* 
\hhline{|>{\arrayrulecolor[rgb]{0.655,0.718,0.537}}->{\arrayrulecolor{black}}--------|}
\rowcolor[rgb]{0.863,0.886,0.812} \multirow{-2}{*}{{\cellcolor[rgb]{0.655,0.718,0.537}}\textbf{\textcolor{white}{Celebrity Name }}} & \textbf{\textit{Reference}} & \textbf{\textit{Target}} & \textbf{\textit{Similarity}} & \textbf{\textit{Reference}} & \textbf{\textit{Target}} & \textbf{\textit{Similarity}} & \textbf{\textit{Reference}} & \textbf{\textit{Target}}\\\hline
\endhead % all the lines above this will be repeated on every page
\hline
\endfoot
\hline
\endlastfoot
\rowcolor[rgb]{0.929,0.941,0.91} {\cellcolor[rgb]{0.655,0.718,0.537}}\textcolor{white}{Aamir Khan} & 94 & 88 & 98 & 99 & 90 & 83 & 33 & 29 \\ 
\hline
\rowcolor[rgb]{0.863,0.886,0.812} {\cellcolor[rgb]{0.655,0.718,0.537}}\textcolor{white}{Angelina Jolie} & 93 & 82 & 96 & 99 & 96 & 68 & 19 & 23 \\ 
\hline
\rowcolor[rgb]{0.929,0.941,0.91} {\cellcolor[rgb]{0.655,0.718,0.537}}\textcolor{white}{Anne Hathaway} & 99 & 84 & 92 & 99 & 97 & 76 & 41 & 29 \\ 
\hline
\rowcolor[rgb]{0.863,0.886,0.812} {\cellcolor[rgb]{0.655,0.718,0.537}}\textcolor{white}{Ben Affleck} & 74 & 83 & 91 & 98 & 91 & 74 & 34 & 40 \\ 
\hline
\rowcolor[rgb]{0.929,0.941,0.91} {\cellcolor[rgb]{0.655,0.718,0.537}}\textcolor{white}{Brad Pitt} & 100 & 88 & 99 & 88 & 97 & 74 & 26 & 23 \\ 
\hline
\rowcolor[rgb]{0.863,0.886,0.812} {\cellcolor[rgb]{0.655,0.718,0.537}}\textcolor{white}{Cameron Diaz} & 92 & 88 & 99 & 99 & 95 & 82 & 19 & 14 \\ 
\hline
\rowcolor[rgb]{0.929,0.941,0.91} {\cellcolor[rgb]{0.655,0.718,0.537}}\textcolor{white}{Carrie-Anne Moss} & 89 & 81 & 100 & 84 & 95 & 82 & 21 & 25 \\ 
\hline
\rowcolor[rgb]{0.863,0.886,0.812} {\cellcolor[rgb]{0.655,0.718,0.537}}\textcolor{white}{Cate Blanchett} & 85 & 76 & 92 & 99 & 98 & 74 & 18 & 18 \\ 
\hline
\rowcolor[rgb]{0.929,0.941,0.91} {\cellcolor[rgb]{0.655,0.718,0.537}}\textcolor{white}{Charlize Theron} & 82 & 80 & 91 & 97 & 96 & 70 & 34 & 29 \\ 
\hline
\rowcolor[rgb]{0.863,0.886,0.812} {\cellcolor[rgb]{0.655,0.718,0.537}}\textcolor{white}{Chloë Grace Moretz} & 91 & 86 & 98 & 96 & 95 & 83 & 27 & 24 \\ 
\hline
\rowcolor[rgb]{0.929,0.941,0.91} {\cellcolor[rgb]{0.655,0.718,0.537}}\textcolor{white}{Chris Evans} & 71 & 84 & 97 & 100 & 96 & 83 & 39 & 27 \\ 
\hline
\rowcolor[rgb]{0.863,0.886,0.812} {\cellcolor[rgb]{0.655,0.718,0.537}}\textcolor{white}{Chris Hemsworth} & 92 & 84 & 97 & 97 & 97 & 83 & 33 & 34 \\ 
\hline
\rowcolor[rgb]{0.929,0.941,0.91} {\cellcolor[rgb]{0.655,0.718,0.537}}\textcolor{white}{Chris Pine} & 98 & 84 & 98 & 99 & 96 & 75 & 29 & 23 \\ 
\hline
\rowcolor[rgb]{0.863,0.886,0.812} {\cellcolor[rgb]{0.655,0.718,0.537}}\textcolor{white}{Colin Farrell} & 99 & 88 & 97 & 100 & 95 & 76 & 31 & 20 \\ 
\hline
\rowcolor[rgb]{0.929,0.941,0.91} {\cellcolor[rgb]{0.655,0.718,0.537}}\textcolor{white}{Denzel Washington} & 98 & 80 & 99 & 89 & 94 & 84 & 32 & 21 \\ 
\hline
\rowcolor[rgb]{0.863,0.886,0.812} {\cellcolor[rgb]{0.655,0.718,0.537}}\textcolor{white}{Don Cheadle} & 97 & 92 & 99 & 100 & 98 & 68 & 33 & 31 \\ 
\hline
\rowcolor[rgb]{0.929,0.941,0.91} {\cellcolor[rgb]{0.655,0.718,0.537}}\textcolor{white}{Edward Norton} & 99 & 91 & 99 & 100 & 98 & 85 & 20 & 20 \\ 
\hline
\rowcolor[rgb]{0.863,0.886,0.812} {\cellcolor[rgb]{0.655,0.718,0.537}}\textcolor{white}{Emma Stone} & 93 & 75 & 99 & 93 & 91 & 76 & 28 & 25 \\ 
\hline
\rowcolor[rgb]{0.929,0.941,0.91} {\cellcolor[rgb]{0.655,0.718,0.537}}\textcolor{white}{Emma Watson} & 93 & 82 & 97 & 98 & 0 & 80 & 38 & 39 \\ 
\hline
\rowcolor[rgb]{0.863,0.886,0.812} {\cellcolor[rgb]{0.655,0.718,0.537}}\textcolor{white}{Ethan Hawke} & 94 & 86 & 98 & 100 & 89 & 85 & 42 & 31 \\ 
\hline
\rowcolor[rgb]{0.929,0.941,0.91} {\cellcolor[rgb]{0.655,0.718,0.537}}\textcolor{white}{Eva Mendes} & 100 & 96 & 97 & 99 & 97 & 77 & 31 & 26 \\ 
\hline
\rowcolor[rgb]{0.863,0.886,0.812} {\cellcolor[rgb]{0.655,0.718,0.537}}\textcolor{white}{Evangeline Lilly} & 92 & 81 & 95 & 98 & 97 & 76 & 43 & 36 \\ 
\hline
\rowcolor[rgb]{0.929,0.941,0.91} {\cellcolor[rgb]{0.655,0.718,0.537}}\textcolor{white}{Famke Janssen} & 98 & 90 & 97 & 99 & 92 & 82 & 45 & 43 \\ 
\hline
\rowcolor[rgb]{0.863,0.886,0.812} {\cellcolor[rgb]{0.655,0.718,0.537}}\textcolor{white}{Gal Gadot} & 88 & 87 & 99 & 98 & 97 & 80 & 36 & 34 \\ 
\hline
\rowcolor[rgb]{0.929,0.941,0.91} {\cellcolor[rgb]{0.655,0.718,0.537}}\textcolor{white}{Gerard Butler} & 94 & 88 & 97 & 100 & 95 & 79 & 36 & 36 \\ 
\hline
\rowcolor[rgb]{0.863,0.886,0.812} {\cellcolor[rgb]{0.655,0.718,0.537}}\textcolor{white}{Gwyneth Paltrow} & 92 & 81 & 97 & 97 & 94 & 77 & 18 & 19 \\ 
\hline
\rowcolor[rgb]{0.929,0.941,0.91} {\cellcolor[rgb]{0.655,0.718,0.537}}\textcolor{white}{Jake Gyllenhaal} & 95 & 93 & 99 & 100 & 93 & 79 & 29 & 24 \\ 
\hline
\rowcolor[rgb]{0.863,0.886,0.812} {\cellcolor[rgb]{0.655,0.718,0.537}}\textcolor{white}{Jamie Foxx} & 85 & 85 & 96 & 98 & 94 & 60 & 17 & 15 \\ 
\hline
\rowcolor[rgb]{0.929,0.941,0.91} {\cellcolor[rgb]{0.655,0.718,0.537}}\textcolor{white}{Jason Statham} & 99 & 90 & 99 & 99 & 83 & 89 & 32 & 32 \\ 
\hline
\rowcolor[rgb]{0.863,0.886,0.812} {\cellcolor[rgb]{0.655,0.718,0.537}}\textcolor{white}{Jennifer Aniston} & 97 & 85 & 95 & 99 & 96 & 72 & 28 & 25 \\ 
\hline
\rowcolor[rgb]{0.929,0.941,0.91} {\cellcolor[rgb]{0.655,0.718,0.537}}\textcolor{white}{Jennifer Lawrence} & 89 & 80 & 99 & 87 & 97 & 80 & 18 & 18 \\ 
\hline
\rowcolor[rgb]{0.863,0.886,0.812} {\cellcolor[rgb]{0.655,0.718,0.537}}\textcolor{white}{Jessica Alba} & 95 & 91 & 98 & 95 & 88 & 77 & 21 & 19 \\ 
\hline
\rowcolor[rgb]{0.929,0.941,0.91} {\cellcolor[rgb]{0.655,0.718,0.537}}\textcolor{white}{Jim Carrey} & 78 & 71 & 97 & 96 & 96 & 79 & 16 & 18 \\ 
\hline
\rowcolor[rgb]{0.863,0.886,0.812} {\cellcolor[rgb]{0.655,0.718,0.537}}\textcolor{white}{John Travolta} & 100 & 93 & 98 & 100 & 95 & 84 & 29 & 31 \\ 
\hline
\rowcolor[rgb]{0.929,0.941,0.91} {\cellcolor[rgb]{0.655,0.718,0.537}}\textcolor{white}{Joseph Gordon-Levitt} & 97 & 94 & 99 & 100 & 96 & 83 & 36 & 28 \\ 
\hline
\rowcolor[rgb]{0.863,0.886,0.812} {\cellcolor[rgb]{0.655,0.718,0.537}}\textcolor{white}{Jude Law} & 100 & 87 & 94 & 99 & 99 & 76 & 45 & 31 \\ 
\hline
\rowcolor[rgb]{0.929,0.941,0.91} {\cellcolor[rgb]{0.655,0.718,0.537}}\textcolor{white}{Justin Timberlake} & 99 & 98 & 99 & 100 & 89 & 85 & 29 & 26 \\ 
\hline
\rowcolor[rgb]{0.863,0.886,0.812} {\cellcolor[rgb]{0.655,0.718,0.537}}\textcolor{white}{Kate Beckinsale} & 85 & 74 & 95 & 85 & 95 & 75 & 55 & 46 \\ 
\hline
\rowcolor[rgb]{0.929,0.941,0.91} {\cellcolor[rgb]{0.655,0.718,0.537}}\textcolor{white}{Kate Winslet} & 94 & 87 & 94 & 95 & 96 & 75 & 16 & 17 \\ 
\hline
\rowcolor[rgb]{0.863,0.886,0.812} {\cellcolor[rgb]{0.655,0.718,0.537}}\textcolor{white}{Keira Knightley} & 88 & 84 & 98 & 95 & 95 & 78 & 33 & 32 \\ 
\hline
\rowcolor[rgb]{0.929,0.941,0.91} {\cellcolor[rgb]{0.655,0.718,0.537}}\textcolor{white}{Leonardo DiCaprio} & 98 & 86 & 97 & 97 & 88 & 84 & 25 & 24 \\ 
\hline
\rowcolor[rgb]{0.863,0.886,0.812} {\cellcolor[rgb]{0.655,0.718,0.537}}\textcolor{white}{Liam Neeson} & 98 & 87 & 97 & 100 & 95 & 83 & 31 & 32 \\ 
\hline
\rowcolor[rgb]{0.929,0.941,0.91} {\cellcolor[rgb]{0.655,0.718,0.537}}\textcolor{white}{Mark Wahlberg} & 94 & 91 & 100 & 100 & 95 & 86 & 90 & 81 \\ 
\hline
\rowcolor[rgb]{0.863,0.886,0.812} {\cellcolor[rgb]{0.655,0.718,0.537}}\textcolor{white}{Megan Fox} & 95 & 89 & 97 & 97 & 96 & 77 & 97 & 79 \\ 
\hline
\rowcolor[rgb]{0.929,0.941,0.91} {\cellcolor[rgb]{0.655,0.718,0.537}}\textcolor{white}{Miley Cyrus} & 83 & 82 & 99 & 96 & 94 & 80 & 28 & 24 \\ 
\hline
\rowcolor[rgb]{0.863,0.886,0.812} {\cellcolor[rgb]{0.655,0.718,0.537}}\textcolor{white}{Natalie Portman} & 98 & 90 & 98 & 97 & 94 & 80 & 29 & 32 \\ 
\hline
\rowcolor[rgb]{0.929,0.941,0.91} {\cellcolor[rgb]{0.655,0.718,0.537}}\textcolor{white}{Nicole Kidman} & 88 & 79 & 98 & 95 & 94 & 75 & 21 & 29 \\ 
\hline
\rowcolor[rgb]{0.863,0.886,0.812} {\cellcolor[rgb]{0.655,0.718,0.537}}\textcolor{white}{Randall Park} & 100 & 95 & 98 & 100 & 96 & 81 & 28 & 27 \\ 
\hline
\rowcolor[rgb]{0.929,0.941,0.91} {\cellcolor[rgb]{0.655,0.718,0.537}}\textcolor{white}{Russell Crowe} & 89 & 91 & 97 & 99 & 93 & 79 & 26 & 31 \\ 
\hline
\rowcolor[rgb]{0.863,0.886,0.812} {\cellcolor[rgb]{0.655,0.718,0.537}}\textcolor{white}{Ryan Gosling} & 99 & 95 & 97 & 100 & 90 & 82 & 25 & 33 \\ 
\hline
\rowcolor[rgb]{0.929,0.941,0.91} {\cellcolor[rgb]{0.655,0.718,0.537}}\textcolor{white}{Ryan Reynolds} & 100 & 89 & 98 & 100 & 94 & 80 & 41 & 39 \\ 
\hline
\rowcolor[rgb]{0.863,0.886,0.812} {\cellcolor[rgb]{0.655,0.718,0.537}}\textcolor{white}{Scarlett Johansson} & 99 & 87 & 100 & 98 & 95 & 74 & 44 & 35 \\ 
\hline
\rowcolor[rgb]{0.929,0.941,0.91} {\cellcolor[rgb]{0.655,0.718,0.537}}\textcolor{white}{Sean Penn} & 92 & 87 & 97 & 98 & 97 & 70 & 48 & 38 \\ 
\hline
\rowcolor[rgb]{0.863,0.886,0.812} {\cellcolor[rgb]{0.655,0.718,0.537}}\textcolor{white}{Shia LaBeouf} & 99 & 90 & 94 & 100 & 96 & 75 & 23 & 23 \\ 
\hline
\rowcolor[rgb]{0.929,0.941,0.91} {\cellcolor[rgb]{0.655,0.718,0.537}}\textcolor{white}{Tom Hanks} & 96 & 89 & 97 & 100 & 98 & 80 & 14 & 19 \\ 
\hline
\rowcolor[rgb]{0.863,0.886,0.812} {\cellcolor[rgb]{0.655,0.718,0.537}}\textcolor{white}{Will Ferrell} & 99 & 91 & 97 & 100 & 95 & 76 & 17 & 19 \\ 
\hline
\rowcolor[rgb]{0.929,0.941,0.91} {\cellcolor[rgb]{0.655,0.718,0.537}}\textcolor{white}{Will Smith} & 95 & 87 & 100 & 100 & 78 & 84 & 16 & 19 \\ 
\hline
\rowcolor[rgb]{0.863,0.886,0.812} {\cellcolor[rgb]{0.655,0.718,0.537}}\textcolor{white}{Yang Lan} & 93 & 91 & 93 & 95 & 96 & 74 & 31 & 26 \\ 
\hline
\rowcolor[rgb]{0.929,0.941,0.91} {\cellcolor[rgb]{0.655,0.718,0.537}}\textbf{Average} & \textbf{93.3} & \textbf{86.4} & \textbf{97.1} & \textbf{97.3} & \textbf{92.6} & \textbf{78.3} & \textbf{31.8} & \textbf{29.2} \\
\hline
\end{longtable}

\end{document}